\documentclass[mnsc, nonblindrev]{informs3} 

\OneAndAHalfSpacedXI 

\usepackage{endnotes}
\usepackage{diagbox}
\usepackage{tikz}
\usepackage{amsmath}
\allowdisplaybreaks
\usepackage{dsfont, pifont, mathrsfs}
\usepackage{multirow}
\usepackage[hidelinks, colorlinks=true, linkcolor=blue, citecolor=blue]{hyperref}
\let\footnote=\endnote
\let\enotesize=\normalsize
\def\notesname{Endnotes}%
\def\makeenmark{$^{\theenmark}$}
\def\enoteformat{\rightskip0pt\leftskip0pt\parindent=1.75em
  \leavevmode\llap{\theenmark.\enskip}}

\usepackage{natbib}
 \bibpunct[, ]{(}{)}{,}{a}{}{,}%
 \def\bibfont{\small}%
\TheoremsNumberedThrough     
\ECRepeatTheorems

\EquationsNumberedThrough    

\newcommand{\cA}{\mathcal A}

\newcommand{\mbP}{\mathbb P}
\newcommand{\mbE}{\mathbb E}

\newcommand{\rad}{\text{\rm rad}}
\newcommand{\CI}{\text{\rm CI}}
\newcommand{\SE}{\text{\rm SE}}
\newcommand{\UCB}{\text{\rm UCB}}
\newcommand{\ThS}{\text{\rm TS}}
\newcommand{\poly}{\mathrm{poly}}
\newcommand{\polylog}{\mathrm{polylog}}

\newcommand*{\circled}[1]{\lower.7ex\hbox{\tikz\draw (0pt, 0pt)%
    circle (.5em) node {\makebox[1em][c]{\small #1}};}}

\usepackage{algorithm}
\usepackage{algorithmicx}
\usepackage{algpseudocode}
\algnewcommand{\algorithmicand}{\textbf{and }}
\algnewcommand{\algorithmicor}{\textbf{or }}
\algnewcommand{\OR}{\algorithmicor}
\algnewcommand{\AND}{\algorithmicand}
\makeatletter
\newenvironment{breakablealgorithm}
  {
  \begin{center}
     \refstepcounter{algorithm}
     \hrule height.8pt depth0pt \kern2pt
     \renewcommand{\caption}[2][\relax]{
      {\raggedright\textbf{\ALG@name~\thealgorithm} ##2\par}%
      \ifx\relax##1\relax 
         \addcontentsline{loa}{algorithm}{\protect\numberline{\thealgorithm}##2}%
      \else 
         \addcontentsline{loa}{algorithm}{\protect\numberline{\thealgorithm}##1}%
      \fi
      \kern2pt\hrule\kern2pt
     }
  }{
     \kern2pt\hrule\relax
  \end{center}
  }

\begin{document}


\RUNAUTHOR{Simchi-Levi, Zheng and Zhu}

\RUNTITLE{Optimal Stochastic Bandit Policies with Light-Tailed Risk}


\TITLE{A Simple and Optimal Policy Design with Safety against Heavy-Tailed Risk for Stochastic Bandits}

\ARTICLEAUTHORS{%
\AUTHOR{David Simchi-Levi}
\AFF{Institute for Data, Systems, and Society, Department of Civil and Environmental Engineering, Operations Research Center, Massachusetts Institute of Technology, Cambridge, Massachusetts 02139, \EMAIL{dslevi@mit.edu}} 
\AUTHOR{Zeyu Zheng}
\AFF{Department of Industrial Engineering and Operations Research, University of California, Berkeley, CA 94720, \EMAIL{zyzheng@berkeley.edu}}
\AUTHOR{Feng Zhu}
\AFF{Institute for Data, Systems, and Society, Massachusetts Institute of Technology, MA 02139,  \EMAIL{fengzhu@mit.edu}}
} 

\ABSTRACT{%
We study the stochastic multi-armed bandit problem and design new policies that enjoy both worst-case optimality for expected regret and light-tailed risk for regret distribution. Specifically, our policy design (i) enjoys the worst-case optimality for the expected regret at order $O(\sqrt{KT\ln T})$ and (ii) has the worst-case tail probability of incurring a regret larger than any $x>0$ being upper bounded by $\exp(-\Omega(x/\sqrt{KT}))$, a rate that we prove to be best achievable with respect to $T$ for all worst-case optimal policies. Our proposed policy achieves a delicate balance between doing more exploration at the beginning of the time horizon and doing more exploitation when approaching the end, compared to standard confidence-bound-based policies. We also enhance the policy design to accommodate the ``any-time" setting where $T$ is unknown a priori, and prove equivalently desired policy performances as compared to the ``fixed-time" setting with known $T$. Numerical experiments are conducted to illustrate the theoretical findings. We find that from a managerial perspective, our new policy design yields better tail distributions and is preferable than celebrated policies especially when (i) there is a risk of under-estimating the volatility profile, or (ii) there is a challenge of tuning policy hyper-parameters. We conclude by extending our proposed policy design to the stochastic linear bandit setting that leads to both worst-case optimality in terms of expected regret and light-tailed risk on the regret distribution.

}%



\maketitle

%


\section{Introduction}

The stochastic multi-armed bandit (MAB) problem is a widely studied problem in the domain of sequential decision-making under uncertainty, with applications such as online advertising, recommendation systems, digital clinical trials, financial portfolio design, etc. In a standard MAB problem, whose formulation will be formally discussed later in Section 2, the decision maker sequentially chooses one of many not fully known arms (designs) in each of many time periods. The objective of decision maker is typically to maximize the expectation of the sum of rewards accumulated across all time period. The MAB problem and its associated policy design also provide notable theoretical insights exhibiting the exploration-exploitation trade-off, where the decision maker objectively needs to balance the exploration of arms whose reward distributions are relatively unknown and the exploitation of arms whose expected rewards are high and relatively more known. There is a vast of literature on MAB problem, and we refer to \cite{slivkins2019introduction} among others for a review.

For policy design and analysis, much of the MAB literature uses the metric of maximizing the \textit{expected} cumulative reward, or equivalently minimizing the \textit{expected regret} (where \textit{regret} refers to the difference between the cumulative reward obtained by always pulling the best arm and by executing a policy that does not a priori know the reward distributions). The optimality of a policy is often characterized through its expected regret's rate (order of dependence) on the experiment horizon $T$.

However, if an MAB policy design only focuses on optimizing the expected regret, the policy design may be exposed to risks that can arise in other aspects. As recently documented in \cite{fan2021fragility} about the standard Upper Confidence Bound (UCB) policy (\citealt{auer2002finite}), as well as will be extended in our work about the Successive Elimination (SE) policy (\citealt{even2006action}) and the Thompson Sampling (TS) policy (\citealt{russo2018tutorial}), these  renowned policies, despite enjoying optimality on expected regret, can incur a ``heavy-tailed risk''. That is, the distribution of the regret has a heavy tail --- the probability of incurring a linear regret slowly decays at a polynomial rate $\Omega(\poly(1/T))$ as $T$ tends to infinity. In contrast, a ``light-tailed'' risk in this MAB setting means that the probability of a policy incurring a linear regret decays at an exponential rate $\exp(-\Omega(T^\beta))$ for some $\beta > 0$. The heavy-tailed risk can be undesired when an MAB policy is used in applications that are sensitive to tail risks, including but not limited to finance, healthcare, supply chain, etc. In fact, understanding heavy-tailed risks and their associated disruptions in the aforementioned applications have been a keen focus in the operational research literature; see \cite{bouchaud1990anomalous, bouchaud2000theory, chopra2004supply, embrechts2013modelling, sodhi2021supply} for example. In the following, we give two concrete examples on the motivation and necessity of studying tail risks.
\begin{itemize}
    \item In the realm of new drug experimentation and pharmaceutical research, tail risks here pertain to the rare but potentially severe side effects or unintended consequences of a new drug. In clinical trials, while the primary goal is to learn and establish the (expected) efficacy of a drug, it is the unexpected low-probability events that can have profound implications for patients' health and thus affect the safety profile of the drug. In other words, while maximizing expected welfare is of primary importance, neglecting to properly account for tail events of incurring a large welfare loss can lead to misleading conclusions on a new drug and even significant legal and financial repercussions for drug manufacturers.

    \item In finance, particularly when dealing with noisy data, understanding tail risk is essential for sound decision-making and risk management. Noisy data can obscure genuine trends and relationships, leading to misleading interpretations and potentially erroneous decisions. If a researcher is interested in learning to maximize the expected profit through an ensemble of several trading strategies, while simultaneously controlling the tail probability of incurring a big money loss, then the researcher has to design an efficient policy robust to potentially misspecified (or underestimated) risk profile of the market. As we will show in this work from both the theoretical and numerical perspective, standard policies can be very sensitive to such misspecification, and we will also show whether and how we can control the tail risk under risk misspecifications.
\end{itemize}

Noting that the renowned policies may incur a heavy-tailed risk on the regret distribution, when achieving the optimality on the rate of expected regret, our work is primarily motivated by an attempt to answer the following question. \textbf{\emph{Is it possible to design a policy that on one hand enjoys the classical notion of optimality regarding expected regret, whereas on the other hand enjoys light-tailed risk for the regret distribution?}} If the answer is yes, then that policy design would enjoy both optimality and safety against heavy-tailed risk. Motivated by this question, we summarize our contributions in Section \ref{ssec:contribution}.

To facilitate describing the results on regret orders and function orders, we adopt the family of Bachmann-Landau notation. That is, we use $O(\cdot)$ ($\tilde O(\cdot)$) and $\Omega(\cdot)$ ($\tilde\Omega(\cdot)$) to present upper and lower bounds on the growth rate up to constant (logarithmic) factors, respectively, and $\Theta(\cdot)$ ($\tilde\Theta(\cdot)$) to characterize the rate when the upper and lower bounds match up to constant (logarithmic) factors. We use $o(\cdot)$ and $\omega(\cdot)$ to present strictly dominating upper bounds and strictly dominated lower bounds, respectively.

\subsection{Our Contributions}

\label{ssec:contribution}

\begin{enumerate}
    \item We first argue that instance-dependent consistency (formal definition to be discussed in Section \ref{sec:setup}) and light-tailed risk are incompatible. Recently, \cite{fan2021fragility} showed that information-theoretically optimized bandit policies as well as general UCB policies suffer from severely heavy-tailed risk. Built upon their analysis and results, we find that the general class of instance-dependent consistent policies cannot avoid heavy-tailed risk: if an instance-dependent consistent policy has the probability of incurring a linear regret decay as $\exp(-f(T))$, then $f(T)$ must be $o(T^\beta)$ as $T\to+\infty$ for any $\beta > 0$. Moreover, any policy that has instance-dependent $O(\ln T)$ expected regret, including the standard UCB and SE policy, and the TS policy, incurs (i) a linear regret with probability $\Omega(\poly(1/T))$ and (ii) an expected regret that is almost linear in $T$ if the risk parameter is severely misspecified. The implication is that if we want to find a policy design that avoids the heavy-tailed risk on regret distribution, we shall explore policies that are different from policies that have instance-dependent $O(\ln T)$ expected regret.
    
    \item We show that worst-case optimality and light-tailed risk can co-exist for policy design. Starting from the two-armed bandit setting, we provide a new policy design and prove that it enjoys both the worst-case optimality $\tilde O(\sqrt T)$ for the expected regret and the light-tailed risk $\exp(-\Omega(\sqrt{T}))$ for the regret distribution. We also prove that such exponential decaying rate of the tail probability is the best achievable within the class of worst-case optimal policies, as a ``lower bound" result. A careful change-of-measure argument is applied to show the optimality of our policy design with respect to the tail risk in the worst-case sense. Our policy design is surprisingly simple: it builds upon the idea of confidence bounds, and constructs different bonus terms compared to the standard ones to ensure safety against heavy-tailed risk and meanwhile maintain the worst-case optimality for expected performance.
    
    \item We extend our results from the two-armed bandit to the general $K$-armed bandit and characterize the tail probability bound for any regret threshold in an explicit form and through non-asymptotic lens. By further improving our policy design, we show that the worst-case probability of incurring a regret larger than $x$ is bounded approximately by $\exp(-\Omega(x/\sqrt{KT}))$ for \textit{any} $x>0$. We then enhance the policy design to accommodate the ``any-time" setting where $T$ is not known a priori, as a more challenging setting compared to the ``fixed-time" setting where $T$ is known a priori. We design a policy for the ``any-time" setting and prove that the policy enjoys an equivalently desired exponential decaying tail and optimal expected regret as in the ``fixed-time" setting. As a result, our policy does not require the knowledge of the risk parameter $\sigma$ or the time horizon $T$, freeing the concern of mis-specifications for these parameters; any tuning parameter is sufficient to achieve the optimal expected regret bound $\tilde O(\sqrt{KT})$ as well as tail risk. Despite the simplicity of our proposed policy design, the associated proof techniques are novel and may be useful for broader analysis on regret distribution and tail risk. In particular, a novel \textit{split-and-conquer} technique is introduced to achieve optimal dependence on both $T$ and $K$ under the case when $T$ is known, which is then further generalized to handle the any-time case without knowing $T$. Our result also partially answers an open question raised in \cite{lattimore2020bandit} for the stochastic MAB problem. A brief account of experiments are conducted to illustrate our theoretical findings. From a managerial perspective, compared to celebrated policies, our new policy designs yield better tail distributions and perform significantly better than celebrated policies especially when (i) the volatility of the environment is underestimated, or (ii) the hyper-parameter of the policy is hard to tune.
    
    \item We extend the idea of our policy design to the linear bandit setting, a setting that sits in the broad stochastic bandits problems and deviates from the MAB setting. In the linear bandit setting, the decision maker chooses an action in each period from a potentially time-varying continuous action set, instead of from $K$ discrete arms (see, e.g., \citealt{abbasi2011improved} for reference). We introduce a novel regret decomposition formula tailored to deal with the uncontrollable arm feature vectors as well as further adapt the split-and-conquer technique. We prove that our simple policy design can be integrated into classical linear bandit algorithms and lead to both worst-case optimality in terms of expected regret and light-tailed risk on the regret distribution.
\end{enumerate}

\subsection{Related Work}
\label{ssec:literature}

Our work builds upon the vast literature of designing and analyzing policies for the stochastic MAB problem and its various extensions. Comprehensive reviews can be found in \cite{bubeck2012regret, russo2018tutorial, slivkins2019introduction, lattimore2020bandit}. A standard paradigm for obtaining a near-optimal regret is to first \textit{fix} some confidence parameter $\delta>0$. Then a ``good event'' is defined such that good properties are retained conditioned on the event (for example, in the stochastic MAB problem, the good event is such that the mean of each arm always lies in the confidence bound). Then one can obtain both high-probability and worst-case expected regret bounds through careful analysis on the good event. It is known that the stochastic MAB problem has the following regret bound: for any fixed $\delta\in(0, 1)$, the regret bound of UCB is bounded by $O(\sqrt{KT\ln(T/\delta)})$ with probability at least $1-\delta$. Or equivalently speaking, the probability of incurring a $\Omega(\sqrt{KT\ln(T/\delta)})$ regret is bounded by $\delta$. However, the parameter $\delta$ must be an input parameter for the policy. We will discuss this issue in more details in Section \ref{sec:main}. In Section 17.1 of \cite{lattimore2020bandit}, an open question is asked: is it possible to design a \textit{single} policy such that the worst-case probability of incurring a $\Omega(\sqrt{KT}\ln(1/\delta))$ regret is bounded by $\delta$ for \textit{any} $\delta>0$ and any $K$-armed bandit problem with 1-subGaussian stochastic rewards? We partially answer this question by designing a policy such that for \textit{any} $\delta>0$, the probability of incurring a
\[
    \Omega\left(\sqrt{KT}\frac{\ln(T/\delta)}{\sqrt{\ln T}}\right)
\]
regret is bounded by $\delta$. We note that there has been a related result in the adversarial bandit setting (see, e.g., \citealt{neu2015explore}, \citealt{lattimore2020bandit}). It is shown that for the $K$-armed bandit problem with adversarial rewards uniformly in $[0, 1]$, there exists a single policy EXP3-IX such that the worst-case probability of incurring a
\begin{align*}
    \Omega\left(\sqrt{KT}\frac{\ln(K/\delta)}{\sqrt{\ln K}}\right)
\end{align*}
regret is bounded by $\delta$ for any $\delta > 0$. The difference between this result and ours are two-folds. From the policy design prospective, the idea behind EXP3-IX is to use exponential weight, while the idea behind our policy is to use a modified confidence bound designed to handle the stochastic setting. From the model setting perspective, in the adversarial setting, rewards are assumed to be uniformly bounded, and the bound is an input to the policy. While in the stochastic setting, the magnitude of a single reward is uncontrollable. In Section \ref{sec:extensions}, we elaborate in more details on why adversarial bandit policies such as EXP3-IX may not work.

There has been not much work on understanding the tail risk of stochastic bandit algorithms. Two earlier works are \cite{audibert2009exploration, salomon2011deviations} and they studied the concentration properties of the regret around the instance-dependent mean $O(\ln T)$. They showed that in general the regret of the policies concentrate only at a polynomial rate. That is, the probability of incurring a regret of $c(\ln T)^p$ (with $c>0$ and $p>1$ fixed) is approximately polynomially decaying with $T$. Different from our work, the concentration in their work is under an instance-dependent environment, and so such polynomial rate might be different across different instances. Nevertheless, their results indicate that standard bandit algorithms generally have undesirable concentration properties. Recently, \cite{ashutosh2021bandit} showed that an online learning policy with the goal of obtaining logarithmic regret can be fragile, in the sense that a mis-specified risk parameter (e.g., the parameter for subGaussian noises) in the policy can incur an instance-dependent expected regret of $\omega(\ln T)$. They then developed robust algorithms to circumvent the issue. Note that their goal is to handle mis-specification related with risk, but still the task is to minimize the expected regret. 

Our work is inspired by \cite{fan2021fragility}, who first provided a rigorous formulation to analyze heavy-tailed risk for bandit algorithms and showed that for any information-theoretically optimized bandit policy, the probability of incurring a linear regret is very heavy-tailed: at least $\Omega(1/T)$. They additionally showed that optimized UCB bandit designs are fragile to mis-specifications and they modified UCB algorithms to ensure a desired polynomial rate of tail risk, which makes the algorithms more robust to mis-specifications. By extending their analysis, we show a  more general incompatibility result. That is, a large family of policies --- all policies that are consistent --- suffer from heavy-tailed risk (see Section \ref{ssec:contribution}). Further, we propose a simple and new policy design that leads to both light-tailed risk (tail bound exponentially decaying with $\sqrt{T}$) and worst-case optimality (expected regret bounded by $\tilde O(\sqrt{T})$). We then show that our proposed simple policy design naturally extends to the general $K$-armed MAB setting and the linear bandit setting, and prove the compatibility between worst-case optimality and light-tailed risk. 

Recently, there is an increasing line of works analyzing the limiting behaviour of standard UCB and TS policies by considering the diffusion approximations (see, e.g., \citealt{araman2021diffusion, wager2021diffusion,fan2021diffusion, kalvit2021closer}). These works typically consider asymptotic limiting regimes that are set such that the gaps between arm means shrink with the total time horizon. We do not consider limiting regimes but instead consider the original problem setting with general parameters (e.g., gaps). We study how the tail probability decays with $T$ under original environments without taking the gaps to zero. 

Another line of works closely related with ours involve solving risk-averse formulations of the stochastic MAB problem (see, e.g., \citealt{sani2012risk, galichet2013exploration, maillard2013robust, zimin2014generalized, vakili2016risk, cassel2018general, tamkin2019distributionally, prashanth2020concentration, zhu2020thompson, baudry2021optimal, khajonchotpanya2021revised,chen2022bridging, chang2022unifying}). Compared to standard stochastic MAB problems, the main difference in their works is that arm optimality is defined using formulations other than the expected value, such as mean-variance criteria and (conditional) value-at-risk. These formulations consider some single metric that is different compared to the expected regret. From the formulation perspective, our work is different in the sense that we develop policies that on one hand maintain the classical worst-case optimal expected regret, whereas simultaneously achieve light-tailed risk bound. The policy design and analysis in our work are therefore also different from the literature and might be of independent interest. 

There has been a trend in the study of stochastic MAB problem within the non-stationary setting (see, e.g., \citealt{garivier2008upper, yu2009piecewise, besbes2014stochastic, luo2018efficient, cheung2019learning, chen2019new, faury2021regret, zhou2021regime, qin2022adaptivity, liu2022non}). However, most papers that study the non-stationary stochastic MAB problem (or its variant) has been focusing on analyzing the expectation of the cumulative reward (or regret). This work takes a first step towards understanding the regret tail risk under structured non-stationarity by showing that the new policy design and analysis could achieve light-tailed risk in the general linear bandit setting. We believe the results derived in this paper could have further implications for broader non-stationary MAB settings.

\subsection{Organization and Notation}

The rest of the paper is organized as follows. In Section \ref{sec:setup}, we discuss the setup and introduce the key concepts: light-tailed risk, instance-dependent consistency, worst-case optimality. In Section \ref{sec:main}, we show results on the incompatibility between light-tailed risk and consistency, and show the compatibility between light-tailed risk and worst-case optimality via a new policy design. In Section \ref{sec:extensions}, we consider the general $K$-armed bandit model and show a precise regret tail bound for our new policy design. We detail the proof road-map and how to further improve the design. In Section \ref{sec:experiments}, we present numerical experiments. In Section \ref{sec:linear}, we show how to extend our policy design into the general linear bandit setting and obtain similar light-tailed regret bound as in the MAB case. Finally, we conclude in Section \ref{sec:conclusion}. All detailed proofs are left to the online appendix.

Before proceeding, we introduce some notation.  For any $a, b\in\mathbb R$, $a\wedge b = \min\{a, b\}$ and $a\vee b = \max\{a, b\}$. For any $a\in\mathbb R$, $a_+ = \max\{a, 0\}$. We denote $[N]=\{1, \cdots, N\}$ for any positive integer $N$.

\section{The Setup} \label{sec:setup}

In this section, we first discuss the model setup. We then formally define three terms that appeared in the introduction and will appear in the rest of this work: light-tailed risk, instance-dependent consistency, and worst-case optimality.

Fix a time horizon of $T$ and the number of arms as $K$. Throughout the paper, we assume that $T\geq 3$, $K\geq 2$, and $T\geq K$. In each time $t\in[T]$, based on all the information prior to time $t$, the decision maker (DM) pulls an arm $a_t\in[K]$ and receives a reward $r_{t, a_t}$. More specifically, let $H_t = \{a_1, r_{1, a_1}, \cdots, a_{t-1}, r_{t-1, a_{t-1}}\}\cup\{T\}$ be the history prior to time $t$. When $t=1$, $H_1=\{T\}$. At time $t$, the DM is free to adopt any admissible policy $\pi_t: H_t\longmapsto a_t$ that maps the history $H_t$ to an action $a_t$ that may be realized from a discrete probability distribution on $[K]$. The environment then independently samples a reward $r_{t, a_t}=\theta_{a_t} + \epsilon_{t, a_t}$ and reveals it to the DM. Here, $\theta_{a_t}$ is the mean reward of arm $a_t$, and $\epsilon_{t, a_t}$ is an independent zero-mean noise term. We assume that $\epsilon_{t, a_t}$ is $\sigma$-subGaussian. That is,  there exists a $\sigma>0$ such that for any time $t$ and arm $k$,
\begin{align*}
    \max\left\{\mbP\left(\epsilon_{t, k} \geq x\right), \mbP\left(\epsilon_{t, k} \leq -x\right)\right\} \leq \exp(-x^2/(2\sigma^2)).
\end{align*}
Let $\theta = (\theta_1, \cdots, \theta_K)$ be the mean vector. Let $\theta_* = \max\{\theta_1, \cdots, \theta_K\}$ be the optimal mean reward among the $K$ arms. Note that DM does not know $\theta$ at the beginning. The empirical regret of the policy $\pi = (\pi_1, \cdots, \pi_T)$ under the mean vector $\theta$ and the noise parameter $\sigma$ over a time horizon of $T$ is defined as
\begin{align*}
    \hat R_{\theta, \sigma}^\pi(T) & = \theta_*\cdot T - \sum_{t=1}^T(\theta_{a_t} + \epsilon_{t, a_t}).
\end{align*}
Let $\Delta_k = \theta_* - \theta_k$ be the gap between the optimal arm and the $k$th arm. We assume $\Delta_k\leq 1\ (\forall k\in[N])$. Let $n_{t, k}$ be the number of times arm $k$ has been pulled up to time $t$. That is, $n_{t, k} = \sum_{s=1}^t\mathds 1\{a_s = k\}$. For simplicity, we will also use $n_{k}=n_{T, k}$ to denote the total number of times arm $k$ is pulled throughout the whole time horizon. We define $t_k(n)$ as the time period that arm $k$ is pulled for the $n$th time. Define the pseudo regret as 
\[
    R_{\theta, \sigma}^\pi(T) = \sum_{k=1}^K n_k\Delta_k
\]
and the genuine noise as
\[
    N^\pi(T) = \sum_{t=1}^T \epsilon_{t, a_t} = \sum_{k=1}^K\sum_{m=1}^{n_k}\epsilon_{t_k(m), k}.
\]
Then the empirical regret can also be written as $\hat R_{\theta, \sigma}^\pi(T) = R_{\theta, \sigma}^\pi(T) - N^\pi(T)$. We note that for most cases considered in this paper, the environment admits $\sigma$-subGaussian noises by default, and we will write $\hat R_\theta^\pi(T)$ instead of $\hat R_{\theta, \sigma}^\pi(T)$ and $R_\theta^\pi(T)$ instead of $R_{\theta, \sigma}^\pi(T)$ unless otherwise specified. 

The following simple lemma gives the mean and the tail probability of the genuine noise $N^\pi(T)$. Intuitively, it shows when bounding the mean or the tail probability of the empirical regret, one only needs to consider the pseudo regret term. We will make it more precise when we discuss the proof of main theorems.
\begin{lemma}\label{lemma:bound-noise}
We have $\mbE[N^\pi(T)] = 0$ and
\[
    \max\left\{\mbP\left(N^\pi(T) \geq x \right), \mbP\left(N^\pi(T) \leq -x\right)\right\} \leq \exp\left(\frac{-x^2}{2\sigma^2T}\right).
\]
\end{lemma}

Before proceeding to introduce the core concepts, we would like to give some remarks on the model assumptions.
\begin{enumerate}
    \item Prior knowledge. In the model above, we assume we have knowledge to the time horizon $T$ and the subGaussian parameter $\sigma$. We start from assuming knowledge of the two parameters to make our results and analysis cleaner and hopefully easier to comprehend. Later in Section \ref{sec:extensions}, we will show that neither of the two parameters need to be known in advance: we can design a policy without knowing $T$ or $\sigma$, while the policy still achieves the optimal expected regret bound and tail regret bound with regard to $T$. The knowledge of $T$ and $\sigma$ may only affect constant terms.
    
    \item SubGaussian noises. In this paper, we focus on subGaussian noises, which includes a broad class of light-tailed random noises including normal noises and bounded noises that are common in practice. Beyond subGaussian noises, there have been works contributed to understanding heavy-tailed bandit problems (see, e.g., \citealt{bubeck2013bandits, lattimore2017scale, yu2018pure, lugosi2019mean, lee2020optimal, agrawal2021regret, bhatt2022nearly, tao2022optimal}). The task is to develop near-optimal \textit{expected} regret bounds when random noises are heavy-tailed than the subGaussian distribution. While our model is not handling the most general class of random noises, our results and insights have the potential to remain effective for a broader class of problems.
\end{enumerate}

\subsection{Light-tailed Risk, Instance-dependent Consistency, Worst-case Optimality}

Now we describe concepts that are needed to formalize the policy design and analysis.

{\noindent\bf 1. Light-tailed risk.} A policy is called light-tailed, if for any constant $c > 0$, there exists some $\beta > 0$ and constant $C > 0$ such that
\begin{align*}
    \limsup_{T\to+\infty} \frac{\ln\left\{\sup_{\theta}\mbP\left(\hat R_{\theta}^\pi(T) > cT\right)\right\}}{T^{\beta}} \leq -C.
\end{align*}
Note that here, we allow $\beta$ and $C$ to be dependent on $c$. In brief, a policy has light-tailed risk if the probability of incurring a linear regret can be bounded by an exponential term of polynomial $T$:
\begin{align*}
    \sup_{\theta}\mathbb P(\hat R_\theta^\pi(T)\geq cT) = \exp(-\Omega(T^{\beta}))
\end{align*}
for some $\beta > 0$. If a policy is not light-tailed, we say it is \textit{heavy-tailed}.

We clarify that conventionally, a distribution is called ``light-tailed" when its moment generating function is finite around a neighborhood of zero. Our definition of ``light-tailed" emphasizes the boundary between heavy and light to separate polynomial rate of decaying versus exponential-polynomial rate of decaying, which is aligned with but technically different from the conventional definition of ``light-tailed". For example, for regret random variables $R(T)$ indexed by $T$, when $T$ is large, if $\mbP(R(T)>T/2) \sim T^{-\beta}$ for some positive $\beta$, then its distribution is heavy-tailed in both our definition and the conventional definition. If $\mbP(R(T)>T/2) \sim \exp(-T^\beta)$ for $\beta \in(0,1)$, then its distribution is light-tailed in our definition and is heavy-tailed in the conventional definition. If $\mbP(R(T)>T/2) \sim \exp(-T^\beta)$ for $\beta \ge 1$, then its distribution is light-tailed in both our definition and the conventional definition. Therefore, when we claim safety against heavy-tailed risk, it indicates tail distribution that is lighter than any polynomial rate of decay. 

We would also like to note that our criterion of light-tailed risk is relatively loose in the sense that it only involves the regret threshold being a linear $T$ term. This is chosen primarily for showing the undesirable regret tail behavior of many celebrated policies. For all the new policy designs in this paper, the corresponding tail upper bounds are built for any regret threshold.
    
{\noindent\bf 2. Instance-dependent consistency.} A policy is called consistent or instance-dependent consistent, if for any underlying true mean vector $\theta$, the policy has that
\begin{align*}
    \limsup_{T\to+\infty} \frac{\mbE\left[\hat R_\theta^\pi(T)\right]}{T^\beta} = 0
\end{align*}
holds for any $\beta > 0$. In brief, a policy is consistent if the expected regret can never be polynomially growing in $T$ for any fixed instance.
    
{\noindent\bf 3. Worst-case optimality.} A policy is said to be worst-case optimal, if for any $\beta > 0$, the policy has that
\begin{align*}
    \limsup_{T\to+\infty} \frac{\sup_{\theta}\mbE\left[\hat R_\theta^\pi(T)\right]}{T^{1/2+\beta}} = 0.
\end{align*}
In brief, a policy is worst-case optimal if the worst-case expected regret can never be growing in a polynomial rate faster than $T^{1/2}$. Note that here we adopt a relaxed definition of optimality, in the sense that we do not clarify how the regret scale with the number of arms $K$ compared to that in literature. The notion of worst-case optimality in this work focuses on the dependence on $T$. For example, a policy with worst-case regret $O(\poly(K)\sqrt{T}\cdot\poly(\ln T))$ is also optimal by our definition.

It is well known that for the stochastic MAB problem, one can design algorithms to achieve both instance-dependent consistency and worst-case optimality. Among them, two types of policies are of prominent interest: Successive Elimination (SE) and Upper Confidence Bound ($\UCB$). We list the algorithm paradigms in Algorithm \ref{alg:SE} and \ref{alg:UCB}. Let $\hat\theta_{t, k}$ be the empirical mean reward of arm $k$ up to time $t$. The bonus term $\rad(n)$ is typically set as
\begin{align} \label{rad:standard}
    \rad(n) = \sigma\sqrt{\frac{\eta\ln T}{n}}
\end{align}
with $\eta>0$ being some tuning parameter.

\bigskip

\begin{minipage}{.45\linewidth}
\begin{breakablealgorithm}
\caption{Successive Elimination}
\label{alg:SE}
\begin{algorithmic}[1]
\State $\cA = [K]$. $t\gets 0$. $n\gets 0$.
\While{$t < T$}
    \State Pull each arm in $\cA$ once. $t\gets t + |\cA|$. $n\gets n + 1$.
    \State Eliminate any $k\in\cA$ from $\cA$ if $\exists k'$:
    \begin{align*}
        \hat\theta_{t_{k'}(n), k'} - \rad(n_{t, k'}) > \hat\theta_{t_k(n), k} + \rad(n_{t, k})
    \end{align*}
\EndWhile
\end{algorithmic}
\end{breakablealgorithm}
\end{minipage}
\hspace{0.5cm}
\begin{minipage}{.45\linewidth}
\begin{breakablealgorithm}
\caption{Upper Confidence Bound}
\label{alg:UCB}
\begin{algorithmic}[1]
\State $\cA = [K]$. $t\gets 0$.
\While{$t < T$}
    \State $t\gets t+1$.
    \State Pull the arm with the highest $\UCB$ (randomly select one if a tie happens):
    \begin{align*}
        \arg\max_k\quad \hat\theta_{t-1, k} + \rad(n_{t-1, k}).
    \end{align*}
\EndWhile
\end{algorithmic}
\end{breakablealgorithm}
\end{minipage}
\bigskip

SE maintains an active action set, and for each arm in the action set, it maintains a confidence interval. In each phase $n$, after pulling each arm in the action set, SE updates the action set by eliminating any arm whose confidence interval is strictly dominated by others. As a comparison, UCB does not shrink the active action set: it always pulls the arm with the highest upper confidence bound. These two algorithms share similar structure, in the sense that they both utilize confidence intervals to guide the actions.

\section{The Basic Case: Two-armed Bandit}
\label{sec:main}

We start from the simple two-armed bandit setting. The general multi-armed setting is deferred to the next section. We first show that all policies that enjoy instance-dependent consistency are heavy-tailed in terms of regret distribution. The result reveals an incompatibility between instance-dependent consistency and light-tailed risk. Then we show how to add a simple twist to standard confidence bound based policies to obtain light-tailed risk. Moreover, we show that our design leads to an optimal tail decaying rate for all policies that enjoy worst-case optimal order of expected regret.

\subsection{Instance-dependent Consistency Causes Heavy-tailed Risk}

\begin{theorem} \label{thm:consistent-vs-light-tail}
If a policy $\pi$ is instance-dependent consistent, then it can never be light-tailed. Moreover, if $\pi$ satisfies
\begin{align} \label{eq:consistent-ln}
    \limsup_{T\to+\infty}\frac{\mbE\left[\hat R_\theta^\pi(T)\right]}{\ln T} < +\infty
\end{align}
for any $\theta$, then we have the following stronger argument. For any $c\in(0, 1/2)$, there exists $C_\pi>0$ such that
\begin{align}
    \liminf_{T\to+\infty} \frac{\ln\left\{\sup_{\theta}\mbP\left(\hat R_{\theta, \sigma_0}^\pi(T) > cT\right)\right\}}{\ln T} \geq -C_\pi\frac{\sigma^2}{\sigma_0^2} \label{eq:heavy-tail-ln}
\end{align}
for any $\sigma_0 \geq \sigma$.
\end{theorem}

Theorem \ref{thm:consistent-vs-light-tail} suggests that a consistent policy must have a risk tail heavier than an exponential one. The proof of Theorem \ref{thm:consistent-vs-light-tail} adapts a change-of-measure argument appeared in \cite{fan2021fragility}. Intuitively speaking, if we want a policy to be adaptive enough to handle different instances, then the cost we have to pay is heavy-tailed risk. Moreover, if the policy achieves $O(\ln T)$ regret for any fixed instance $\theta$ (the constant is typically dependent on $\theta$), then the probability of incurring a linear regret becomes $\exp(-O(\ln T)) = \Omega(\poly(1/T))$. To make things worse, if such a policy that achieves instance-dependent $O(\ln T)$ regret under $\sigma$-subGaussian noises is used in an environment where the true risk parameter $\sigma_0$ is much larger than $\sigma$, then the probability of incurring a linear regret becomes $\Omega(1/T^\varepsilon)$, where $\varepsilon>0$ can be arbitrarily close to $0$ as $\sigma_0$ increases. As a result, the worst-case expected regret scales almost linearly in $T$. 

Our argument resonates with Theorem 1 in \cite{ashutosh2021bandit} and Corollary 2 in \cite{fan2021fragility}. \cite{ashutosh2021bandit} showed that a policy cannot achieve $O(\ln T)$ regret if the policy is always consistent for \textit{all} environments regardless of the value of the risk parameter $\sigma$, and \cite{fan2021fragility} showed that $\pi=\UCB$ optimized for i.i.d Gaussian rewards with variance $\sigma^2$ satisfies \eqref{eq:heavy-tail-ln} for $C_\pi=1$. We note that policies that achieve $O(\ln T)$ regret are special cases within the family of instance-dependent consistent policies: a policy with $O((\ln T)^{100})$ regret for any fixed $\theta$ remains consistent. Further, information-theoretically optimized policies are special cases of policies that achieve $O(\ln T)$ regret: they try to further optimize the constant term in front of the $\ln T$ term (or the constant in \eqref{eq:consistent-ln}) according to different noise distributions (see, e.g., \citealt{garivier2011kl}). In fact, $\UCB$ or $\SE$ with a large $\eta$ retains instance-dependent consistency but may fail to be information-theoretically optimal. Therefore, our arguments generalize beyond optimized policies in \cite{fan2021fragility} and show the intrinsic incompatibility between instance-dependent consistency and light-tailed risk.

Many standard policies are known to achieve instance-dependent $O(\ln T)$ regret. One special case is the family of confidence bound related policies ($\SE$ and $\UCB$). From Theorem \ref{thm:consistent-vs-light-tail}, the standard bonus term \eqref{rad:standard} will always lead to a tail polynomially dependent on $T$. Another example is the Thompson Sampling ($\ThS$) policy. It has been established that $\pi=\ThS$ with Beta or Gaussian priors has the property \eqref{eq:consistent-ln} (see, e.g., Theorem 1 and 2 in \citealt{agrawal2012analysis}, proof of Theorem 1.3 in \citealt{agrawal2017near}). Theorem \ref{thm:consistent-vs-light-tail} then suggests that \eqref{eq:heavy-tail-ln} also holds for $\pi=\ThS$.

We need to remark on the difference between Theorem \ref{thm:consistent-vs-light-tail} and high-probability bounds in the stochastic MAB literature. It has been well-established that $\UCB$ or $\SE$ with
\begin{align*}
    \rad(n) = \sigma\sqrt{\frac{\eta\ln(1/\delta)}{n}}
\end{align*}
achieves $\tilde O(\sqrt{T\cdot\polylog (T/\delta)})$ regret with probability at least $1-\delta$ (see, e.g., Section 1.3 in \cite{slivkins2019introduction}, Section 7.1 in \cite{lattimore2020bandit}). Such design also leads to a consistent policy. However, the bound holds only for fixed $\delta$. In fact, one can see that the bonus design is dependent on the confidence parameter $\delta$. If $\delta = \exp(-\Omega(T^\beta))$ with $\beta > 0$, then the scaling speed of the regret with respect to $T$ can only be greater than $1/2$, which is sub-optimal. As a comparison, in our problem, ideally we seek to find a single policy such that it achieves $\tilde O(\sqrt{T\cdot\polylog(T/\delta)})$ regret for \textit{any} $\delta>0$.

Up till now, we have made two observations. First, from standard stochastic MAB results, consistency and optimality can hold simultaneously. Second, from Theorem \ref{thm:consistent-vs-light-tail}, consistency and light-tailed risk are always incompatible. Then a natural question arises: Can we design a policy that enjoys both optimality and light-tailed risk? If we can, then can we make the tail risk decay with $T$ at an optimal rate? We answer these two questions with an affirmative ``yes'' in the next section.

\subsection{Worst-case Optimality Permits Light-tailed Risk}

In this section, we propose a new policy design that achieves both light-tailed risk and worst-case optimality. The design is very simple. We still use the idea of confidence bounds, but instead of setting the bonus as \eqref{rad:standard}, we set
\begin{align} \label{rad:new}
    \rad(n) = \sigma\frac{\sqrt{\eta T\ln T}}{n}
\end{align}
with $\eta > 0$ being a tuning parameter. Theorem \ref{thm:exp-2-armed} gives performance guarantees for the mean and the tail probability of the empirical regret when $\pi=\SE$.

\begin{theorem} \label{thm:exp-2-armed}
For the two-armed bandit problem, the SE policy with $\eta\geq 4$ and the bonus term being \eqref{rad:new}
satisfies the following two properties.
\begin{enumerate}
    \item $\sup_{\theta}\mbE[\hat R_\theta^\pi(T)] = O(\sqrt{T\ln T})$.
    \item For any $c > 0$ and any $\alpha\in(1/2, 1]$, we have
    \begin{align*}
        \sup_{\theta}\mathbb P(\hat R_\theta^\pi(T)\geq cT^\alpha) = \exp(-\Omega(T^{\alpha-1/2})).
    \end{align*}
\end{enumerate}
\end{theorem}

The first item in Theorem \ref{thm:exp-2-armed} means that with the modified bonus term, the worst case regret is still bounded by $O(\sqrt{T\ln T})$, which is the same as the regret bounds for $\SE$ and $\UCB$ with the standard bonus term \eqref{rad:standard}. The second item shows that the tail probability of incurring a $\Omega(T^\alpha)$ regret ($\alpha > 1/2$) is exponentially decaying in $\Omega(T^{\alpha-1/2})$, and thus the policy is light-tailed. The detailed proof of Theorem \ref{thm:exp-2-armed} is provided in the online appendix. The illustrative road-map of the proof is delegated to Section \ref{sec:extensions}, where we provide the proof idea for Theorem \ref{thm:exp-K-armed} that is a strict generalization of Theorem \ref{thm:exp-2-armed}. In Theorem \ref{thm:exp-K-armed}, we also demonstrate that the new bonus design \eqref{rad:new} allows robustness and gives exponential decaying tail risk even under an misspecified deviation parameter, avoiding \eqref{eq:heavy-tail-ln}. Here, we give some intuition on the new bonus design. Our new bonus term inflates the standard one by a factor of $\sqrt{T/n}$. This means our policy is more conservative than the traditional confidence bound methods, especially at the beginning. In fact, one can observe that for the first $\Theta(\sqrt{T})$ time periods, our policy consistently explores between arm $1$ and $2$, regardless of the environment. A naturally corollary is that our policy is never ``consistent'', following Theorem \ref{thm:consistent-vs-light-tail}. However, the bonus term \eqref{rad:new} decays at a faster rate on the number of pulling times $n$ compared to \eqref{rad:standard}. This means as the experiment goes on, the policy leans towards exploitation. We note that this is not the same as the explore-then-commit policy, which is well-known to achieve a sub-optimal $\Theta(T^{2/3})$ order of expected regret.

The following theorem shows that the risk tail in Theorem \ref{thm:exp-2-armed} is not improvable in term of order on $T$. That is, if the policy $\pi$ is worst-case optimal, then for \textit{any} $\alpha\in(1/2, 1]$, the exponent of $\alpha-1/2$ is tight.

\begin{theorem} \label{thm:tightness}
Let $c\in (0, 1/2)$. Consider the 2-armed bandit problem with Gaussian noise. Let $\pi$ be a worst-case optimal policy. That is, for any $\alpha > 1/2$, 
\begin{align*}
    \limsup_T \frac{\sup_{\theta}\mbE[\hat R_\theta^\pi(T)]}{T^\alpha} = 0.
\end{align*}
Then for any $\alpha\in(1/2, 1]$,
\begin{align*}
    \liminf_T\frac{\ln\left\{\sup_{\theta}\mbP(\hat R_\theta^\pi(T)\geq cT^{\alpha})\right\}}{T^{\beta}} = 0
\end{align*}
holds for any $\beta > \alpha-1/2$.
\end{theorem}

Theorem \ref{thm:tightness} also relies on the change-of-measure argument appeared in the proof of Theorem \ref{thm:consistent-vs-light-tail}. However, there are two notable differences: we only have worst-case optimality rather than consistency, and the regret threshold $cT^\alpha$ is in general not linear in $T$. Therefore, we need to take care of constructing the specific ``hard'' instance when doing the change-of-measure. The detailed proof is delegated to the online appendix.

\section{The General Case: Multi-armed Bandit}
\label{sec:extensions}

In this section, we provide step-by-step extensions to our previous results in Section \ref{sec:main} to the general multi-armed bandit setting. We first give a direct extension where the bonus term is set as \eqref{rad:new}. It turns out that such bonus design only yields a $\tilde O(K\sqrt{T})$ expected regret, which has a linear dependence on $K$, and so we study how to achieve the optimal dependence on both $K$ and $T$ by slightly modifying the design. Finally, we relax the assumption of knowing $T$ a priori and give an any-time policy that enjoys an equivalent tail probability bound as compared to the fixed-time case.

\subsection{The Direct Extension}

We first provide a generalization of our previous tail probability bound in Theorem \ref{thm:exp-2-armed} from the following aspects: (a) a general $K$-armed bandit setting; (b) an analysis for $\UCB$ aside from $\SE$; (c) a detailed characterization of the tail bound for any fixed regret threshold.

\begin{theorem} \label{thm:exp-K-armed}
For the $K$-armed bandit problem, both the policy $\pi=\SE$ and the policy $\pi=\UCB$ with
\begin{align*}
    \rad(n) = \sigma\frac{\sqrt{\eta T\ln T}}{n}
\end{align*}
satisfy the following two properties.
\begin{enumerate}
    \item If $\eta \geq 4$, then $\sup_{\theta}\mbE\left[\hat R_\theta^{\pi}(T)\right] \leq 4K + 4K\sigma\sqrt{\eta T\ln T}$.
    \item If $\eta > 0$, then for any $x > 0$, we have
    \begin{align*}
        & \quad\sup_{\theta}\mbP(\hat R_\theta^\pi(T)\geq x) \\
        & \leq \exp\left(-\frac{x^2}{2K\sigma^2T}\right) + 2K\exp\left(-\frac{(x-2K-4K\sigma\sqrt{\eta T\ln T})_+^2}{32\sigma^2K^2T}\right) + K^2T\exp\left(-\frac{x\sqrt{\eta\ln T}}{8\sigma K\sqrt{T}}\right).
    \end{align*}
\end{enumerate}
\end{theorem}

{\noindent\bf Proof Idea.} We provide a road-map of proving Theorem \ref{thm:exp-K-armed}. The expected regret bound is proved using standard techniques. That is, we define ``the good event" to be such that the mean of each arm always lies in the confidence bounds throughout the whole time horizon. Conditioned on the good event, the regret of each arm is bounded by $O(\sqrt{T\ln T})$, and thus the total expected regret is $O(K\sqrt{T\ln T})$.

The proof of the tail bound requires new analysis different from that in the literature. Without loss of generality, we assume arm $1$ is optimal. We first illustrate the proof for $\pi=\SE$. 
\begin{enumerate}
\item We use
\begin{align*}
    \sup_{\theta}\mbP(\hat R_\theta^\pi(T)\geq x) \leq \mbP\left(N^\pi(T) \leq -x/\sqrt{K}\right) + \sup_{\theta}\mbP\left(R_\theta^\pi(T) \geq x(1-1/\sqrt{K})\right)
\end{align*}
The term with the genuine noise can be easily bounded using Lemma \ref{lemma:bound-noise}. We are left to bound the tail of the pseudo regret. By a union bound, we observe that
\begin{align}
    \mbP\left(R_\theta^\pi(T) \geq x(1-1/\sqrt{K})\right) \leq \sum_{k\neq 1} \mbP\left(n_k\Delta_k \geq x/(K+\sqrt{K})\right) \leq \sum_{k\neq 1} \mbP\left(n_k\Delta_k \geq x/(2K)\right) \label{eq:split}
\end{align}
Thus, we reduce bounding the sum of the regret incurred by different arms to bounding that by a single sub-optimal arm.

\item For any $k\neq 1$, we define
\begin{align*}
    S_k = \{\text{Arm }1\text{ is not eliminated before arm }k\}.
\end{align*}
With a slight abuse of notation, we let $n_0 = \lceil x/(2K\Delta_k)\rceil-1$. Consider the case when both $n_k\Delta_k\geq x/(2K)$ and $S_k$ happen. This corresponds to the risk of \textit{spending too much time before correctly discarding a sub-optimal arm}. Then arm $1$ and $k$ are both not eliminated after each of them being pulled $n_0$ times. This indicates
\begin{align*}
    \hat\theta_{t_1(n_0), 1} - \frac{\sigma\sqrt{\eta T\ln T}}{n_0}\leq \hat\theta_{t_k(n_0), k} + \frac{\sigma\sqrt{\eta T\ln T}}{n_0}
\end{align*}
The probability of this event can be bounded using concentration of subGaussian variables, which yields the second term in the tail probability bound in Theorem \ref{thm:exp-K-armed}. We note that the choice of $n_0$ is important. Also, at this step, even if we replace our new bonus term by the standard one, the bound still holds.

\item Now consider the situation when both $n_k\Delta_k\geq x/(2K)$ and $\bar S_k$ (complement of $S_k$) happen. This corresponds to the risk of \textit{wrongly discarding the optimal arm}. Then after some phase $n$, the optimal arm $1$ is eliminated by some arm $k'$, while arm $k$ is not eliminated. Note that $k=k'$ does not necessarily hold when $K > 2$. As a consequence, we have the following two events hold simultaneously:
\begin{align*}
    \hat\theta_{t_{k'}(n), k'} - \frac{\sigma\sqrt{\eta T\ln T}}{n} \geq \hat\theta_{t_1(n), 1} + \frac{\sigma\sqrt{\eta T\ln T}}{n}\quad \text{ and } \quad
    \hat\theta_{t_k(n), k} + \frac{\sigma\sqrt{\eta T\ln T}}{n} \geq \hat\theta_{t_1(n), 1} + \frac{\sigma\sqrt{\eta T\ln T}}{n}.
\end{align*}
We note that the second event holds because otherwise, combined with the first event, we have
\begin{align*}
    \hat\theta_{t_{k'}(n), k'} - \frac{\sigma\sqrt{\eta T\ln T}}{n} \geq \hat\theta_{t_1(n), 1} + \frac{\sigma\sqrt{\eta T\ln T}}{n} > 
    \hat\theta_{t_k(n), k} + \frac{\sigma\sqrt{\eta T\ln T}}{n}.
\end{align*}
This contradicts with the fact that arm $k$ is not eliminated.

Now, the first event leads to
\begin{align*}
    \text{Mean of some noise terms } \geq 2\cdot\rad(n) + \Delta_{k'} \geq 2\cdot\rad(n) = \frac{2\sigma\sqrt{\eta T\ln T}}{n}.
\end{align*}
The second event leads to
\begin{align*}
    \text{Mean of some noise terms } \geq \Delta_k \geq \frac{x}{2KT}.
\end{align*}
To deal with an arbitrary $n$, we bound the probabilities of the two events separately and take the minimum of the two probabilities ($\mbP(A\cap B)\leq\min\{\mbP(A), \mbP(B)\}\ (\forall A, B)$). Then such minimum can be further bounded using the basic inequality $\min\{a, b\}\leq \sqrt{ab}\ (\forall a, b\geq 0)$. Roughly speaking, the probability can be bounded by
\begin{align*}
    & \quad \exp\left(-\Theta\left(n\cdot\rad(n)^2\right)\right) \wedge \exp\left(-\Theta\left(n\frac{x^2}{K^2T^2}\right)\right) \\
    & \precsim \exp\left(-\Theta\left(n\frac{\rad(n)\cdot x}{KT}\right)\right) \asymp \exp\left(-\Theta\left(\frac{x}{K\sqrt{T}}\right)\right),
\end{align*}
which yields the last term in Theorem \ref{thm:exp-K-armed}. We note that at this step, the $\sqrt{T}/n$ design in our new bonus term plays a crucial role. The standard bonus term \eqref{rad:standard} does not suffice to get an exponential bound since
\begin{align*}
    \exp\left(-\Theta\left(n\cdot\rad(n)^2\right)\right) \wedge \exp\left(-\Theta\left(n\frac{x^2}{K^2T^2}\right)\right) \precsim \exp\left(-\Theta(\ln T)\right) = \polylog(1/T).
\end{align*}
In fact, such a tail upper bound coincides with our tail lower bound in Theorem \ref{thm:consistent-vs-light-tail} and highlights \textit{why and when} standard policies fail: they fail because they are too adaptive to different environments, leading to greedy-like behavior after $O(\ln T)$ rounds, and they fail when they wrongly discard the optimal arm at the beginning several rounds. Our bonus design inflates the standard one at the beginning when $n$ is small, but decays faster when $n$ grows. It turns out that such inflation does not sacrifice the order of the worst-case expected regret with respect to $T$, while at the same time optimally control the worst-case regret tail. Rather, the hidden cost is consistency --- we are abandoning instance-dependent consistency in exchange for light-tailed risk.
\end{enumerate}

We next illustrate the proof for $\pi=\UCB$. Unlike the proof for $\pi=\SE$, the difference here stems from the fact that $S_k$ is no longer valid. Following the first step in the proof for $\pi=\SE$, we proceed as follows. For fixed $k$, we let $n_0 = \lceil x/(2K\Delta_k)\rceil - 1$. When arm $k$ is pulled for the $(n_0+1)$th time, by the design of the $\UCB$ policy, there exists some $n$ such that
\begin{align*}
    \theta_1 + \frac{\sum_{m=1}^{n}\epsilon_{t_1(m), 1} + \sigma\sqrt{\eta T\ln T}}{n}\leq \theta_k + \frac{\sum_{m=1}^{n_0}\epsilon_{t_k(m), k} + \sigma\sqrt{\eta T\ln T}}{n_0}.
\end{align*}
An important observation is that the event above is covered by a union of two events as follows:
\begin{align*}
    \frac{\sum_{m=1}^{n_0}\epsilon_{t_k(m), k} + \sigma\sqrt{\eta T\ln T}}{n_0} \geq \frac{\Delta_k}{2} \quad \text{ and } \quad \exists n:\ \frac{\sum_{m=1}^{n}\epsilon_{t_1(m), 1} + \sigma\sqrt{\eta T\ln T}}{n} \leq - \frac{\Delta_k}{2}.
\end{align*}
The probability of each of the two events can be bounded using similar techniques adopted for $\pi=\SE$. In fact, it is implicitly shown in our proof that $\UCB$ can yield better constants than $\SE$. We still need to emphasize that when bounding the second event, similar to the argument for $\pi=\SE$, the choice of our new bonus term is crucial.

\smallskip

{\noindent\bf Remarks.} Some remarks for Theorem \ref{thm:exp-K-armed} are as follows.

\begin{enumerate}
    \item For the regret bound, we note that compared to the optimal $\tilde\Theta(\sqrt{KT})$ bound, we have an additional $\sqrt{K}$ term. We should point out that the additional $\sqrt{K}$ term is not surprising under the bonus term \eqref{rad:new}. An intuitive explanation is as follows. Compared to the bonus term \eqref{rad:standard}, we widen the bonus term by a factor of $\sqrt{T/n}$. Among the $K-1$ arms, there must exist an arm such that it is pulled no more than $T/K$ times throughout the whole time horizon. That is, the bonus term of this arm is always inflated by a factor of at least $\sqrt{K}$. The standard regret bound analysis will, as a result, lead to an additional $\sqrt{K}$ factor compared to the optimal regret bound $\tilde\Theta(\sqrt{KT})$. 
    \item From the proof road-map, one can see that the tail bound in Theorem \ref{thm:exp-K-armed} is also valid for the pseudo regret $\sup_{\theta}\mbP(R_\theta^\pi(T)\geq x)$. To simplify the tail bound, one can notice that the last term in the bound can be written as
    \begin{align*}
        K\exp\left(-\frac{x\sqrt{\eta\ln T} - 8\sigma K\sqrt T\ln (KT)}{8\sigma K\sqrt T}\right) \leq K\exp\left(-\frac{(x - 16K\sigma\sqrt{1/\eta\cdot T\ln T})\sqrt{\eta\ln T}}{8\sigma K\sqrt T}\right)
    \end{align*}
    Since the tail probability has a trivial upper bound of $1$, the last term can be replaced by
    \begin{align*}
        K\exp\left(-\frac{(x - 16K\sigma\sqrt{1/\eta\cdot T\ln T})_+\sqrt{\eta\ln T}}{8\sigma K\sqrt T}\right).
    \end{align*}
    Therefore, if we let
    \begin{align*}
        y = \frac{\left(x-2K - 16\sigma K\sqrt{(\eta\vee1/\eta) T\ln T}\right)_+}{8\sigma K\sqrt{T}},
    \end{align*}
    then for any $x\geq 0$, we have
    \begin{align*}
    & \quad\sup_{\theta}\mbP(R_\theta^\pi(T)\geq x) \\
    & \leq \exp\left(-y^2\right) + K\exp\left(-y^2\right) + K\exp\left(-y\sqrt{\eta\ln T}\right) \leq 4K\exp\left(-(y^2\wedge y\sqrt{\eta\ln T})\right).
\end{align*}
    One can observe that for any $\eta > 0$, our policy always yields a $\tilde O(\sqrt{T})$ expected regret (although with a constant larger than that in the first result in Theorem \ref{thm:exp-K-armed}). In fact, notice that for any $x > 0$
    \begin{align*}
        \mbE[\hat R_\theta^\pi(T)] = \mbE[R_\theta^\pi(T)] \leq x + \mbP(R_\theta^\pi(T)\geq x)\cdot T.
    \end{align*}
    If we let $x = 2K + C\sigma K\sqrt{(\eta\vee1/\eta)T\ln T}$ with the absolute constant $C$ being moderately large, then $\mbP(R_\theta^\pi(T)\geq x)\cdot T = O(1)$. As a result, the worst-case regret becomes
    \begin{align*}
        O\left(K\sigma\sqrt{(\eta\vee1/\eta)T\ln T}\right).
    \end{align*}
    This observation has two implications.
    \begin{itemize}
        \item[(a)] First, our policy design is not sensitive to the parameter $\eta$ regarding the growth rate on $T$, as opposed to the standard $\UCB$ policy with \eqref{rad:standard}, where a very small $\eta$ can possibly make the $\UCB$ policy no longer enjoy a $\tilde O(\sqrt{T})$ worst-case regret. For completeness, we summarize this point with a proof in the online appendix. In fact, we show that when $\eta$ is very small, the regret for either $\SE$ or $\UCB$ is lower bounded by $\tilde\Omega(T^{1-2\eta})$, the order of which can be arbitrarily close to $1$.
        \item[(b)] Second, our policy design is not sensitive to the risk parameter $\sigma$ regarding the growth rate on $T$. Moreover, our policy does not even require the knowledge of $\sigma$ in advance! To be more concrete, if our policy uses a (severely) misspecified risk parameter $\sigma' \neq \sigma$, then note that
        \begin{align*}
            \sigma'\sqrt{\eta} = \sigma\sqrt{\eta\frac{\sigma'^2}{\sigma^2}} \triangleq \sigma\sqrt{\eta'}, 
        \end{align*}
        we can treat our policy as if we are using the true risk parameter $\sigma$ but with a scaled tuning parameter $\eta'>0$. The tail probability of incurring a linear regret still decays at an $\exp(-\Omega(\sqrt{T}))$ rate, and meanwhile, the expected regret still grows at a $\sqrt{T}$ rate. This is in contrast with standard $\UCB$ or $\ThS$ policies, where a misspecified risk parameter smaller than the true one will, by Theorem \ref{thm:consistent-vs-light-tail}, possibly make the expected regret scale at an order larger than $1/2$ (see also, e.g., Corollary 2 in \citealt{fan2021fragility}). 
    \end{itemize}
    
    We note that the two implications above also hold for refined policies discussed in the following sections.
    
\end{enumerate}

\subsection{Optimal Expected Regret}

A natural question is whether we can improve the regret bound in Theorem \ref{thm:exp-K-armed} to $\tilde\Theta(\sqrt{KT})$ and get a probability bound of
\begin{align*}
    \ln\left\{\sup_{\theta}\mbP(\hat R_\theta^\pi(T)\geq x)\right\} = -\Omega\left(\frac{x}{\sqrt{KT}}\right)
\end{align*}
for large $x$. By slightly modifying the bonus term \eqref{rad:new}, we give a ``yes'' answer to this question.

\begin{theorem} \label{thm:exp-K-armed-optimal}
For the $K$-armed bandit problem, both the policy $\pi=\SE$ and the policy $\pi=\UCB$ with
\begin{align} \label{rad:optimal}
    \rad(n) = \sigma\sqrt{\frac{\ln T}{n}}\cdot \max\left\{\sqrt{\frac{\eta_1 T}{nK}}, \sqrt{\eta_2}\right\}
\end{align}
satisfy the following two properties.
\begin{enumerate}
    \item If $\eta_1, \eta_2\geq 4$, then $\sup_{\theta}\mbE\left[R_\theta^\pi\right] \leq 4K + 8\sigma\sqrt{\max\{\eta_1, \eta_2\}KT\ln T}$.
    \item If $\eta_1 > 0, \eta_2 \geq 0$, then for any $x > 0$, we have
    \begin{align*}
        \sup_{\theta}\mbP(\hat R_\theta^\pi(T)\geq x) & \leq \exp\left(-\frac{x^2}{8K\sigma^2T}\right) + 4K\exp\left(-\frac{(x-2K-8\sigma\sqrt{(\eta_1\vee\eta_2) KT\ln T})_+^2}{128\sigma^2KT}\right) \\
        & \quad \quad + 2K^2T\exp\left(-\frac{(x-2K)_+\sqrt{\eta_1\ln T}}{16\sigma\sqrt{KT}}\right).
    \end{align*}
\end{enumerate}
\end{theorem}

The proof of Theorem \ref{thm:exp-K-armed-optimal} requires additional techniques compared to that of Theorem \ref{thm:exp-K-armed}: the main technical challenge lies in reducing the $K$ factor in Theorem \ref{thm:exp-K-armed} into a $\sqrt{K}$ factor. To address the challenge, we define a (random) arm set as
\begin{align*}
    \cA_0 = \left\{k\neq 1: n_k\leq 1 + T/K\right\}.
\end{align*}
The expected regret bound is then proved using standard techniques, but by considering arms in or not in $\cA_0$ separately. We stress that the standard techniques are feasible only when $\eta_1$ and $\eta_2$ are both not too small. Otherwise, it is not valid to show that the good event (the mean of each arm always lies in the confidence bounds throughout the whole time horizon) happens with high probability. To obtain better bounds when $\eta_1$ and $\eta_2$ are small, we should resort to the tail bound.

To prove the tail bound, we introduce a \textit{split-and-conquer} technique. To put it simple, we split the tail events into two categories depending on whether an arm is pulled more than $T/K$ times: when $k\in\cA_0$, we consider the event $n_k\Delta_k = \Omega(x/K)$; when $k\notin\cA_0$, we consider the event $\Delta_k=\Omega(x/T)$. As a result, in each event we consider, $\Delta_k$ is guaranteed to be $\Omega(x/T)$, and when $k\in\cA_0$, $\Delta_k$ enjoys a possibly better lower bound. Combined with the bonus design of $\sqrt{T/K}/n$, we can get an exponential $-\Omega(x/\sqrt{KT})$ term for the tail probability. We note that applying \eqref{eq:split} only yields $\Delta_k = \Omega(x/KT)$, leading to an exponential $-\Omega(x/K\sqrt{T})$ term. Detailed derivations are left to the online appendix.

For the tail bound, we can further simplify it as in Theorem \ref{thm:exp-K-armed}. If we let $\eta_1 = \eta > 0$, $\eta_2\in[0, \eta]$ and
\begin{align*}
    y = \frac{\left(x - 2K - 32\sigma \sqrt{(\eta\vee1/\eta)KT\ln T}\right)_+}{16\sigma \sqrt{KT}},
\end{align*}
then for any $x\geq 0$, we have
\begin{align*}
    \sup_{\theta}\mbP(R_\theta^\pi(T)\geq x) \leq 8K\exp\left(-y^2\wedge y\sqrt{\eta\ln T}\right).
\end{align*}
There are two observations:
\begin{enumerate}
\item For any $\eta > 0$, our policy always yields a $O(\sqrt{KT\ln T})$ expected regret (although with a varying constant term). In fact, notice that for any $x > 0$
\begin{align*}
    \mbE[\hat R_\theta^\pi(T)] = \mbE[R_\theta^\pi(T)] \leq x + \mbP(R_\theta^\pi(T)\geq x)\cdot T.
\end{align*}
If we let $x = 2K + C\sigma\sqrt{(\eta\vee1/\eta)KT\ln T}$ with the absolute constant $C$ being moderately large, then $\mbP(R_\theta^\pi(T)\geq x)\cdot T = O(1)$. As a result, the worst-case regret becomes
\begin{align*}
    O\left(\sigma\sqrt{(\eta\vee1/\eta)KT\ln T}\right).
\end{align*}

\item Let $\eta = 1$. For \textit{any} $\delta$, let
\begin{align*}
    x = 2K + 32\sigma\sqrt{KT\ln T} + 16\sigma\sqrt{KT}y
\end{align*}
where
\begin{align*}
    y = \sqrt{\ln(8K/\delta)} \vee \frac{\ln(8K/\delta)}{\sqrt{\ln T}}. 
\end{align*}
Then one can see that with probability at least $1-\delta$, the regret of our policy is bounded by
\begin{align*}
    O\left(\sigma \sqrt{KT}\left(\sqrt{\ln T} + \sqrt{\ln(8K/\delta)} \vee \frac{\ln(8K/\delta)}{\sqrt{\ln T}}\right)\right) = O\left(\sigma \sqrt{KT}\frac{\ln(T/\delta)}{\sqrt{\ln T}}\right).
\end{align*}
This partially answers the open question in Section 17.1 of \cite{lattimore2020bandit} (see Section \ref{ssec:literature}) up to a logarithmic factor. 

{\noindent\bf Why adversarial bandit policies may not work?} Before proceeding to further extensions, we would like to make a point on adversarial bandit policies. One may be tempted to apply the celebrated EXP3-IX algorithm under the adversarial setting (see Section \ref{ssec:literature}) to the stochastic setting and wonder if the adversarial approach can deliver similar performances as our proposed algorithm. However, this adversarial approach can have certain drawbacks. In the following, we take Algorithm 10 in \cite{lattimore2020bandit} as a concrete example.

1. From the algorithm design perspective, the uniform upper bound $C$ on all rewards need to be known a priori to transform the ``reward'' setting into the ``loss'' setting. To be more precise, let $\gamma$ be some regularization parameter and $p_{t, k}$ be the sampling probability of arm $k$ at time $t$, then rather than use
\begin{align*}
    \frac{r_{t, k}\mathds 1\{a_t=k\}}{\gamma+p_{t, k}}\geq 0
\end{align*}
as the estimator for the mean ``reward'' of arm $i$, it is recommended to use
\begin{align*}
    \frac{(C-r_{t, k})\mathds 1\{a_t=k\}}{\gamma+p_{t, k}} \geq 0
\end{align*}
as the estimator for the mean ``loss'' of arm $i$. Clearly, this approach requires the knowledge of $C$. Otherwise, the algorithm analysis can fail. In fact, if $C$ is not known a priori and one applies the ``reward'' estimator instead of the ``loss'' estimator, EXP3-IX can perform very bad due to its sampling strategy that violates the ``shift-invariant'' property --- an online learning policy is expected to perform similarly well if only the mean reward vector $\theta$ is shifted by a constant --- which holds true for stochastic bandit policies such as $\SE$ and $\UCB$. Consider two noiseless environments ($\sigma=0$) with $\theta = [0.2, 0.8]$ and $\tilde\theta = [5.2, 5.8]$. Let $T = 500$. Figure \ref{fig:exp3-ix} shows the empirical cumulative regret distribution of 5000 simulation paths by EXP3-IX for each environment. The tuning parameters are chosen according to Theorem 12.1.1 in \cite{lattimore2020bandit}. It indicates that if one has no prior knowledge of the reward upper bound $C$ so that the parameters are not tuned appropriately, applying the ``reward" estimator can leave the EXP3-IX policy rather questionable. The intuition is that if one samples a sub-optimal arm $a_t$ at time $t$ with a small probability $p_{t, a_t}$ and a small parameter $\gamma$, then a large $r_{t, a_t}$ may push the probabilities for the next time period $p_{t+1, k}$ towards zero for all $k\neq a_t$. As a result, the policy may frequently visit or even stick to a sub-optimal arm.

\begin{figure}
    \centering
    \includegraphics[width=12.5cm]{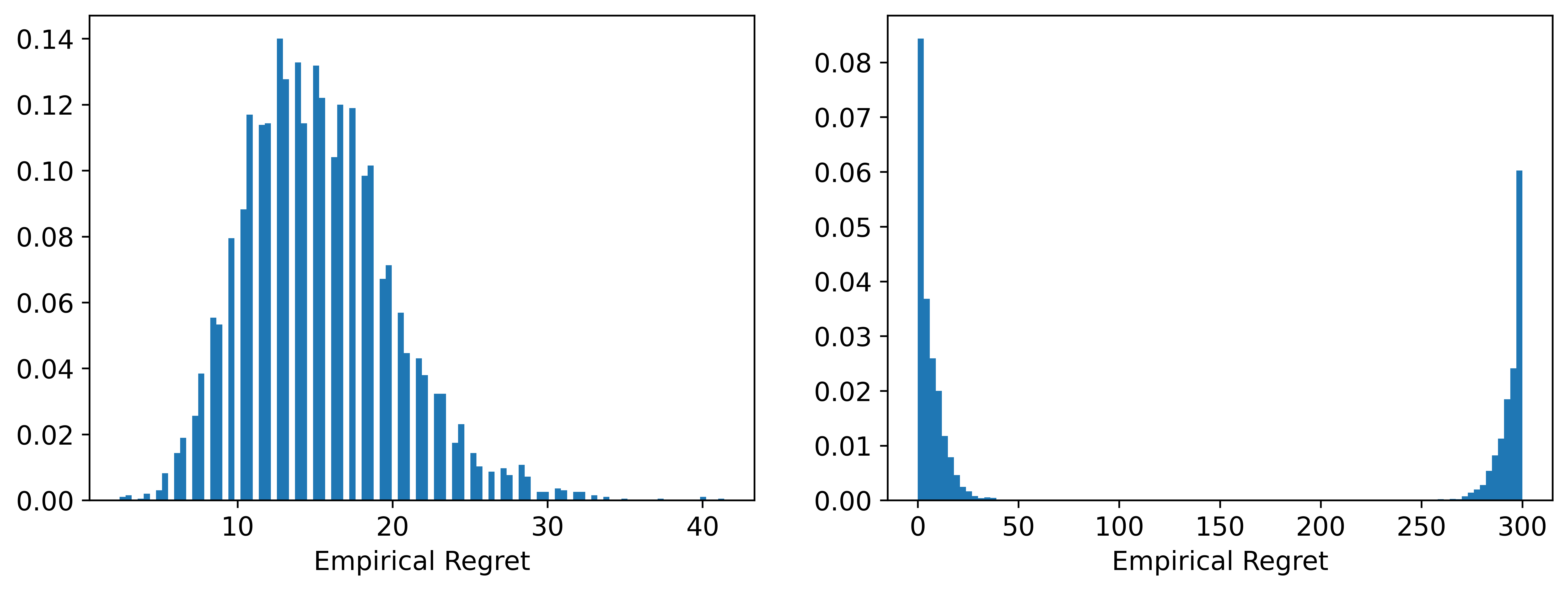}
    \caption{Regret Distribution of EXP3-IX (left for $\theta$ and right for $\tilde\theta$)}
    \label{fig:exp3-ix}
\end{figure}

2. With $\sigma>0$, one way to reduce the high-probability regret bound for EXP3-IX into one under the stochastic setting is as follows. For simplicity, we assume  $\theta\in[0, 1]^K$ and Gaussian noise with $\sigma=1$. First, a simple union bound suggests that with probability at least $1-2T\exp(-C^2/2)$, all the stochastic rewards are bounded within $[-C, 1+C]$. Then by applying Theorem 12.1.1 in \cite{lattimore2020bandit}, one knows that for any $\delta>0$, with probability at least $1-\delta/2 - 2T\exp(-C^2/2)$, the cumulative regret in the stochastic setting is bounded by
\begin{align*}
    O\left(C\sqrt{KT}\frac{\ln(K/\delta)}{\sqrt{\ln K}}\right).
\end{align*}
Now let $\delta/2 = 2T\exp(-C^2/2)$, then with probability at least $1-\delta$, the cumulative regret is
\begin{align*}
    O\left(\sqrt{\ln(T/\delta)}\sqrt{KT}\frac{\ln(K/\delta)}{\sqrt{\ln K}}\right).
\end{align*}
However, such bound requires $C$ to be dependent on the confidence parameter $\delta$, meaning a direct application of adversarial bandit policy requires knowing $\delta$ in advance (which is not desirable for obtaining a light-tailed policy), not to mention that the regret bound has a sub-optimal dependence of approximately $\ln(1/\delta)^{3/2}$ on $\delta$ (remember the optimal dependence should be $\ln(1/\delta$)).

3. Finally, when it comes to risk misspecification (e.g., we assume $\sigma=1$, but the truth is $\sigma_0\gg \sigma$), EXP3-IX may require a careful tuning of the input parameters as well as how $C$ is selected. Further, the EXP3-IX policy does not present a clear path to avoid the sensitivity to misspecifications. As a comparison, our new policy designs based on confidence bounds have advantages in self-adaptively handling risk misspecifications --- even if the risk profile is underestimated, our policies remain effective in achieving optimal rate with respect to $T$ for both expected regret bound and regret tail risk.

Apart from the points mentioned above, we would like to emphasize that the celebrated EXP3-IX remains to be a good policy, provided that the user of the policy has access to a well-specified uniform upper bound on all stochastic rewards. Meanwhile, if the environment keeps changing, then EXP3-IX is also a solid tool to hedge against irregular and strong non-stationarities. It would be interesting to see if our policy designs can be combined with EXP3-IX to achieve ideal performance in more complex environments, and we leave it for future work.

\end{enumerate}

\subsection{From Fixed-time to Any-time}

Finally, we enhance the policy design to accommodate the ``any-time" setting where $T$ is not known a priori, as a more challenging setting compared to the ``fixed-time" setting where $T$ is known a priori. We design a policy for the ``any-time" setting and prove that 
the policy enjoys an equivalently desired exponential decaying tail and optimal expected regret as in the ``fixed-time" setting. That is, our {any-time} policy enjoys a similar tail bound in Theorem \ref{thm:exp-K-armed-optimal}. In the following, we use $\rad_t(n)$ to denote the bonus term at time $t$.

\begin{theorem} \label{thm:any-time}
For the $K$-armed bandit problem, $\pi=\UCB$ with
\begin{align*}
    \rad_t(n) = \sigma\frac{\sqrt{\eta t(1\vee\ln(Kt))}}{n\sqrt{K}}
\end{align*}
satisfies the following property: fix any $\eta>0$, for any $x > 0$, we have
\begin{align*}
    \sup_{\theta}\mbP(\hat R_\theta^\pi(T)\geq x) & \leq \exp\left(-\frac{x^2}{8K\sigma^2T}\right) + 2KT^2\exp\left(-\frac{(x-2K-16\sigma\sqrt{2\eta KT\ln T})_+^2}{512\sigma^2KT}\right) \\
    & \quad \quad + 2KT^3\exp\left(-\frac{(x-2K)_+\sqrt{\eta\ln T}}{16\sigma\sqrt{KT}}\right).
\end{align*}

\end{theorem}

It is clear that for any $\eta > 0$, the policy in Theorem \ref{thm:any-time} always yields an expected regret of
\begin{align*}
    O(\sigma\sqrt{(\eta\vee 1/\eta)KT\ln T}).
\end{align*}
The reason is the same as that for Theorem \ref{thm:exp-K-armed-optimal}. Another remark is that Theorem \ref{thm:any-time} only involves the $\UCB$ policy. In fact, the $\SE$ policy can always fail under an any-time bonus design. This is because $\SE$ will never pull an arm if this arm was eliminated previously. Therefore, even in the basic 2-armed setting, at the beginning when $t$ is small compared to $T$, the behaviour of $\SE$ with $\rad_t(n)$ can be nearly as worse as that of $\SE$ with \eqref{rad:standard}: it may eliminate the optimal arm with a probability heavy-tailed in $T$. On the contrary, in $\UCB$, arms are always activated, and so the gradually time-increasing numerator in the bonus term will take effect and prevents the optimal arm from being discarded forever.

The bonus design $\rad_t(n)$ in Theorem \ref{thm:any-time} can be approximately regarded as replacing the $T$ term in \eqref{rad:optimal} with $t$. We use $1\vee \ln (Kt)$ instead of $\ln t$ primarily out of convenience for analysis. The proof of Theorem \ref{thm:any-time} posits additional challenges to that of Theorem \ref{thm:exp-K-armed-optimal} due to the unfixed $t$ in the bonus term. In the proof of Theorem \ref{thm:exp-K-armed-optimal}, in each event induced by the split-and-conquer technique, we always have $\Delta_k=\Omega(x/T)$. However, such bound may not be enough for the any-time case, since if we follow the proof of Theorem \ref{thm:exp-K-armed-optimal}, the tail probability of wrongly discarding the optimal arm can only be bounded by
\begin{align*}
    \exp\left(-\Omega(x/T\cdot\sqrt{t/K})\right),
\end{align*}
which is not an informative bound with an uncontrolled $t$. Also, it is unclear whether a $\sqrt{\ln T}$ term can be produced in the last term of the tail bound (the probability of wrongly discarding the optimal arm) under an any-time bonus design, which is essential to obtain an expected $O(\sqrt{T\ln T})$ regret bound. Both issues show that we need to rectify the set $\cA_0$ such that $\Delta_k$ enjoys a possibly better bound depending on $t_k$, and that $\ln t_k$ is connected with $\ln T$ in the analysis. Our proof further improves the split-and-conquer technique by defining a corrected arm set
\begin{align*}
    \cA_1 = \left\{k\neq 1: n_k\leq 1 + \frac{t_k^{3/4}T^{1/4}}{K}\right\},
\end{align*}
where $t_k=t_k(n_{T, k})$ is the last time period we pull arm $k$ across the time horizon of $T$. When $k\in\cA_1$, we consider the event $n_k\Delta_k = \Omega(x/K)$; when $k\notin\cA_0$, we consider the event $\Delta_k=\Omega(x/\sqrt{t_k T})$. The specific choice of $\cA_1$ allows us to ensure $\Delta_k = \Omega(x/\sqrt{t_k T})$, and meanwhile ensure the additional $\sqrt{\ln T}$ factor in the last term of the tail bound. Details are all left to the online appendix. We also need to note that since $n_k$ and $t_k$ are both random variables, when bounding the probabilities, we must use a union bound to cover through all possible pairs $(n_k, t_k)$. This is the reason why we have an additional $T^2$ factor in front of the exponential tail.

\section{Numerical Illustration}

\label{sec:experiments}

We now provide numerical experiments to illustrate our theoretical findings.

\subsection{Impact of Hyper-parameters}
\label{ssec:experiments:hyper-tune}
In this section, we study how different hyper-parameters affect the performance of different policies under a \textit{fixed} environment. To be specific, we will fix an environment, and for each policy we consider, we adjust the hyper-parameters to see how sensitive the policy is with respect to tail risk.

\begin{figure}[!ht]
    \centering
    \includegraphics[width=15cm]{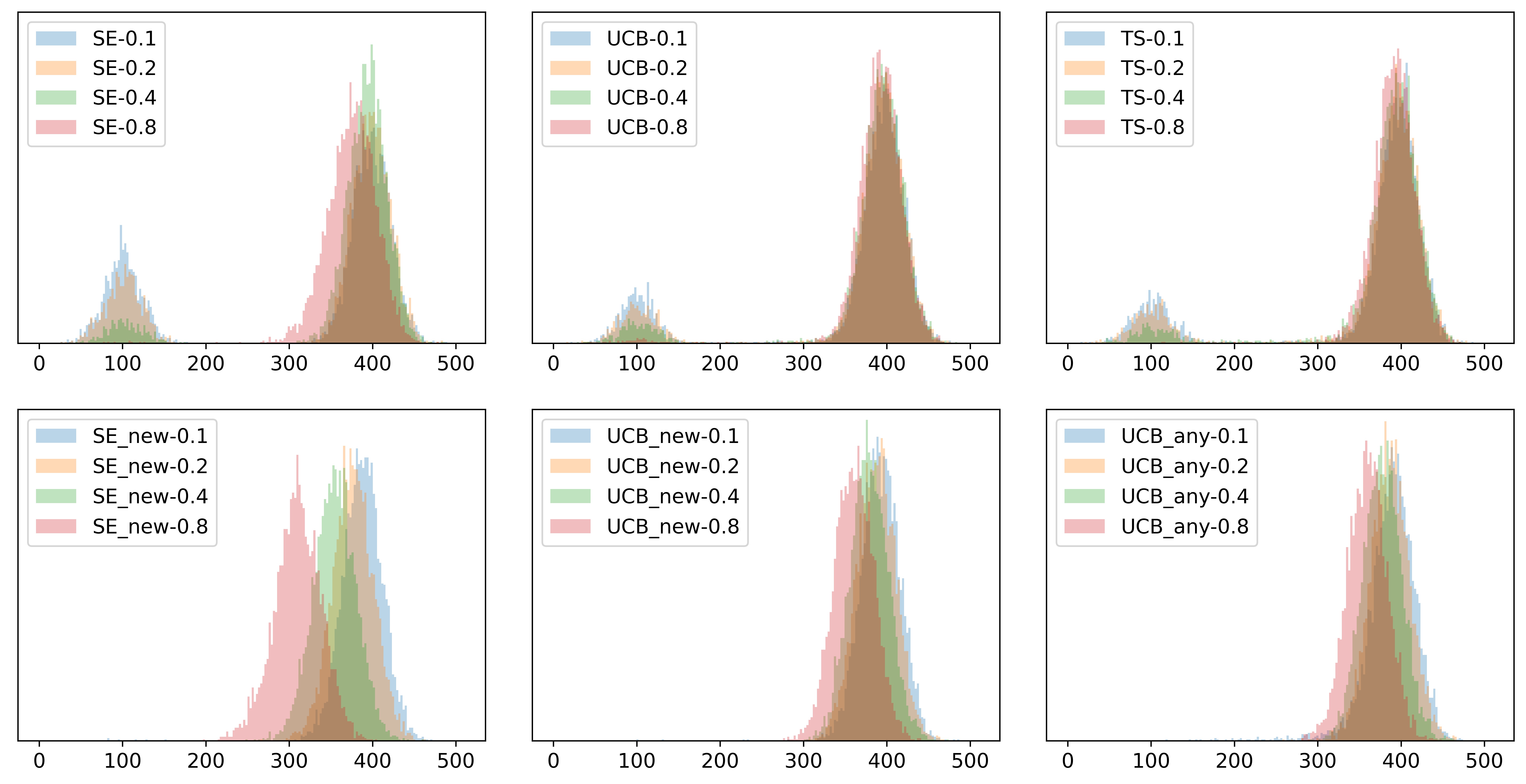}
    \caption{Empirical distribution for the cumulative reward}
    \label{fig:reward-distribution-2}
\end{figure}

We first consider a $2$-armed bandit problem with $\theta = (0.2, 0.8)$, $\sigma=1$, $T = 500$ and Gaussian noise. We test five confidence bound type policies: $\SE$ and $\UCB$ with the classical bonus design described in \eqref{rad:standard}, $\SE\_\text{new}$ and $\UCB\_\text{new}$ with the proposed new bonus design in \eqref{rad:new}, and $\UCB\_\text{any}$ with the bonus design $\rad_t(n)$ in Theorem \ref{thm:any-time}. We let $\kappa\triangleq \sigma\sqrt{\eta}$. The tuning parameter $\kappa$ has 4 choices: $\kappa\in\{0.1, 0.2, 0.4, 0.8\}$. We also consider the $\ThS$ policy assuming the mean reward of each arm $i$ following the prior $\mathcal N(0, 1)$ and the sample from each arm $i$ following $\mathcal N(\theta_i, \kappa^2)$. That said, we evaluate $\ThS$ under mis-specified risk parameters. For each policy and $\kappa$, we run $5000$ simulation paths and for each path we collect the cumulative reward. We provide the empirical mean for the cumulative reward in Table \ref{table:mean-2}. We also plot the empirical distribution (histogram) for a policy's  cumulative reward in Figure \ref{fig:reward-distribution-2}. 

\begin{table}[!ht]
    \centering
    \begin{tabular}{|c|c|c|c|c|}\hline
       \diagbox{Policy}{$\kappa$}& $0.1$ & $0.2$ & $0.4$ & $0.8$ \\\hline
       $\SE$ & $311.60$ & $336.46$ & $375.53$ & $374.69$ \\\hline
       $\UCB$ & $349.68$ & $359.68$ & $377.17$ & $390.23$ \\\hline
       $\ThS$ & $351.00$ & $360.71$ & $377.94$ & $390.32$ \\\hline
       $\SE\_\text{new}$ & $388.16$ & $376.69$ & $354.25$ & $309.58$ \\\hline
       $\UCB\_\text{new}$ & $393.27$ & $387.48$ & $377.72$ & $360.69$ \\\hline
       $\UCB\_\text{any}$ & $391.66$ & $387.60$ & $377.37$ & $359.59$ \\\hline
    \end{tabular}
    \caption{Empirical mean for the cumulative reward}
    \label{table:mean-2}
\end{table}

Table \ref{table:mean-2} shows that, SE\_new (or UCB\_new) achieves empirical mean for the cumulative reward as high as SE (or UCB) can achieve. The highest empirical mean for the cumulative reward that can be achieved by SE\_new (or UCB\_new) with various choices of $\kappa$ is comparable to the highest empirical mean that can be achieved by SE (or UCB). We note that there is no direct implication by comparing all different algorithms at the same value of $\kappa$, because the algorithms use the parameter $\kappa$ in different ways. For example, for some value of $\kappa$, SE has a higher empirical mean for the cumulative reward compared to SE\_new, whereas for some other value of $\kappa$, SE has a smaller empirical mean compared to SE\_new. There is no direct implication by fixing a value of $\kappa$ and comparing different algorithms. Nevertheless, we do observe that $\ThS$ performs similarly to $\UCB$, from both Figure \ref{fig:reward-distribution-2} and Table \ref{table:mean-2}, and so we put our discussion on confidence bound policies. Figure \ref{fig:reward-distribution-2} shows that, compared to SE, $\SE\_\text{new}$ has much lower probability of incurring a low cumulative reward. The implication is that (i) in terms of the empirical mean of cumulative reward, SE\_new is as good as SE; (ii) in terms of the risk of incurring a low cumulative reward, SE\_new can be much better (i.e., lower risk) than SE, particularly when $\kappa$ is relatively small. The same implication holds analogously for the comparison between UCB\_new and UCB. Indeed, one can observe that for both $\SE$ and $\UCB$ with \eqref{rad:standard} and $\ThS$, there is a significant part of distribution around $100$, indicating a significant risk of incurring a linear regret when $\kappa$ is relatively small. In contrast, with the new design \eqref{rad:new}, the reward is highly concentrated for every $\kappa > 0$ with almost no tail risk of getting a low total reward. Particularly, when $\kappa=0.1$ or $\kappa=0.2$, $\UCB\_\text{new}$ achieves both high empirical mean and light-tailed distribution.

\begin{figure}[!ht]
    \centering
    \includegraphics[width=15cm]{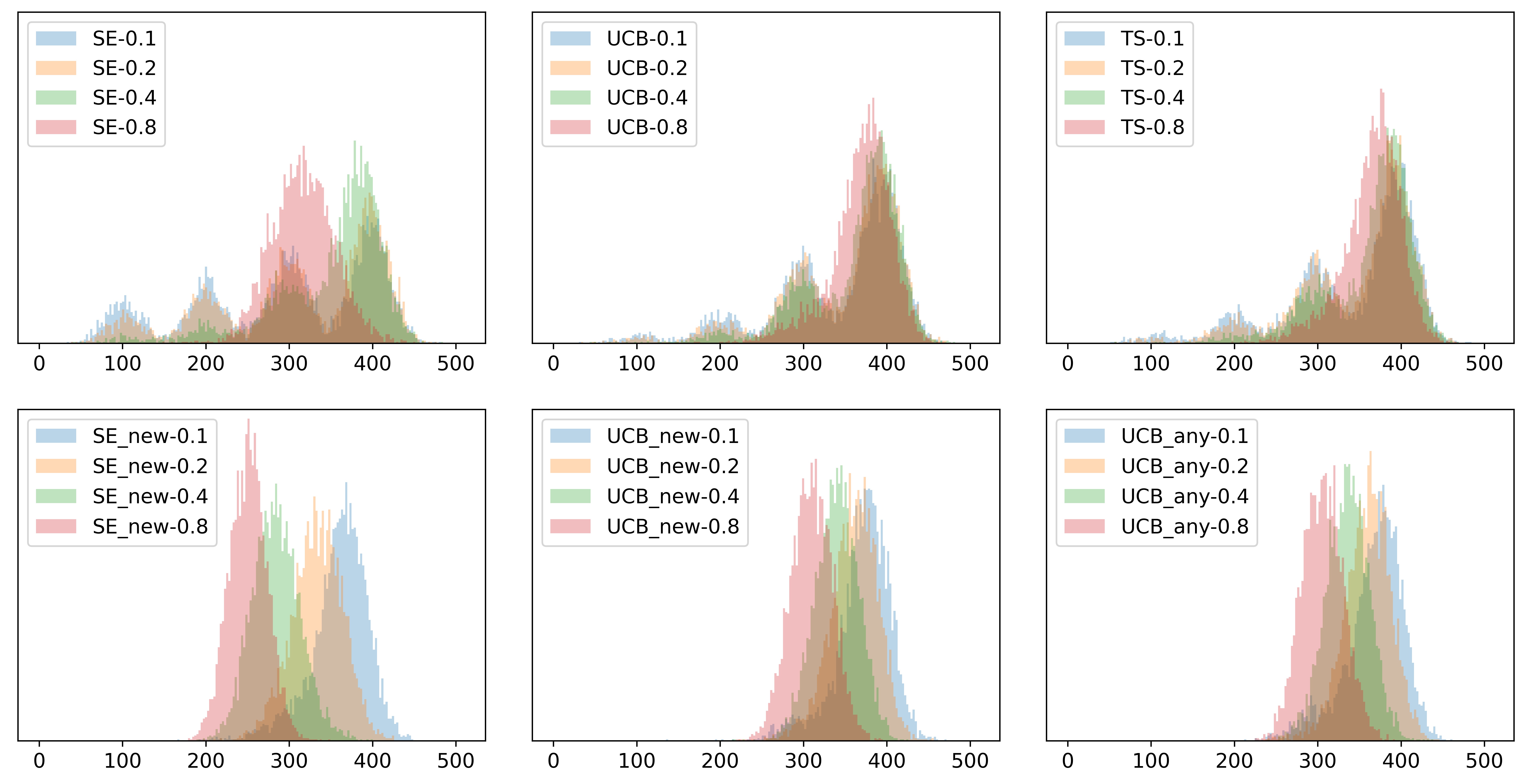}
    \caption{Empirical distribution for the cumulative reward}
    \label{fig:reward-distribution-4}
\end{figure}

Next, we consider a $4$-armed bandit problem with $\theta = (0.2, 0.4, 0.6, 0.8), \sigma=1, T = 500$ and Gaussian noise. Same as in the two-armed case, we test six policies: $\SE$, $\UCB$, $\ThS$, $\SE\_\text{new}$, $\UCB\_\text{new}$, $\UCB\_\text{any}$. The tuning parameter has 4 choices: $\kappa\in\{0.1, 0.2, 0.4, 0.8\}$. For each policy and $\kappa$, we run $5000$ simulation paths and for each path we collect the cumulative reward. We plot the empirical distribution (histogram) for a policy's  cumulative reward in Figure \ref{fig:reward-distribution-4}. We also report the empirical mean in Table \ref{table:mean-4}. 

\begin{table}[!ht]
    \centering
    \begin{tabular}{|c|c|c|c|c|}\hline
       \diagbox{Policy}{$\kappa$}& $0.1$ & $0.2$ & $0.4$ & $0.8$ \\\hline
       $\SE$ & $293.11$ & $311.74$ & $351.81$ & $316.64$ \\\hline
       $\UCB$ & $339.41$ & $348.52$ & $360.26$ & $369.25$ \\\hline
       $\ThS$ & $341.05$ & $349.86$ & $359.82$ & $365.26$ \\\hline
       $\SE\_\text{new}$ & $361.93$ & $334.18$ & $283.69$ & $251.52$ \\\hline
       $\UCB\_\text{new}$ & $371.10$ & $361.13$ & $339.29$ & $309.71$ \\\hline
       $\UCB\_\text{any}$ & $368.86$ & $359.87$ & $335.68$ & $305.88$ \\\hline
    \end{tabular}
    \caption{Empirical mean for the cumulative reward}
    \label{table:mean-4}
\end{table}

Indeed, one can observe that for both $\SE$ and $\UCB$ with \eqref{rad:standard}, there is a significant part of distribution around $200$ and $300$ when $\kappa$ is relatively small, which means that with an non-negligible probability the two policies always pull arm $2$ or $3$, incurring a linear regret. In contrast, with the new design \eqref{rad:new}, the reward is highly concentrated for every $\kappa > 0$. Particularly, when $\kappa=0.1$, either $\SE\_\text{new}$ or $\UCB\_\text{new}$ achieves both high empirical mean and light-tailed distribution. When $\kappa$ is relatively large, e.g., $\kappa=0.8$, the empirical mean is not very satisfactory. This is consistent with Theorem \ref{thm:exp-K-armed}, which indicates an additional $\sqrt{K}$ factor compared to the optimal $\tilde O(\sqrt{KT})$ expected regret, if $\kappa$ is not scaled by a factor of $\sqrt{K}$ as in \eqref{rad:optimal}.

Apart from the observations above, we would like to emphasize that from a managerial perspective, celebrated policies ($\SE$ and $\UCB$ with the standard bonus term \eqref{rad:standard}, and $\ThS$) remain to be good policies (see, e.g., Figure \ref{fig:reward-distribution-2} and \ref{fig:reward-distribution-4} when $\kappa=0.8$), provided that the user (i) has access to the true risk profile of the environment (or at least a reasonable upper bound of $\sigma$) and (ii) tunes the hyper-parameter $\kappa$ appropriately. After all, a tail probability of $O(1/T^{2})$ may not be substantial in real experiments when $T$ is not too small. However, if either of the two above-mentioned requirement is not satisfied, the celebrated policies can deteriorate significantly. As a comparison, the new policy designs yield better tail distributions as well as more robustness/insensitivity towards hyper-paramter tuning by little or no sacrifice on the regret expectation. This point will be further illustrated in Section \ref{ssec:experiments:noise-mis}.

\subsection{Tail Decaying Rate}
In this section, we testify how different policies behave on the tail risk of incurring a linear regret. For each policy we consider, we will adopt its well tuned version built on the results in Section \ref{ssec:experiments:hyper-tune} and test it in a \textit{fixed} environment (with $T$ growing), and see how sensitive the policy is with respect to tail risk.

\begin{figure}[!ht]
    \centering
    \includegraphics[width=15cm]{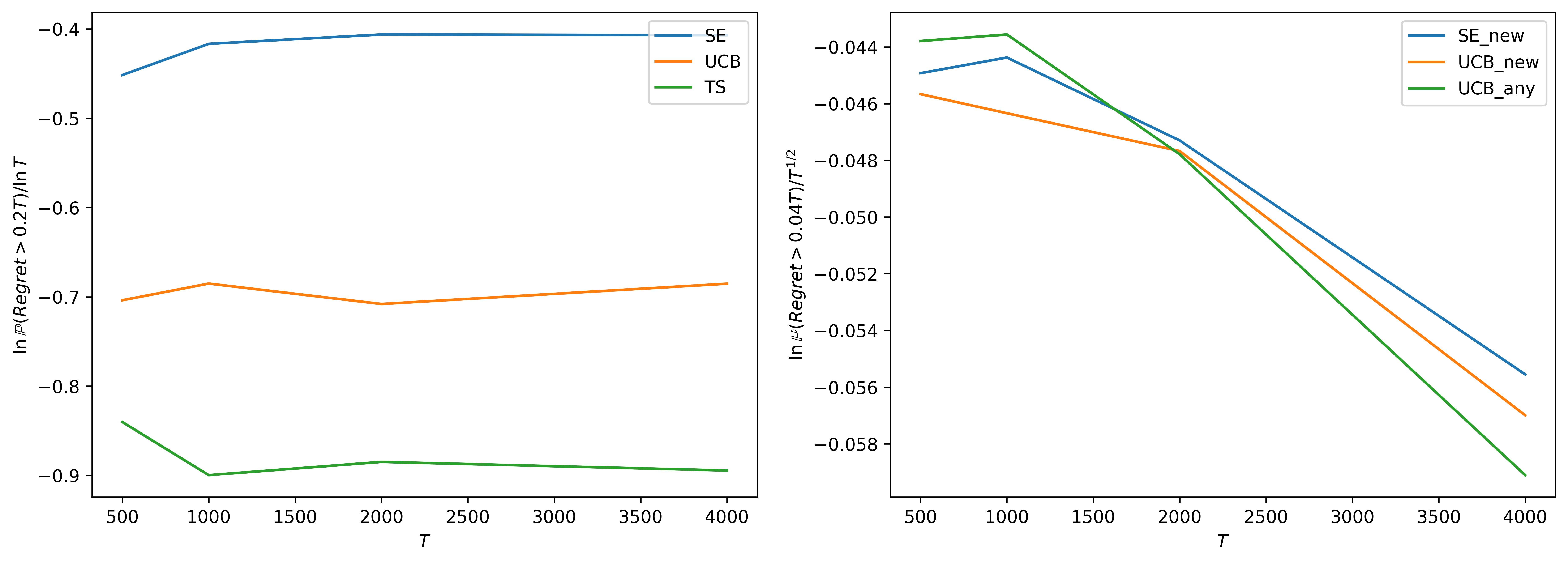}
    \caption{Regret tail decaying rate. Left: $\ln\mbP(\hat R_\theta^\pi(T)>0.2T)/\ln T$ vs $T$. Right: $\ln\mbP(\hat R_\theta^\pi(T)>0.04T)/\sqrt{T}$ vs $T$.}
    \label{fig:reward-distribution-tail-2}
\end{figure}

We first consider the two armed-bandit problem with $\theta = (0.2, 0.8)$, and Gaussian noise with $\sigma=1$. With all other factors fixed, we vary $T$ by taking $T\in\{500\cdot i|i = 1, 2, 4, 8\}$. Similar as before, we test $6$ policies, but each policy is now endowed with a fixed hyper-parameter $\kappa$ (remember $\kappa\triangleq \sigma\sqrt{\eta}$) fine tuned from the experiments in Section \ref{ssec:experiments:hyper-tune}: $\SE$ with the classical bonus design described in \eqref{rad:standard} ($\kappa=0.4$), $\UCB$ with the classical bonus design described in \eqref{rad:standard} ($\kappa=0.8$), $\ThS$ ($\kappa=0.8$), $\SE\_\text{new}$ with the proposed new bonus design in \eqref{rad:new} ($\kappa=0.1$), $\UCB\_\text{new}$ with the proposed new bonus design in \eqref{rad:new} ($\kappa=0.2$), and $\UCB\_\text{any}$ with the bonus design $\rad_t(n)$ in Theorem \ref{thm:any-time} ($\kappa=0.2$). For each policy, we run $5000$ simulation paths and for each path we collect the cumulative reward. Figure \ref{fig:reward-distribution-tail-2} shows the empirical regret tail decaying rate. For the standard policies $\SE$, $\UCB$, $\ThS$, we plot $\ln\mbP(\hat R_\theta^\pi(T)>0.2T)/\ln T$ versus $T$. For our new proposed policies $\SE\_\text{new}$, $\UCB\_\text{new}$, $\UCB\_\text{any}$, we plot $\ln\mbP(\hat R_\theta^\pi(T)>0.04T)/\sqrt T$ versus $T$. 

Figure \ref{fig:reward-distribution-tail-2} shows numerical evidence consistent with our theoretical findings. Note that the more negative the value on the $y$-axis, the more light tailed the policy behaves. In particular, for a standard policy (on the left), the probability of incurring a linear regret scales as a polynomial $T$ term. For our new policy designs, the probability of incurring a linear regret decays at an $\exp(-\Omega(\sqrt{T}))$ rate. We would like to note that we set the threshold $0.04$ for new policy designs rather than $0.2$ (as in the left) because we found in our simulations that in some cases the empirical probability $\mbP(\hat R_\theta^\pi(T)>0.2T)$ can be zero, particularly when $T$ grows large. 

\begin{figure}[!ht]
    \centering
    \includegraphics[width=15cm]{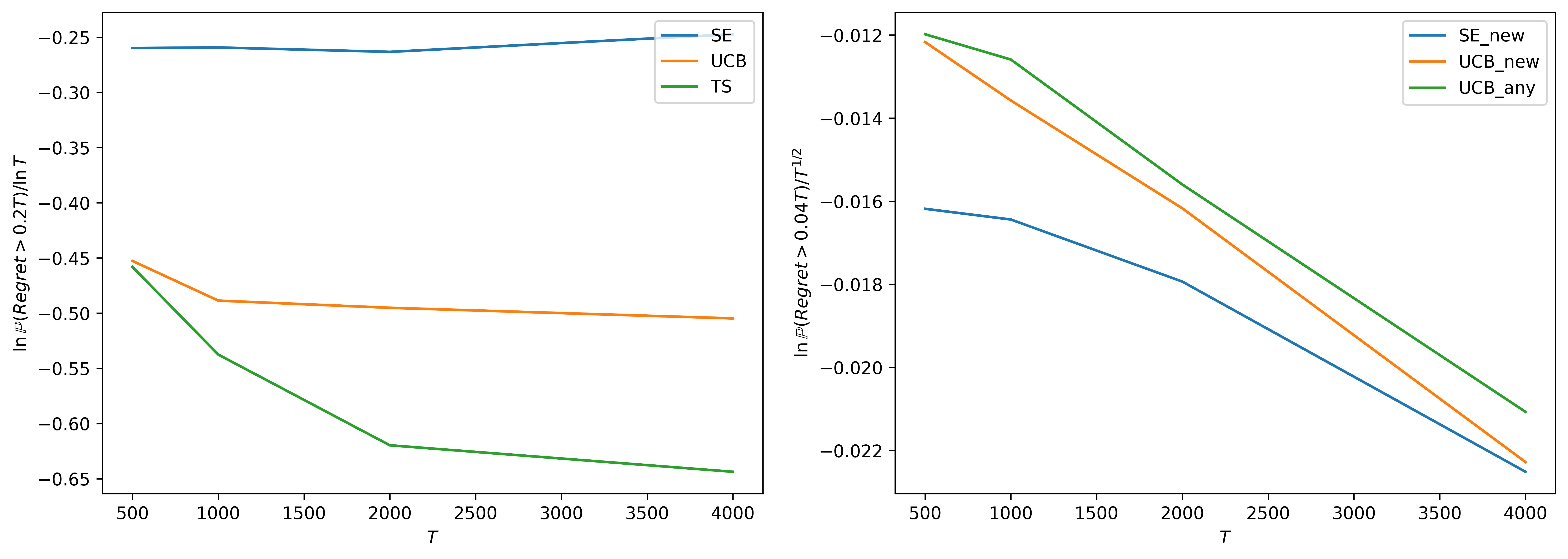}
    \caption{Regret tail decaying rate. Left: $\ln\mbP(\hat R_\theta^\pi(T)>0.2T)/\ln T$ vs $T$. Right: $\ln\mbP(\hat R_\theta^\pi(T)>0.04T)/\sqrt{T}$ vs $T$.}
    \label{fig:reward-distribution-tail-4}
\end{figure}

Next, we consider a $4$-armed bandit problem. All the settings are exactly the same as in the two-armed bandits case stated above (including the environments and the policies with corresponding hyper-parameters). The only difference is that we let $\theta=(0.2, 0.4, 0.6, 0.8)$. For each policy, we run $5000$ simulation paths and for each path we collect the cumulative reward. Figure \ref{fig:reward-distribution-tail-4} shows the empirical regret tail decaying rate. Same as in the $2$-armed bandit case, for the standard policies $\SE$, $\UCB$, $\ThS$, we plot $\ln\mbP(\hat R_\theta^\pi(T)>0.2T)/\ln T$ versus $T$. For our new proposed policies $\SE\_\text{new}$, $\UCB\_\text{new}$, $\UCB\_\text{any}$, we plot $\ln\mbP(\hat R_\theta^\pi(T)>0.04T)/\sqrt T$ versus $T$. We observe the same phenomenon as in the $2$-armed bandit case. 

\subsection{Noise Misspecification} \label{ssec:experiments:noise-mis}
In this section, we study how misspecified noise parameters affect the performance of different policies under a \textit{fixed} hyper-parameter. For each policy we consider, we will put its well tuned version built on the results in Section \ref{ssec:experiments:hyper-tune} (under a less noisier environment) into a much noisier environment (with the hyper-parameter fixed) and see how sensitive the policy is with respect to tail risk.

We first consider the $2$-armed bandit base problem with $\theta = (0.2, 0.8)$, $\sigma=1$, $T = 500$ and Gaussian noise. We construct $4$ scaled environments as follows. In instance $i\in\{1, 2, 4, 8\}$, $\theta=(0.2, 0.8)$ remains the same. Noises are still Gaussian, but with $\sigma_0^2 = k\sigma^2$. Accordingly, $T_0 = kT$. The instances are constructed such that the efficiency of estimating an unknown arm remains the same: intuitively speaking, given $T_0$ samples with $\sigma_0$-Gaussian noise for one arm, the mean squared error of estimation error for an arm is at the order of $\sigma_0^2/T_0$. 

\begin{figure}[!ht]
    \centering
    \includegraphics[width=15cm]{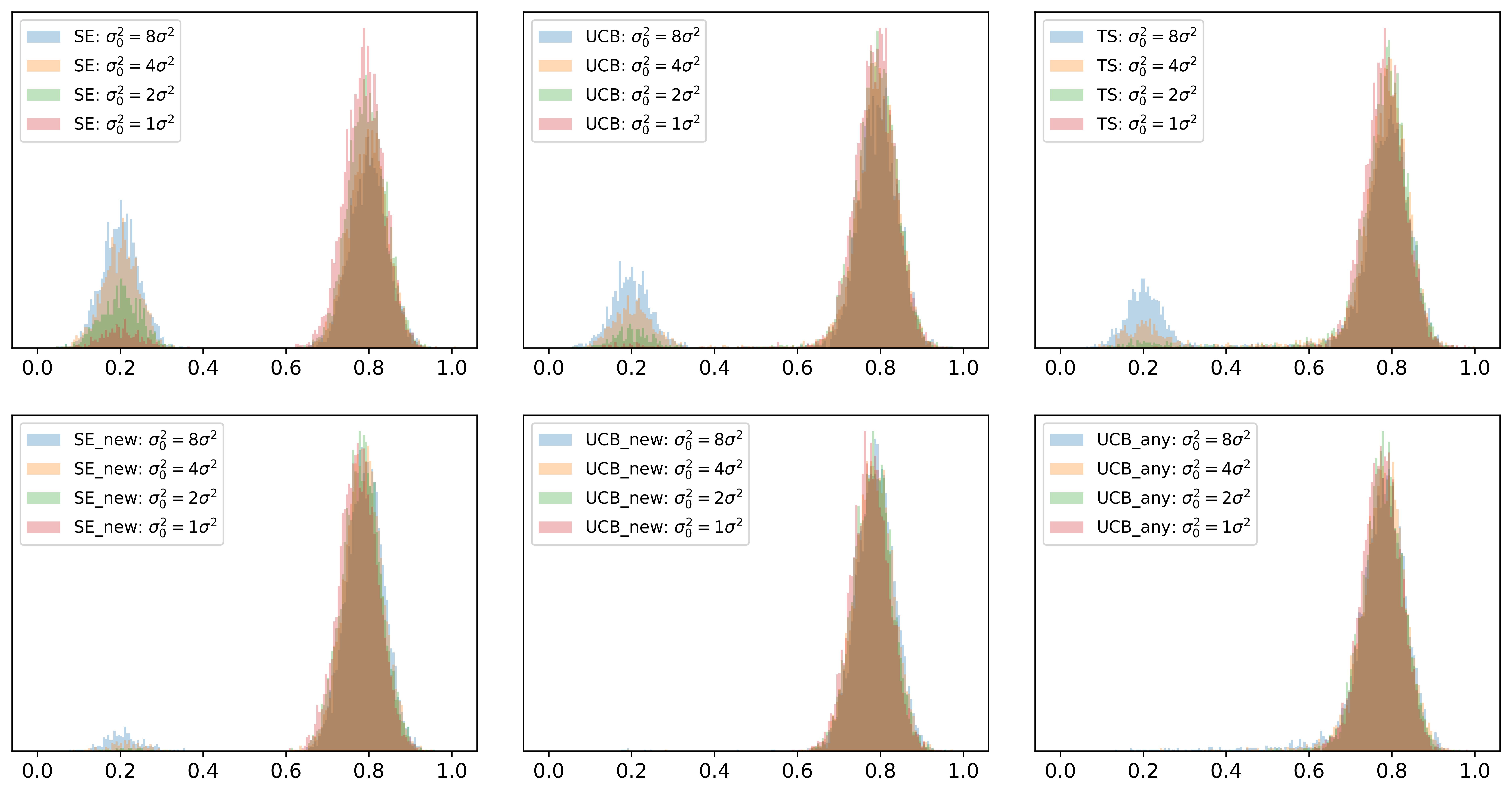}
    \caption{Empirical distributions for the cumulative reward; $x$-axis normalized}
    \label{fig:reward-distribution-mis-2}
\end{figure}

Similar as before, we test $6$ policies, but each policy is now endowed with a fixed hyper-parameter $\kappa$ (remember $\kappa\triangleq \sigma\sqrt{\eta}$) fine tuned from the previous experiments: $\SE$ with the classical bonus design described in \eqref{rad:standard} ($\kappa=0.4$), $\UCB$ with the classical bonus design described in \eqref{rad:standard} ($\kappa=0.8$), $\ThS$ ($\kappa=0.8$), $\SE\_\text{new}$ with the proposed new bonus design in \eqref{rad:new} ($\kappa=0.1$), $\UCB\_\text{new}$ with the proposed new bonus design in \eqref{rad:new} ($\kappa=0.2$), and $\UCB\_\text{any}$ with the bonus design $\rad_t(n)$ in Theorem \ref{thm:any-time} ($\kappa=0.2$). For each policy, we run $5000$ simulation paths and for each path we collect the cumulative reward. We plot the empirical distribution (histogram) for a policy's cumulative reward in Figure \ref{fig:reward-distribution-mis-2}. Note that since $T_0$ can be different, for each environment we do a normalization on the $x$-axis via scaling it by a factor of $T_0$.

Figure \ref{fig:reward-distribution-mis-2} shows that, compared to SE, $\SE\_\text{new}$ is much less sensitive to misspecified noise parameters $\sigma_0$. The same phenomenon holds analogously for the comparison between UCB\_new and UCB. Indeed, one can observe that for both $\SE$ and $\UCB$ with \eqref{rad:standard} and $\ThS$, there is a significant part of distribution around $0.2$, indicating a significant risk of incurring a linear regret, particularly when $\sigma_0^2=8\sigma^2$. In contrast, with the new design \eqref{rad:new}, the reward is highly concentrated with almost no tail risk of getting a low total reward, \textit{regardless of $\sigma_0$}. We also observe that $\UCB$-type policies perform better than $\SE$-type policies.

\begin{figure}[!ht]
    \centering
    \includegraphics[width=15cm]{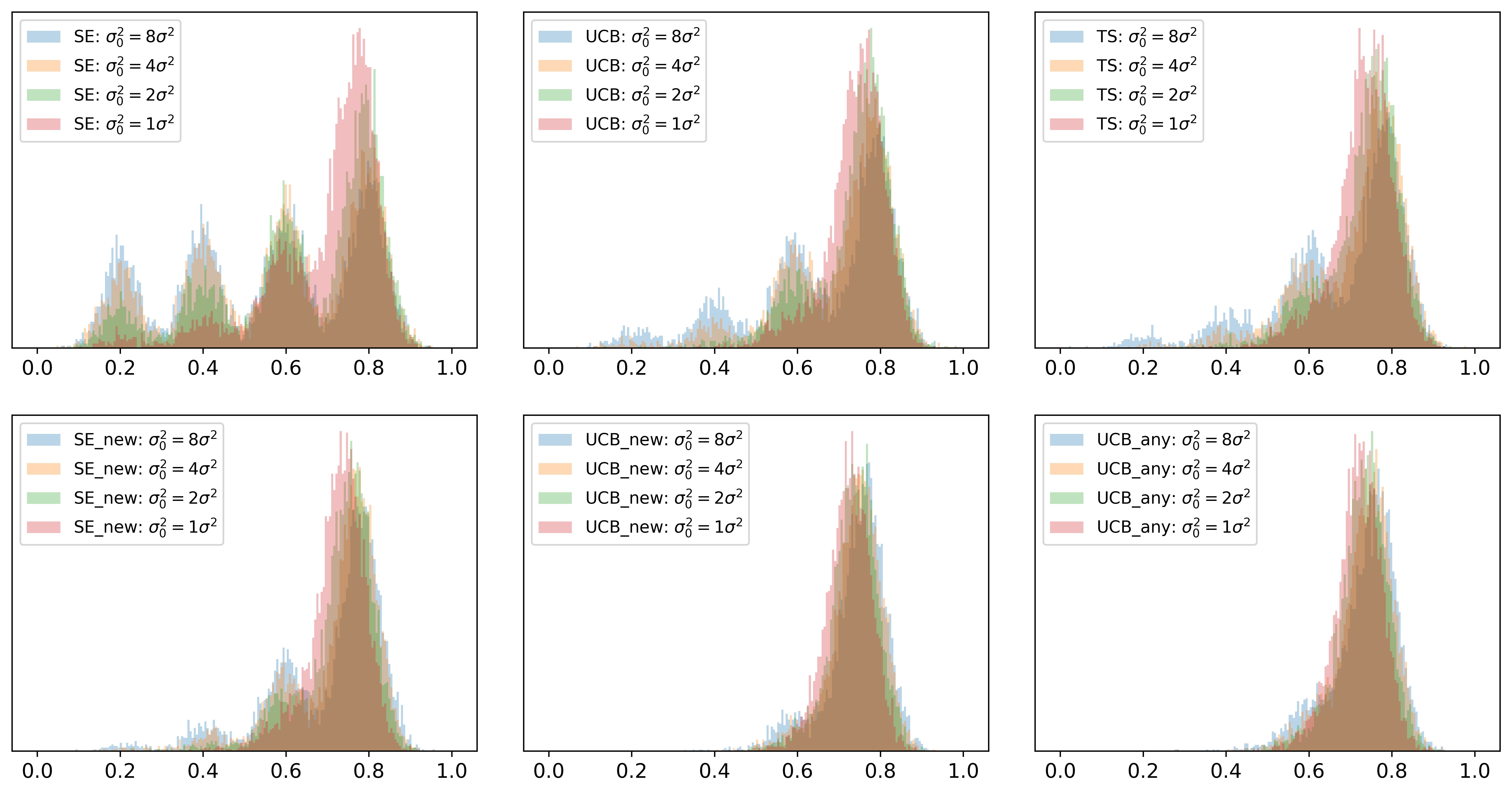}
    \caption{Empirical distribution for the cumulative reward; $x$-axis normalized}
    \label{fig:reward-distribution-mis-4}
\end{figure}

Next, we consider a $4$-armed bandit problem. All the settings are exactly the same as in the two-armed bandits case stated above (including the scaled environments and the policies with corresponding hyper-parameters). The only difference is that we let $\theta=(0.2, 0.4, 0.6, 0.8)$. For each policy, we run $5000$ simulation paths and for each path we collect the cumulative reward. We plot the empirical distribution (histogram) for a policy's cumulative reward in Figure \ref{fig:reward-distribution-mis-4}. 

Indeed, one can observe that for both $\SE$ and $\UCB$ with \eqref{rad:standard}, there is a significant part of distribution around $0.4$ and $0.6$, especially when $\sigma_0$ becomes larger. In contrast, with the new design \eqref{rad:new}, the reward is much more concentrated for every $\sigma_0$. Notably, the $\UCB$-type policies perform better than the $\SE$-type policies. Nevertheless, the results show that compared to standard policy designs, our new policy designs significantly reduce both the tail risk of incurring a large regret and the sensitivity to model misspecifications.

\section{Extension to Linear Bandits} \label{sec:linear}
In this section, we further extend our policy design to the setting of linear bandits. We show that the simple policy design that leads to both optimality and light-tailed risk for the multi-armed bandit setting can be naturally extended to the linear bandit setting. We briefly review the setting of linear bandits as follows (see, e.g., \citealt{dani2008stochastic}, \citealt{abbasi2011improved}, for reference of more details). In each time period $t$, the decision maker (DM) is given an action set $\cA_t\subseteq \mathbb R^d$ from which the DM needs to select one action $a_t\in\cA_t$ to take for the time period $t$. Subsequently a reward of $r_t = \theta^{\top}a_t + \epsilon_{t, a_t}$ is collected, where $\theta\in\mathbb R^d$ is an unknown parameter and $\epsilon_{t, a_t}$ is an independent $\sigma$-subGaussian mean-zero noise. More specifically, let $H_t = \{a_1, r_{1, a_1}, \cdots, a_{t-1}, r_{t-1, a_{t-1}}\}$ be the history prior to time $t$. When $t=1$, $H_1=\emptyset$. At time $t$, the DM adopts a policy $\pi_t: H_t\longmapsto a_t$ that maps the history $H_t$ to an action $a_t$, where $a_t$ may be realized from some probability distribution on $\cA_t$. Adopting the standard assumptions in the linear bandits literature, we presume that $\|\theta\|_{\infty} \leq 1$ and $\|a\|_2 \leq 1$ for any $a\in\cA_t$ and any $t$. Let $a_t^*=\arg\max_{a\in\cA_t}\theta^\top a$. The empirical regret is defined as
\begin{align*}
    \hat R_{\theta}^\pi(T) = \sum_{t=1}^T \theta^\top a_t^{*} - \sum_{t=1}^T r_{t, a_t} = \sum_{t=1}^T\theta^\top (a_t^* - a_t) - \sum_{t=1}^T\epsilon_{t, a_t} \triangleq R_{\theta}^\pi(T) - N^\pi(T).
\end{align*}
Same as in the MAB setting, $N^\pi(T)$ also enjoys the fast concentration property in Lemma \ref{lemma:bound-noise}. We provide the Linear UCB policy ($\UCB\text{\rm -L}$) in Algorithm \ref{alg:UCB-lin} and show in Theorem \ref{thm:linear} an explicit exponentially decaying regret tail bound under a carefully specified bonus term $\rad_t(z)$. Note that in standard bonus design, $\rad_t(z)\propto \sqrt{z}$ (see, e.g., the OFUL policy in \citealt{abbasi2011improved}). We need to emphasize that the additional $\sqrt{dz}$ term in \eqref{rad:linear} is necessary, without which the policy may be lack of exploration at the very beginning and then always stick to a suboptimal action for a small $\sigma$. 

\bigskip
\begin{breakablealgorithm}
\caption{Linear UCB (UCB-L)}
\label{alg:UCB-lin}
\begin{algorithmic}[1]
\State $t\gets 0$, $V_0=I$, $\hat\theta_0=0$.
\While{$t < T$}
    \State $t\gets t+1$. Observe $\cA_t$.
    \State Take the action with the highest $\UCB$:
    \begin{align*}
        a_t=\arg\max_{a\in\cA_t}\left\{\hat\theta_{t-1}^\top a  + \rad_t(a^\top V_{t-1}^{-1}a)\right\}.
    \end{align*}
    \State $V_t=V_{t-1} + a_ta_t^\top$, $\hat\theta_t = V_t^{-1}(\sum_{s\leq t}a_sr_s)$.
\EndWhile
\end{algorithmic}
\end{breakablealgorithm}
\bigskip

\begin{theorem} \label{thm:linear}
Let $T\geq d$. For the stochastic linear bandit problem, $\pi=\UCB\text{\rm -L}$ with
\begin{align} \label{rad:linear}
    \rad_t(z) = z\sigma\sqrt{\frac{\eta t}{d}}+\sqrt{dz}
\end{align}
satisfies the following property: for any fixed $\eta > 0$, for any $x>0$, we have
\begin{align*}
    & \quad \sup_{\theta}\mbP(\hat R_{\theta}^\pi(T)\geq x) \\
    & \leq \exp\left(-\frac{x^2}{2\sigma^2d^2T}\right)+ 2d(T/d)^{2d+1}\exp\left(-\frac{\left(x-4\sqrt{d} - 32d\sqrt{T}\ln T - 16\sigma\sqrt{\eta dT}\ln T\right)_+^2}{512\sigma^2dT\ln^2T}\right) \\
    & \quad\quad + 2d(T/d)^{2d+1}\exp\left(-\frac{(x-4\sqrt{d})_+\sqrt{\eta}}{8\sigma\sqrt{dT}\ln T}\right).
\end{align*}
\end{theorem}

The main technical challenge to prove Theorem \ref{thm:linear} is that the analysis for the MAB setting is not directly applicable --- the estimation of the unknown parameter $\theta$ is entangled with uncontrollable arm feature vectors in the linear bandit setting. It is also not straightforward how the equation \eqref{eq:split} can be adapted to accommodate the linear bandit setting and prevent the appearance of $K$ (the number of arms may even be infinite). The main idea to resolve these two obstacles is by noticing that $a_t^\top V_{t-1}^{-1}a_t$ can be regarded as a counterpart of $n_i$ in the MAB setting (though they are not equivalent). The tail bound is obtained by further adapting and refining the analysis from Theorem \ref{thm:exp-K-armed-optimal}. All proof details are provided in the online appendix.

\smallskip

Some interpretation and remarks for Theorem \ref{thm:linear} are as follows.
\begin{enumerate}
    \item To have a more interpretable representation of the upper bound derived in Theorem \ref{thm:linear}, we can do a similar change-of-variable trick as in Theorem \ref{thm:exp-K-armed}. If we denote a variable $y$ as
    \begin{align*}
        y = \frac{\left(x - 4\sqrt{d} - 32d\sqrt{T}\ln T - 16\sigma\sqrt{\eta\vee1/\eta}d\sqrt{dT}\ln^2 T\right)_+}{32\sigma \sqrt{dT}\ln T},
    \end{align*}
    then for any $x\geq 0$, we have
    \begin{align*}
        \sup_{\theta}\mbP(R_\theta^\pi(T)\geq x) \leq 8d(T/d)^{2d+1}\exp\left(-y^2\wedge y\sqrt{\eta}\right) \leq 8T^{3d}\exp\left(-y^2\wedge y\sqrt{\eta}\right).
    \end{align*}
    One can observe that for any $\eta > 0$, our policy always yields a $\tilde O(\sqrt{T})$ expected regret, achieving the optimal order. In fact, notice that for any $x > 0$
    \begin{align*}
        \mbE[\hat R_\theta^\pi(T)] = \mbE[R_\theta^\pi(T)] \leq x + \mbP(R_\theta^\pi(T)\geq x)\cdot \sqrt{d}T.
    \end{align*}
    If we let
    \begin{align*}
        x = 4\sqrt{d} + 32d\sqrt{T}\ln T + C\sigma\sqrt{\eta\vee1/\eta}d\sqrt{dT}\ln^2 T
    \end{align*}
    with the absolute constant $C$ being moderately large, then $\mbP(R_\theta^\pi(T)\geq x)\cdot T = O(1)$. As a result, the worst-case expected regret becomes controlled by the order of
    \begin{align*}
        O\left(d\sqrt{T}\ln T + \sigma\sqrt{\eta\vee1/\eta}d\sqrt{dT}\ln^2 T\right).
    \end{align*}
    We have to point out that compared to the $O(d\sqrt{T}\ln T)$ regret in previous linear bandits literature (see, e.g., \citealt{abbasi2011improved}), our result has an additional factor of $\sqrt{d}\ln T$. Even though this additional factor does not affect the optimal $\tilde O(\sqrt{T})$ expected regret order achieved by our policy design on the linear bandit setting, we think as future work it might be useful to see whether our analysis can be refined to improve on the additional factor.
    
    \item Regarding computation, the main step is Line 4 in Algorithm \ref{alg:UCB-lin}, where the objective function is a convex function of $a$. When $|\cA_t|$ is finite and small (e.g., $|\cA_t|=O(\poly(d))$), we can enumerate all choices for $a$. However, in general, even when $\cA_t$ is a convex set, maximizing a convex function can be NP-Hard. A similar discussion can be found in \cite{dani2008stochastic}, and in the future work it might be interesting to see whether the maximization problem is efficiently solvable under other special cases.
\end{enumerate}

\section{Conclusion}

\label{sec:conclusion}

In this work, we consider the stochastic multi-armed bandit problem with a joint goal of minimizing the worst-case expected regret and obtaining light-tailed probability bound of the regret distribution. We demonstrate that light-tailed risk and instance-dependent consistency are incompatible, and show that light-tailed risk and worst-case optimality can co-exist through a simple new policy design. We also discuss generalizations of our results and show how to achieve the optimal rate dependence on both the number of arms $K$ and the time horizon $T$ with or without knowing $T$. We extend the simple and optimal policy design to the linear bandit setting.

There are several prospective future directions. Technically, one direction is to improve our policy design for linear bandits on the tail bound and the computation efficiency. Another direction is to study extensions of our results to general heavy-tailed bandit problems, where a robust mean estimator (rather than the simple empirical mean) for each arm is essential. Meanwhile, it would be interesting to understand the cost of embracing both worst-case optimality and light-tailed regret distribution in a more precise way, especially with regard to the constant term in front of the expected regret bounds. Empirically, it would be interesting to see how the policy design works in various practical settings. Methodologically, it is tempting to see whether our policy design can be integrated into more complex settings such as reinforcement learning.

One interesting future work from both the theoretical and numerical sides is to study the relations among optimality, consistency, and tail risk. From Theorem \ref{thm:consistent-vs-light-tail}, we know that one cannot hope for light-tailed risk and instance-dependent consistency simultaneously. This suggests that there might exist some ``Pareto frontier'' where the optimal regret tail risk can be dependent on to what extent we allow consistency. This requires further generalizing our lower bound results as well as policy designs to handle the more sophisticated situation where worst-case optimality and instance-dependent consistency are entangled with each other under varying regret thresholds. Nevertheless, we hope our results and analysis in this paper may bring about new insights on understanding and alleviating the tail risk of learning algorithms under a stochastic environment.

\ACKNOWLEDGMENT{The authors would like to gratefully acknowledge the Editor and the Reviewers for their time and comments that greatly helped improve the manuscript. The work of David Simchi-Levi and Feng Zhu is partially supported by the MIT Data Science Laboratory. The authors are listed in alphabetical order.}

\bibliographystyle{informs2014}
\bibliography{main}

\newpage
\setcounter{page}{1}
\begin{APPENDICES}
\section*{Online Appendix}

To prove Theorem \ref{thm:consistent-vs-light-tail}, we need the following lemma. 

\begin{lemma} \label{lemma:sub-linear}
Consider the two-armed bandit problem with $\sigma$-Gaussian noise. Let $\pi$ be a policy such that for any true mean vector $\theta$, 
\begin{align*}
    \limsup_{T\to+\infty}\frac{\mathbb E\left[\hat R_\theta^\pi(T)\right]}{T} = 0.
\end{align*}
That is, the expected regret under $\pi$ is always sub-linear in $T$. Then for any $\tilde\theta = (\tilde\theta_1, \tilde\theta_2)$ where $\tilde\theta_1 > \tilde\theta_2$, and any $\delta > 0$, we have
\begin{align*}
    \limsup_{T\to+\infty}\mbP_{\tilde\theta}^\pi(|\hat\theta_{T, 2} - \tilde\theta_2| > \delta) = 0.
\end{align*}
\end{lemma}

{\noindent\bf Proof of Lemma \ref{lemma:sub-linear}.} Define
\begin{align*}
    E_T = \left\{|\hat\theta_{T, 2} - \tilde\theta_2|\leq \delta\right\}.
\end{align*}
Fix any positive integer $N$, we have
\begin{align*}
\mathbb P_{\tilde\theta}^\pi(\bar E_T) 
& = \mathbb P_{\tilde\theta}^\pi(\bar E_T; n_{T, 2} < N) +  \mathbb P_{\tilde\theta}^\pi(\bar E_T; n_{T, 2} \geq N) \\
& \leq \mathbb P_{\tilde\theta}^\pi(n_{T, 2} < N) + \sum_{n=N}^{+\infty} \mathbb P_{\tilde\theta}^\pi(\bar E_T; n_{T, 2} = n) \\
& \leq \mathbb P_{\tilde\theta}^\pi(n_{T, 2} < N) + \sum_{n=N}^{+\infty}2\exp(-\frac{n\delta^2}{2\sigma^2}).
\end{align*}
Note that the last inequality holds because
\begin{align*}
    & \quad \mathbb P_{\tilde\theta}^\pi(\bar E_T; n_{T, 2} = n) \\
    & = \mathbb P_{\tilde\theta}^\pi\left(
    \left|\frac{\sum_{n'=1}^{n_{T, 2}}r_{t_2(n'), 2}}{n_{T, 2}} - \tilde\theta_2\right|>\delta; n_{T, 2}=n\right) \\
    & = \mathbb P_{\tilde\theta}^\pi\left(
    \left|\frac{\sum_{n'=1}^{n_{T, 2}}\epsilon_{t_2(n'), 2}}{n_{T, 2}}\right|>\delta; n_{T, 2}=n\right) \\
    & \leq \mathbb P\left(
    \left|\frac{\sum_{n'=1}^{n}\epsilon_{t_2(n'), 2}}{n}\right|>\delta\right) \\
    & \leq 2\exp(-\frac{n\delta^2}{2\sigma^2}).
\end{align*}
Thus,
\begin{align*}
\limsup_{T}\mathbb P_{\tilde\theta}^\pi(\bar E_T) \leq \limsup_{T}\mathbb P_{\tilde\theta}^\pi(n_{T, 2} < N) + \sum_{n=N}^{+\infty}2\exp(-\frac{n\delta^2}{2\sigma^2})
\end{align*}
holds for any $N$. Note that the last term converges to $0$ as $N\to+\infty$. It suffices to show $\mathbb P_{\tilde\theta}^\pi(n_{T, 2} < N)\to 0$ as $T\to+\infty$ for any fixed $N$. Suppose this does not hold, then we can find $p > 0$ and a sequence $\left\{T(m)\right\}_{m=1}^{+\infty}$ such that
$$
\mathbb P_{\tilde\theta}^\pi(n_{T(m), 2} < N)>p.
$$
Let $M$ be some large number such that $q\triangleq p - N\exp(-\frac{M^2}{2\sigma^2}) > 0$. Consider an alternative environment $\theta = (\theta_1, \theta_2)$ where $\theta_2 > \theta_1 = \tilde\theta_1$. Using the change-of-measure argument, we have
\begin{align*}
& \quad\ \mathbb P_{\theta}^\pi(n_{T(m), 2} < N) \\
& = \mathbb E_{\theta}^\pi[\mathds 1\{n_{T(m), 2} < N\}] \\
& = \mbE_{\tilde\theta}^\pi\left[\exp\left(\sum_{n=1}^{n_{T(m), 2}}\frac{(r_{t_2(n), 2} - \tilde\theta_2)^2 - (r_{t_2(n), 2} - \theta_2)^2}{2\sigma^2}\right)\mathds 1\{n_{T(m), 2} < N\}\right] \\
& = \mathbb E_{\tilde\theta}^\pi\left[\exp\left(n_{T(m), 2}\left(\frac{\tilde\theta_2^2 - \theta_2^2}{2\sigma^2} + \frac{(\theta_2 - \tilde\theta_2)\hat\theta_{T(m), 2}}{\sigma^2}\right)\right)\mathds 1\{n_{T(m), 2} < N\}\right] \\
& \geq \mathbb E_{\tilde\theta}^\pi\left[\exp\left(n_{T(m), 2}\left(\frac{\tilde\theta_2^2 - \theta_2^2}{2\sigma^2} + \frac{(\theta_2 - \tilde\theta_2)\hat\theta_{T(m), 2}}{\sigma^2}\right)\right)1\{\hat\theta_{T(m), 2} > \tilde\theta_2 - M, n_{T(m), 2} < N\}\right] \\
& \geq \mathbb E_{\tilde\theta}^\pi\left[\exp\left(N\left(-\frac{(\tilde\theta_2 - \theta_2)^2}{2\sigma^2} - \frac{M(\theta_2 - \tilde\theta_2)}{\sigma^2}\right)\right)1\{\hat\theta_{T(m), 2} > \tilde\theta_2 - M, n_{T(m), 2} < N\}\right] \\
& = \exp\left(N\left(-\frac{(\tilde\theta_2 - \theta_2)^2}{2\sigma^2} - \frac{M(\theta_2 - \tilde\theta_2)}{\sigma^2}\right)\right)\mathbb P_{\tilde\theta}^\pi(\hat\theta_{T(m), 2} > \tilde\theta_2 - M, n_{T(m), 2} < N).
\end{align*}
Note that
\begin{align*}
& \quad\ \mathbb P_{\tilde\theta}^\pi(\hat\theta_{T(m), 2} > \tilde\theta_2 - M, n_{T(m), 2} < N) \\
& = p - \sum_{n=1}^{N-1}\mathbb P_{\tilde\theta}^\pi(\hat\theta_{T(m), 2} \leq \tilde\theta_2 - M, n_{T(m), 2} = n) \\
& \geq p - \sum_{n=1}^{N-1}\exp(-\frac{nM^2}{2\sigma^2}) \geq p - N\exp(-\frac{M^2}{2\sigma^2})=q>0.
\end{align*}
Therefore, there exists a constant positive probability such that $\pi$ pulls arm $2$ no more than $N$ times under $\theta$. As a result, $\pi$ incurs a linear expected regret under $\theta$, leading to a contradiction.

$\hfill\Box$

{\noindent\bf Proof of Theorem \ref{thm:consistent-vs-light-tail}.} 

To prove the first statement, we consider the environment where the noise $\epsilon$ is Gaussian with standard deviation $\sigma$. Let $\theta_1 = 1/2$. Let $\theta = (\theta_1, \theta_2)$ and $\tilde\theta = (\theta_1, \tilde\theta_2)$, where $\theta_2 = \theta_1 + \frac{1}{2}$ and $\tilde\theta_2 = \theta_1 - \frac{1}{2}$. Let $c'\in(c, 1/2)$. Define
\begin{align*}
    E_T = \left\{|\hat\theta_{T, 2} - \tilde\theta_2|\leq \delta\right\}
\end{align*}
where $\delta>0$ is a small number, and
\begin{align*}
    F_T = \{n_2\leq f(T)\}.
\end{align*}
Here, $f(T) > 0$ is a strictly increasing function such that
\begin{align*}
    \limsup_T\frac{f(T)}{T} < 1-2c'.
\end{align*}
We will detail how $f(T)$ should be chosen under different conditions in the last step of the proof. Then there exists $T_0$ such that $f(T) < (1-2c') T$ for any $T > T_0$. Now we fix any $T > T_0$. Under the environment $\tilde\theta$, we have
\begin{align*}
    \mbP_{\tilde\theta}^\pi(\bar F_T) & = \mbP_{\tilde\theta}^\pi\left(n_2 > f(T)\right) \leq \frac{\mbE_{\tilde\theta}^\pi[n_2]}{f(T)} \leq \frac{2\mbE[R_{\tilde \theta}^\pi(T)]}{f(T)} = \frac{2\mbE[\hat R_{\tilde \theta}^\pi(T)]}{f(T)}.
\end{align*}
Combined with Lemma \ref{lemma:sub-linear}, we have
\begin{align} \label{eq:bar-F}
    \liminf_T \mbP_{\tilde\theta}^\pi(E_T, F_T) \geq 1 - \limsup_T \frac{2\mbE[\hat R_{\tilde\theta}^\pi(T)]}{f(T)}.
\end{align}
Notice that
\begin{align*}
    & \quad \mbP\left(\hat R_{\theta}^\pi(T)\geq cT\right) \\
    & \geq \mbP\left(R_{\theta}^\pi(T)\geq c'T, -N^\pi(T) \geq -(c'-c)T\right) \\
    & = \mbP\left(R_{\theta}^\pi(T)\geq c'T\right) - \mbP\left(R_{\theta}^\pi(T)\geq c'T, N^\pi(T) > (c'-c)T\right) \\
    & \geq \mbP\left(R_{\theta}^\pi(T)\geq c'T\right) - \mbP\left(N^\pi(T) > (c'-c)T\right) \\
    & \geq \mbP\left(R_{\theta}^\pi(T)\geq c'T\right) - \exp\left(-\frac{(c'-c)^2T}{2\sigma^2}\right)
\end{align*}
The last inequality holds from Lemma \ref{lemma:bound-noise}. Now
\begin{align*}
    & \quad \mbP\left(R_{\theta}^\pi(T)\geq c'T\right) \\
    & \geq \mbP_{\theta}^\pi(n_1\geq 2c'T) \\
    & \geq \mbP_{\theta}^\pi(n_2\leq (1-2c')T) \\
    & \geq \mbP_{\theta}^\pi(n_2\leq f(T)) \\
    & \geq \mbP_{\theta}^\pi(E_T, F_T) \\
    & = \mbE_{\theta}^\pi[\mathds 1\{E_T F_T\}]
\end{align*}
Now we apply the change-of-measure argument (see also, e.g., (13) or (51) in \citealt{fan2021fragility}; note that they fix arm $2$ while we fix arm $1$).
\begin{align*}
    & \quad \mbE_{\theta}^\pi[\mathds 1\{E_T F_T\}] \\
    & = \mbE_{\tilde\theta}^\pi\left[\exp\left(\sum_{n=1}^{n_2}\frac{(r_{t_2(n), 2} - \tilde\theta_2)^2}{2\sigma^2} - \frac{(r_{t_2(n), 2} - \theta_2)^2}{2\sigma^2}\right)\mathds 1\{E_T F_T\}\right] \\
    & = \mbE_{\tilde\theta}^\pi\left[\exp\left(n_2\left(\frac{\tilde\theta_2^2 - \theta_2^2}{2\sigma^2} + \frac{(\theta_2 - \tilde\theta_2)\hat\theta_{T, 2}}{\sigma^2}\right)\right)\mathds 1\{E_T F_T\}\right] \\
    & \geq \mbE_{\tilde\theta}^\pi\left[\exp\left(n_2\left(\frac{\tilde\theta_2^2 - \theta_2^2}{2\sigma^2} + \frac{(\theta_2 - \tilde\theta_2)(\tilde\theta_{2} - \delta)}{\sigma^2}\right)\right)\mathds 1\{E_T F_T\}\right] \\
    & = \mbE_{\tilde\theta}^\pi\left[\exp\left(n_2\left(-\frac{(\tilde\theta_2 - \theta_2)^2}{2\sigma^2} - \frac{\delta(\theta_2 - \tilde\theta_2)}{\sigma^2}\right)\right)\mathds 1\{E_T F_T\}\right] \\
    & \geq \mbE_{\tilde\theta}^\pi\left[\exp\left(f(T)\left(-\frac{(\tilde\theta_2 - \theta_2)^2}{2\sigma^2} - \frac{\delta(\theta_2 - \tilde\theta_2)}{\sigma^2}\right)\right)\mathds 1\{E_T F_T\}\right] \\
    & = \exp(-f(T)(1/(2\sigma^2) + \delta/\sigma^2))\mbP_{\tilde\theta}^\pi(E_T, F_T).
\end{align*}
Therefore, 
\begin{align}
    & \quad \liminf_T\frac{\ln\left\{\sup_{\theta'}\mbP\left(\hat R_{\theta'}^\pi(T)\geq cT\right)\right\}}{f(T)} \nonumber\\
    & \geq \liminf_T \frac{\ln \left\{\exp(-f(T)(1/(2\sigma^2) + \delta/\sigma^2))\mbP_{\tilde\theta}^\pi(E_T, F_T) - \exp\left(-\frac{(c'-c)^2T}{2\sigma^2}\right) \right\}}{f(T)}.
    \label{eq:ln-P}
\end{align}
Now assume that $\pi$ is consistent. Then we set $f(T) = T^\beta$ with $\beta\in (0, 1)$. From \eqref{eq:bar-F}, we have
\begin{align*}
    \liminf_T \mbP_{\tilde\theta}^\pi(E_T, F_T) \geq 1 - \limsup_T \frac{2\mbE[\hat R_{\tilde\theta}^\pi(T)]}{T^\beta} = 1.
\end{align*}
Then from \eqref{eq:ln-P}, we have
\begin{align*}
    \liminf_T\frac{\ln\left\{\sup_{\theta'} \mbP(\hat R_{\theta'}^\pi(T)\geq cT)\right\}}{T^\beta} \geq -(1/(2\sigma^2) + \delta/\sigma^2).
\end{align*}
Since $\delta > 0$ is arbitrary, we have
\begin{align*}
    & \quad \liminf_T\frac{\ln\left\{\sup_{\theta'}\mbP(\hat R_{\theta'}^\pi(T)\geq cT)\right\}}{T^\beta} \geq -1/(2\sigma^2).
\end{align*}
Note again that $\beta > 0$ is arbitrary. Now let $0<\beta' < \beta$, we have
\begin{align*}
    \liminf_T\frac{\ln \left\{\sup_{\theta'}\mbP(\hat R_{\theta'}^\pi(T)\geq  cT)\right\}}{T^\beta} & = \liminf_T\frac{\ln \left\{\sup_{\theta'}\mbP(\hat R_{\theta'}^\pi(T)\geq  cT)\right\}}{T^{\beta'}} \cdot \liminf_T T^{\beta' - \beta} \geq -1/(2\sigma^2) \cdot 0 = 0.
\end{align*}

To prove \eqref{eq:heavy-tail-ln}, we consider two environments. The first environment is $\theta = (1/2, 1)$ where the noise $\epsilon$ is Gaussian with standard deviation $\sigma$ for the first arm and $\sigma_0 > \sigma$ for the second arm. The second environment is $\tilde\theta=(1/2, 0)$ where the noise $\epsilon$ is Gaussian with standard deviation $\sigma$. To simplify notations, we will write $\hat R_{\theta}^\pi(T)$ instead of $\hat R_{\theta, \sigma_0}^\pi(T)$ and $R_{\theta}^\pi(T)$ instead of $R_{\theta, \sigma_0}^\pi(T)$. One need to keep in mind that under the environment with mean vector $\theta$, the second arm has a larger deviation. We assume that $\pi$ satisfies
\begin{align*}
    \limsup_T \frac{\mbE[\hat R_{\tilde\theta}^\pi(T)]}{\ln T} = c_\pi\sigma^2 < +\infty.
\end{align*}
Let $c'\in(c, 1/2)$. Define
\begin{align*}
    E_T = \left\{|\hat\theta_{T, 2} - \tilde\theta_2|\leq \delta\right\}
\end{align*}
where $\delta>0$ is a small number, and
\begin{align*}
    F_T = \{n_2\leq  4c_\pi\sigma^2\ln T\}.
\end{align*}
Then there exists $T_0$ such that $4c_\pi\sigma^2\ln T < (1-2c') T$ for any $T > T_0$. Now we fix any $T > T_0$. Under the environment $\tilde\theta$, we have
\begin{align*}
    \mbP_{\tilde\theta}^\pi(\bar F_T) & = \mbP_{\tilde\theta}^\pi\left(n_2 >  4c_\pi\sigma^2\ln T\right) \leq \frac{\mbE_{\tilde\theta}^\pi[n_2]}{4c_\pi\sigma^2\ln T} \leq \frac{2\mbE[R_{\tilde \theta}^\pi(T)]}{ 4c_\pi\sigma^2\ln T} = \frac{2\mbE[\hat R_{\tilde \theta}^\pi(T)]}{ 4c_\pi\sigma^2\ln T}.
\end{align*}
Combined with Lemma \ref{lemma:sub-linear}, we have
\begin{align*} 
    \liminf_T \mbP_{\tilde\theta}^\pi(E_T, F_T) \geq 1 - \limsup_T \frac{2\mbE[\hat R_{\tilde\theta}^\pi(T)]}{ 4c_\pi\sigma^2\ln T}\geq 1/2.
\end{align*}
Notice that
\begin{align*}
    & \quad \mbP\left(\hat R_{\theta}^\pi(T)\geq cT\right) \\
    & \geq \mbP\left(R_{\theta}^\pi(T)\geq c'T, -N^\pi(T) \geq -(c'-c)T\right) \\
    & = \mbP\left(R_{\theta}^\pi(T)\geq c'T\right) - \mbP\left(R_{\theta}^\pi(T)\geq c'T, N^\pi(T) > (c'-c)T\right) \\
    & \geq \mbP\left(R_{\theta}^\pi(T)\geq c'T\right) - \mbP\left(N^\pi(T) > (c'-c)T\right) \\
    & \geq \mbP\left(R_{\theta}^\pi(T)\geq c'T\right) - \exp\left(-\frac{(c'-c)^2T}{2\sigma^2}\right)
\end{align*}
The last inequality holds from Lemma \ref{lemma:bound-noise}. Now
\begin{align*}
    & \quad \mbP\left(R_{\theta}^\pi(T)\geq c'T\right) \\
    & \geq \mbP_{\theta}^\pi(n_1\geq 2c'T) \\
    & \geq \mbP_{\theta}^\pi(n_2\leq (1-2c')T) \\
    & \geq \mbP_{\theta}^\pi(n_2\leq  4c_\pi\sigma^2\ln T) \\
    & \geq \mbP_{\theta}^\pi(E_T, F_T) \\
    & = \mbE_{\theta}^\pi[\mathds 1\{E_T F_T\}]
\end{align*}
Now we apply the change-of-measure argument.
\begin{align*}
    & \quad \mbE_{\theta}^\pi[\mathds 1\{E_T F_T\}] \\
    & = \mbE_{\tilde\theta}^\pi\left[\exp\left(\sum_{n=1}^{n_2}\frac{(r_{t_2(n), 2} - \tilde\theta_2)^2}{2\sigma^2} - \frac{(r_{t_2(n), 2} - \theta_2)^2}{2\sigma_0^2}\right)\mathds 1\{E_T F_T\}\right] \\
    & \geq \mbE_{\tilde\theta}^\pi\left[\exp\left(\sum_{n=1}^{n_2}\frac{(r_{t_2(n), 2} - \tilde\theta_2)^2 - (r_{t_2(n), 2} - \theta_2)^2}{2\sigma_0^2}\right)\mathds 1\{E_T F_T\}\right] \\
    & = \mbE_{\tilde\theta}^\pi\left[\exp\left(n_2\left(\frac{\tilde\theta_2^2 - \theta_2^2}{2\sigma_0^2} + \frac{(\theta_2 - \tilde\theta_2)\hat\theta_{T, 2}}{\sigma_0^2}\right)\right)\mathds 1\{E_T F_T\}\right] \\
    & \geq \mbE_{\tilde\theta}^\pi\left[\exp\left(n_2\left(\frac{\tilde\theta_2^2 - \theta_2^2}{2\sigma_0^2} + \frac{(\theta_2 - \tilde\theta_2)(\tilde\theta_{2} - \delta)}{\sigma_0^2}\right)\right)\mathds 1\{E_T F_T\}\right] \\
    & = \mbE_{\tilde\theta}^\pi\left[\exp\left(n_2\left(-\frac{(\tilde\theta_2 - \theta_2)^2}{2\sigma_0^2} - \frac{\delta(\theta_2 - \tilde\theta_2)}{\sigma_0^2}\right)\right)\mathds 1\{E_T F_T\}\right] \\
    & \geq \mbE_{\tilde\theta}^\pi\left[\exp\left(4c_\pi\sigma^2\ln T\left(-\frac{(\tilde\theta_2 - \theta_2)^2}{2\sigma_0^2} - \frac{\delta(\theta_2 - \tilde\theta_2)}{\sigma_0^2}\right)\right)\mathds 1\{E_T F_T\}\right] \\
    & = \exp(-4c_\pi\sigma^2\ln T(1/(2\sigma_0^2) + \delta/\sigma_0^2))\mbP_{\tilde\theta}^\pi(E_T, F_T).
\end{align*}
Therefore, 
\begin{align*}
    & \quad \liminf_T\frac{\ln\left\{\sup_{\theta'}\mbP\left(\hat R_{\theta', \sigma_0}^\pi(T)\geq cT\right)\right\}}{\ln T} \nonumber\\
    & \geq \liminf_T \frac{\ln \left\{\exp(-4c_\pi\sigma^2\ln T(1/(2\sigma_0^2) + \delta/\sigma_0^2))\mbP_{\tilde\theta}^\pi(E_T, F_T) - \exp\left(-\frac{(c'-c)^2T}{2\sigma^2}\right) \right\}}{\ln T} \\
    & \geq -(2c_\pi\sigma^2/\sigma_0^2 + 4c_\pi\delta\sigma^2/\sigma_0^2).
\end{align*}
Since $\delta > 0$ is arbitrary, we have
\begin{align*}
    & \quad \liminf_T\frac{\ln\left\{\sup_{\theta'}\mbP(\hat R_{\theta', \sigma_0}^\pi(T)\geq cT)\right\}}{\ln T} \geq -2c_\pi\sigma^2/\sigma_0^2.
\end{align*}
Let $C_\pi = 2c_\pi$, we have
\begin{align*}
    \liminf_T\frac{\ln\left\{\sup_{\theta'}\mbP(\hat R_{\theta', \sigma_0}^\pi(T)\geq cT)\right\}}{\ln T} \geq -C_\pi\frac{\sigma^2}{\sigma_0^2}.
\end{align*}

$\hfill\Box$

{\noindent\bf Proof of Theorem \ref{thm:exp-2-armed}.}
Without loss of generality, we assume $\theta_1 > \theta_2$. We prove the results one by one. Since the environment $\theta$ is fixed, we will write $\mbP$ ($\mbE$) instead of $\mbP^\pi_\theta$ ($\mbE_\theta^\pi$).
\begin{enumerate}
\item From Lemma \ref{lemma:bound-noise}, 
\begin{align*}
    \mbE[\hat R_\theta^\pi(T)] = \mbE[R_\theta^\pi(T)] = \mbE[n_2]\cdot\Delta_2.
\end{align*}
Let $G$ be the event such that
\begin{align*}
    G = \{\theta_k\in \CI_{t, k},\ \forall (t, k)\}.
\end{align*}
Then
\begin{align*}
    \mbP(\bar G) \leq \sum_{(t, k)}\mbP(\theta_k\notin\CI_{t, k}) \leq 2\sum_{n=1}^T2\exp(-2\frac{\eta T\ln T}{n}) \leq 4T^{1-2\eta}.
\end{align*}
Thus, 
\begin{align*}
    \mbE[n_2] & = \mbE[n_2|G]\mbP(G) + \mbE[n_2|\bar G]\mbP(\bar G) \\
    & \leq \mbE[n_2|G] + T\cdot\mbP(\bar G) \\
    & \leq \mbE[n_2|G] + 4T^{2-2\eta} \\
    & \leq \mbE[n_2|G] + 4.
\end{align*}
With a slight abuse of notation, we let $t$ be the largest time period such that arm $2$ is pulled but subsequently not eliminated from $\cA$. Then under $G$, we have
\begin{align*}
    \theta_1 - 2\sigma\frac{\sqrt{\eta T\ln T}}{n_{t, 2} - 1} & \leq \theta_1 - 2\sigma\frac{\sqrt{\eta T\ln T}}{n_{t, 1}} \leq \hat\theta_{t, 1} - \sigma\frac{\sqrt{\eta T\ln T}}{n_{t, 1}} \leq \hat\theta_{t, 2} + \sigma\frac{\sqrt{\eta T\ln T}}{n_{t, 2} - 1} \leq \theta_2 + 2\sigma\frac{\sqrt{\eta T\ln T}}{n_{t, 2}-1}.
\end{align*}
Therefore, 
\begin{align*}
    n_{t, 2} \leq 1 + 4\sigma\frac{\sqrt{\eta T\ln T}}{\Delta_2}
\end{align*}
and thus, 
\begin{align*}
    n_2 \leq 2 + 4\sigma\frac{\sqrt{\eta T\ln T}}{\Delta_2}.
\end{align*}
As a result, 
\begin{align*}
    \mbE[R_\theta^\pi(T)] \leq 2\Delta_2 + 4\sigma\sqrt{\eta T\ln T} + 4 = O(\sqrt{T\ln T}).
\end{align*}

\item We have
\begin{align*}
    \mbP(\hat R_{\theta}^\pi(T) \geq cT^\alpha) & \leq \mbP(R_\theta^\pi(T) \geq cT^\alpha/2) + \mbP(N_\theta^\pi(T) \leq -cT^\alpha/2)
\end{align*}
From Lemma \ref{lemma:bound-noise}, the second term can be bounded as
\begin{align} \label{eq:bound-noise}
    \mbP(N_\theta^\pi(T) \leq -cT^\alpha/2) \leq \exp\left(-\frac{c^2T^{2\alpha}}{2\sigma^2T}\right) = \exp\left(-\frac{c^2T^{2\alpha-1}}{2\sigma^2}\right).
\end{align}

We are left to bound $\mbP(R_\theta^\pi(T) \geq cT^\alpha/2)$. Let $S$ be the event defined as
    \begin{align*}
        S = \{\text{Arm }1\text{ is never eliminated throughout the whole time horizon}\}.
    \end{align*}
    Then
    \begin{align*}
        \bar S = \{\exists t\text{ such that arm $1$ is eliminated at time }t\}.
    \end{align*}
    So
    \begin{align*}
        \mbP(R_\theta^\pi(T) \geq cT^\alpha/2) = \mbP(R_\theta^\pi(T) \geq cT^\alpha/2, S) + \mbP(R_\theta^\pi(T) \geq cT^\alpha/2, \bar S).
    \end{align*}
    Let $T$ be such that
    \begin{align*}
        cT^\alpha \geq \max\{4, 16\sigma\sqrt{\eta T\ln T}\}.
    \end{align*}
    Let $n_0 = \lceil cT^\alpha/(2\Delta_2)\rceil$ - 1, then
    \begin{align*}
        n_0 \geq cT^\alpha/(4\Delta_2).
    \end{align*}
    Also, if $R_\theta^\pi(T) \geq cT^\alpha/2$, then since $R_\theta^\pi(T)=n_2\Delta_2\leq T\Delta_2$, we must have
    \begin{align*}
        T \geq cT^{\alpha}/(2\Delta_2),
    \end{align*}
    which means $\Delta_2\geq cT^{\alpha-1}/2$. We have
    \begin{align}
        & \quad \mbP(R_\theta^\pi(T) \geq cT^\alpha/2, S) \nonumber\\
        &  = \mbP(n_2\Delta_2 \geq cT^\alpha/2, S) \nonumber\\
        & = \mbP(n_2 \geq cT^\alpha/(2\Delta_2), S) \nonumber\\
        & \leq \mbP(n_2\geq n_0+1,\text{ arm }1\text{ and }2\text{ are pulled in turn for }(n_0+1)\text{ times}) \nonumber\\
        & \leq \mbP(\text{arm }1\text{ and }2\text{ are pulled in turn for }n_0\text{ times and arm }1\text{ and }2\text{ are both not eliminated}) \nonumber\\
        & \leq \mbP\left(\hat\theta_{t_1(n_0), 1} - \frac{\sigma\sqrt{\eta T\ln T}}{n_0}\leq \hat\theta_{t_2(n_0), 2} + \frac{\sigma\sqrt{\eta T\ln T}}{n_0}\right) \quad\quad\text{(otherwise arm $2$ is eliminated)}\nonumber\\
        & = \mbP\left(\theta_1 - \frac{\sum_{m=1}^{n_0}\epsilon_{t_1(m), 1} + \sigma\sqrt{\eta T\ln T}}{n_0}\leq \theta_2 + \frac{\sum_{m=1}^{n_0}\epsilon_{t_2(m), 2} + \sigma\sqrt{\eta T\ln T}}{n_0}\right) \nonumber\\
        & = \mbP\left(\frac{\sum_{m=1}^{n_0}(\epsilon_{t_1(m), 1} - \epsilon_{t_2(m), 2})}{n_0} \geq \Delta_2 - \frac{2\sigma\sqrt{\eta T\ln T}}{n_0}\right) \nonumber\\
        & \leq \mbP\left(\frac{\sum_{m=1}^{n_0}\epsilon_{t_1(m), 1}}{n_0} \geq \frac{\Delta_2}{2} - \frac{\sigma\sqrt{\eta T\ln T}}{n_0}\right) + \mbP\left(\frac{\sum_{m=1}^{n_0}\epsilon_{t_2(m), 2}}{n_0} \geq \frac{\Delta_2}{2} - \frac{\sigma\sqrt{\eta T\ln T}}{n_0}\right) \nonumber\\
        & \leq 2\exp\left(-n_0\left(\frac{\Delta_2}{2} - \frac{\sigma\sqrt{\eta T\ln T}}{n_0}\right)^2\big/(2\sigma^2)\right) \nonumber\\
        & = 2\exp\left(-n_0\Delta_2^2\left(1 - \frac{8\sigma\sqrt{\eta T\ln T}}{cT^\alpha}\right)^2\big/(2\sigma^2)\right) \nonumber\\
        & \leq 2\exp\left(-\frac{n_0\Delta_2^2}{8\sigma^2}\right) \nonumber\\
        & \leq 2\exp\left(-\frac{cT^{\alpha}\cdot cT^{\alpha-1}}{128\sigma^2}\right) \nonumber\\
        & = 2\exp\left(-\frac{c^2T^{2\alpha-1}}{128\sigma^2}\right). \label{eq:bound-pseudo-identified}
    \end{align}
    Meanwhile, 
    \begin{align}
        & \quad \mbP(R_\theta^\pi(T) \geq cT^\alpha/2, \bar S) \nonumber\\
        & \leq  \mbP\left(\exists n\leq T/2: \hat\theta_{t_1(n), 1} + \frac{\sigma\sqrt{\eta T\ln T}}{n} < \hat\theta_{t_2(n), 2} - \frac{\sigma\sqrt{\eta T\ln T}}{n}\right) \nonumber\\
        & = \mbP\left(\exists n\leq T/2: \theta_1 + \frac{\sum_{m=1}^{n}\epsilon_{t_1(m), 1} + \sigma\sqrt{\eta T\ln T}}{n} < \theta_2 + \frac{\sum_{m=1}^{n}\epsilon_{t_2(m), 2} - \sigma\sqrt{\eta T\ln T}}{n}\right) \nonumber\\
        & \leq \sum_{n=1}^{\lfloor T/2\rfloor}\mbP\left(\frac{\sum_{m=1}^{n}(\epsilon_{t_2(m), 2} - \epsilon_{t_1(m), 1})}{n} > \Delta_2 + \frac{2\sigma\sqrt{\eta T\ln T}}{n}\right) \nonumber\\
        & \leq \sum_{n=1}^{\lfloor T/2\rfloor}\left(\mbP\left(\frac{\sum_{m=1}^{n}\epsilon_{t_2(m), 2}}{n} > \frac{\Delta_2}{2} + \frac{\sigma\sqrt{\eta T\ln T}}{n}\right) + \mbP\left(\frac{\sum_{m=1}^{n}-\epsilon_{t_1(m), 1}}{n} > \frac{\Delta_2}{2} + \frac{\sigma\sqrt{\eta T\ln T}}{n}\right)\right) \nonumber\\
        & \leq 2\sum_{n=1}^{\lfloor T/2\rfloor}\exp\left(-n\left(\frac{\Delta_2}{2} + \frac{\sigma\sqrt{\eta T\ln T}}{n}\right)^2\big/(2\sigma^2)\right) \quad\quad\text{(Hoeffding's inequality for subGaussian noises)}\nonumber\\
        & \leq T\exp\left(-2n\Delta_2\frac{\sigma\sqrt{\eta T\ln T}}{n}\big/(2\sigma^2)\right) \nonumber\\
        & \leq T\exp(-\sigma\cdot cT^{\alpha-1}\cdot\sqrt{\eta T\ln T}\big/ 2\sigma^2) \nonumber\\
        & = \exp\left(-\frac{cT^{\alpha-1/2}\sqrt{\eta\ln T} - \sigma\ln T}{2\sigma}\right) \nonumber\\
        & \leq \exp\left(-\frac{cT^{\alpha-1/2}\sqrt{\eta\ln T}}{4\sigma}\right) \nonumber\\
        & \leq \exp\left(-\frac{cT^{\alpha-1/2}}{16\sigma}\right). \label{eq:bound-pseudo-unidentified}
    \end{align}

Note that the equations above hold for any instance $\theta$. Combining (\ref{eq:bound-noise}), (\ref{eq:bound-pseudo-identified}), (\ref{eq:bound-pseudo-unidentified}) yields
\begin{align*}
    \sup_{\theta}\mbP(R_\theta^\pi(T)\geq cT^\alpha) \leq 4\exp\left(\frac{cT^{\alpha-1/2}}{16\sigma}\right).
\end{align*}
\end{enumerate}

$\hfill\Box$

Before we prove Theorem \ref{thm:tightness}, we need to introduce Lemma \ref{lemma:sub-linear-w}, the worst-case version of Lemma \ref{lemma:sub-linear}.

\begin{lemma} \label{lemma:sub-linear-w}
Consider the two-armed bandit problem with $\sigma$-Gaussian noise. Let $\pi$ be a policy such that
\begin{align*}
    \limsup_{T\to+\infty}\sup_{\theta}\frac{\sup_\theta\mathbb E\left[\hat R_\theta^\pi(T)\right]}{T} = 0.
\end{align*}
That is, the expected regret under $\pi$ is always sub-linear in $T$. Let $\omega\in[0, 1/2)$. Then for any $\delta > 0$, we have
\begin{align*}
    \limsup_{T\to+\infty}\sup_{\tilde\theta: \tilde\theta_1 > \tilde\theta_2}\mbP_{\tilde\theta}^\pi(|\hat\theta_{T, 2} - \tilde\theta_2| > \delta/n_{T, 2}^\omega) = 0.
\end{align*}
\end{lemma}

{\noindent\bf Proof of Lemma \ref{lemma:sub-linear-w}.} Define
\begin{align*}
    E_T = \left\{|\hat\theta_{T, 2} - \tilde\theta_2|\leq \delta/n_{T, 2}^\omega\right\}.
\end{align*}
Fix any positive integer $N$, we have
\begin{align*}
\mathbb P_{\tilde\theta}^\pi(\bar E_T) 
& = \mathbb P_{\tilde\theta}^\pi(\bar E_T; n_{T, 2} < N) +  \mathbb P_{\tilde\theta}^\pi(\bar E_T; n_{T, 2} \geq N) \\
& \leq \mathbb P_{\tilde\theta}^\pi(n_{T, 2} < N) + \sum_{n=N}^{+\infty} \mathbb P_{\tilde\theta}^\pi(\bar E_T; n_{T, 2} = n) \\
& \leq \mathbb P_{\tilde\theta}^\pi(n_{T, 2} < N) + \sum_{n=N}^{+\infty}2\exp(-\frac{n^{1-2\omega}\delta^2}{2\sigma^2}).
\end{align*}
Note that the last inequality holds because
\begin{align*}
    & \quad \mathbb P_{\tilde\theta}^\pi(\bar E_T; n_{T, 2} = n) \\
    & = \mathbb P_{\tilde\theta}^\pi\left(
    \left|\frac{\sum_{n'=1}^{n_{T, 2}}r_{t_2(n'), 2}}{n_{T, 2}} - \tilde\theta_2\right|>\delta/n_{T, 2}^\omega; n_{T, 2}=n\right) \\
    & = \mathbb P_{\tilde\theta}^\pi\left(
    \left|\frac{\sum_{n'=1}^{n_{T, 2}}\epsilon_{t_2(n'), 2}}{n_{T, 2}}\right|>\delta/n_{T, 2}^\omega; n_{T, 2}=n\right) \\
    & \leq \mathbb P\left(
    \left|\frac{\sum_{n'=1}^{n}\epsilon_{t_2(n'), 2}}{n}\right|>\delta/n^\omega\right) \\
    & \leq 2\exp(-\frac{n^{1-2\omega}\delta^2}{2\sigma^2}).
\end{align*}
Thus,
\begin{align*}
    \limsup_{T}\sup_{\tilde\theta}\mathbb P_{\tilde\theta}^\pi(\bar E_T) \leq \limsup_{T}\sup_{\tilde\theta}\mathbb P_{\tilde\theta}^\pi(n_{T, 2} < N) + \sum_{n=N}^{+\infty}2\exp(-\frac{n^{1-2\omega}\delta^2}{2\sigma^2})
\end{align*}
holds for any $N$. Note that the last term converges to $0$ as $N\to+\infty$. It suffices to show $\sup_{\tilde\theta}\mathbb P_{\tilde\theta}^\pi(n_{T, 2} < N)\to 0$ as $T\to+\infty$ for any fixed $N$. Suppose this does not hold, then we can find $p > 0$ and a sequence $\left\{T(m)\right\}_{m=1}^{+\infty}$ and a sequence of vectors $\{\tilde\theta(m)\}_{m=1}^{+\infty}$ such that
$$
\mathbb P_{\tilde\theta(m)}^\pi(n_{T(m), 2} < N)>p.
$$
Let $M$ be some large number such that $q\triangleq p - N\exp(-\frac{M^2}{2\sigma^2}) > 0$. Consider a sequence of alternative environments $\theta(m) = (\theta_1(m), \theta_2(m))$ where $\theta_2(m) > \theta_1(m) = \tilde\theta_1(m)$. Using the change-of-measure argument, we have
\begin{align*}
& \quad\ \mathbb P_{\theta(m)}^{\pi}(n_{T(m), 2} < N) \\
& = \mathbb E_{\theta(m)}^{\pi}[\mathds 1\{n_{T(m), 2} < N\}] \\
& = \mbE_{\tilde\theta(m)}^{\pi}\left[\exp\left(\sum_{n=1}^{n_{T(m), 2}}\frac{(X_{t_2(n), 2} - \tilde\theta(m)_2)^2 - (X_{t_2(n), 2} - \theta(m)_2)^2}{2\sigma^2}\right)\mathds 1\{n_{T(m), 2} < N\}\right] \\
& = \mathbb E_{\tilde\theta(m)}^{\pi}\left[\exp\left(n_{T(m), 2}\left(\frac{\tilde\theta(m)_2^2 - \theta(m)_2^2}{2\sigma^2} + \frac{(\theta(m)_2 - \tilde\theta(m)_2)\hat\theta_{T(m), 2}}{\sigma^2}\right)\right)\mathds 1\{n_{T(m), 2} < N\}\right] \\
& \geq \mathbb E_{\tilde\theta(m)}^{\pi}\left[\exp\left(n_{T(m), 2}\left(\frac{\tilde\theta(m)_2^2 - \theta(m)_2^2}{2\sigma^2} + \frac{(\theta(m)_2 - \tilde\theta(m)_2)\hat\theta_{T(m), 2}}{\sigma^2}\right)\right) \right.\\
& \quad\quad\quad\quad\quad \left.\mathds 1\{\hat\theta_{T(m), 2} > \tilde\theta(m)_2 - M, n_{T(m), 2} < N\}\right] \\
& \geq \mathbb E_{\tilde\theta(m)}^{\pi}\left[\exp\left(N\left(-\frac{(\tilde\theta(m)_2 - \theta(m)_2)^2}{2\sigma^2} - \frac{M(\theta(m)_2 - \tilde\theta(m)_2)}{\sigma^2}\right)\right)\mathds 1\{\hat\theta_{T(m), 2} > \tilde\theta_2 - M, n_{T(m), 2} < N\}\right] \\
& = \exp\left(N\left(-\frac{(\tilde\theta(m)_2 - \theta(m)_2)^2}{2\sigma^2} - \frac{M(\theta(m)_2 - \tilde\theta(m)_2)}{\sigma^2}\right)\right)\mathbb P_{\tilde\theta}^{\pi}(\hat\theta_{T(m), 2} > \tilde\theta(m)_2 - M, n_{T(m), 2} < N) \\
& \geq \exp\left(N\left(-\frac{1}{2\sigma^2} - \frac{M}{\sigma^2}\right)\right)\mathbb P_{\tilde\theta}^{\pi}(\hat\theta_{T(m), 2} > \tilde\theta(m)_2 - M, n_{T(m), 2} < N)
\end{align*}
Note that
\begin{align*}
& \quad\ \mathbb P_{\tilde\theta}^\pi(\hat\theta_{T(m), 2} > \tilde\theta_2 - M, n_{T(m), 2} < N) \\
& = p - \sum_{n=1}^{N-1}\mathbb P_{\tilde\theta}^\pi(\hat\theta_{T(m), 2} \leq \tilde\theta_2 - M, n_{T(m), 2} = n) \\
& \geq p - \sum_{n=1}^{N-1}\exp(-\frac{nM^2}{2\sigma^2}) \geq p - N\exp(-\frac{M^2}{2\sigma^2})=q>0.
\end{align*}
Therefore, there exists a constant positive probability such that $\pi$ pulls arm $2$ no more than $N$ times under $\theta(m)$. As a result, $\pi$ incurs a linear expected regret under $\theta(m)$, leading to a contradiction.

$\hfill\Box$

{\noindent\bf Proof of Theorem \ref{thm:tightness}.} 

We consider the environment where the noise $\epsilon$ is Gaussian with standard deviation $\sigma$. Fix any $\alpha > 1/2$. Let $\theta_1 = 1/2$. Let $\theta(T) = (\theta_1, \theta_2(T))$ and $\tilde\theta(T) = (\theta_1, \tilde\theta_2(T))$, where $\theta_2(T) = \theta_1 + \frac{1}{2T^{1-\alpha}}$ and $\tilde\theta_2(T) = \theta_1 - \frac{1}{2T^{1-\alpha}}$. Let \begin{align*}
    \gamma\in(3/2-\alpha, \quad \min\{1, \beta+2-2\alpha\}).
\end{align*}
Such $\gamma$ always exists because $\beta+2-2\alpha > \alpha-1/2 + 2 - 2\alpha = 3/2-\alpha$ and $3/2-\alpha < 3/2 - 1/2 = 1$. For notation simplicity, we will write $\theta$ ($\tilde\theta$) instead of $\theta(T)$ ($\tilde\theta(T)$), but we must keep in mind that $\theta$ ($\tilde\theta$) is dependent on $T$. Define
\begin{align*}
    E_T = \left\{|\hat\theta_{T, 2} - \tilde\theta_2|\leq \delta/n_{T, 2}^{(1-\alpha)/\gamma}\right\}
\end{align*}
where $\delta>0$ is a small number, and
\begin{align*}
    F_T = \{n_2\leq T^\gamma\}.
\end{align*}
Note that in $E_T$ we have $(1-\alpha)/\gamma\in[0, 1/2)$ since $\alpha\in(1/2, 1]$. Then under the environment $\tilde\theta$, we have
\begin{align*}
    \mbP_{\tilde\theta}^\pi(\bar F_T) & = \mbP_{\tilde\theta}^\pi\left(n_2 > T^\gamma\right) \leq \frac{\mbE_{\tilde\theta}^\pi[n_2]}{T^\gamma} \leq \frac{\mbE[R_{\tilde\theta}^\pi(T)]}{T^{\gamma+\alpha - 1}} \leq \frac{\sup_{\theta'}\mbE[R_{\theta'}^\pi(T)]}{T^{\gamma+\alpha-1}} \longrightarrow 0
\end{align*}
as $T\to+\infty$. Combined with Lemma \ref{lemma:sub-linear-w}, we have
\begin{align*}
    \liminf_T \mbP_{\tilde\theta}^\pi(E_T, F_T) = 1.
\end{align*}
Let $c'\in(c, 1/2)$. There exists $T_0$ such that $(1-2c')T > T^\gamma$ for any $T > T_0$. Fix $T > T_0$. Notice that
\begin{align*}
    & \quad \mbP\left(\hat R_{\theta}^\pi(T)\geq cT^\alpha\right) \\
    & \geq \mbP\left(R_{\theta}^\pi(T)\geq c'T^\alpha, N^\pi(T) \geq -(c'-c)T^\alpha\right) \\
    & = \mbP\left(R_{\theta}^\pi(T)\geq c'T\right) - \mbP\left(R_{\theta}^\pi(T)\geq c'T^\alpha, N^\pi(T) < -(c'-c)T^\alpha\right) \\
    & \geq \mbP\left(R_{\theta}^\pi(T)\geq c'T^\alpha\right) - \mbP\left(N^\pi(T) < -(c'-c)T^\alpha\right) \\
    & \geq \mbP\left(R_{\theta}^\pi(T)\geq c'T^\alpha\right) - \exp\left(-\frac{(c'-c)^2T^{2\alpha-1}}{2\sigma^2}\right)
\end{align*}
The last inequality holds from Lemma \ref{lemma:bound-noise}. Now
\begin{align*}
    & \quad \mbP\left(R_{\theta}^\pi(T)\geq c'T^\alpha\right) \\
    & \geq \mbP_{\theta}^\pi(n_1\geq 2c'T) \\
    & \geq \mbP_{\theta}^\pi(n_2\leq (1-2c')T) \\
    & \geq \mbP_{\theta}^\pi(n_2\leq T^\gamma) \\
    & \geq \mbP_{\theta}^\pi(E_T, F_T) \\
    & = \mbE_{\theta}^\pi[\mathds 1\{E_T F_T\}]
\end{align*}
Now we apply the change-of-measure argument.
\begin{align*}
    & \quad \mbE_{\theta}^\pi[\mathds 1\{E_T F_T\}] \\
    & = \mbE_{\tilde\theta}^\pi\left[\exp\left(\sum_{n=1}^{n_2}\frac{(r_{t_2(n), 2} - \tilde\theta_2)^2 - (r_{t_2(n), 2} - \theta_2)^2}{2\sigma^2}\right)\mathds 1\{E_T F_T\}\right] \\
    & = \mbE_{\tilde\theta}^\pi\left[\exp\left(n_2\left(\frac{\tilde\theta_2^2 - \theta_2^2}{2\sigma^2} + \frac{(\theta_2 - \tilde\theta_2)\hat\theta_{T, 2}}{\sigma^2}\right)\right)\mathds 1\{E_T F_T\}\right] \\
    & \geq \mbE_{\tilde\theta}^\pi\left[\exp\left(n_2\left(\frac{\tilde\theta_2^2 - \theta_2^2}{2\sigma^2} + \frac{(\theta_2 - \tilde\theta_2)(\tilde\theta_{2} - \delta/n_2^{(1-\alpha)/\gamma})}{\sigma^2}\right)\right)\mathds 1\{E_T F_T\}\right] \\
    & = \mbE_{\tilde\theta}^\pi\left[\exp\left(n_2\left(-\frac{(\tilde\theta_2 - \theta_2)^2}{2\sigma^2} - \frac{\delta(\theta_2 - \tilde\theta_2)}{\sigma^2n_2^{(1-\alpha)/\gamma}}\right)\right)\mathds 1\{E_T F_T\}\right] \\
    & \geq \mbE_{\tilde\theta}^\pi\left[\exp\left(-T^\gamma\frac{(\tilde\theta_2 - \theta_2)^2}{2\sigma^2} - T^{\gamma-1+\alpha}\frac{\delta(\theta_2 - \tilde\theta_2)}{\sigma^2}\right)\mathds 1\{E_T F_T\}\right] \\
    & \geq \exp(-T^{\gamma + 2\alpha - 2}/(2\sigma^2) - \delta T^{\gamma+2\alpha-2}/\sigma^2)\mbP_{\tilde\theta}^\pi(E_T, F_T).
\end{align*}
Notice that
\begin{align*}
    \gamma+2\alpha-2 < \beta,
\end{align*}
and $\delta > 0$ can be arbitrary. Therefore, 
\begin{align*}
    & \quad \liminf_T\frac{\ln\left\{\sup_{\theta'} \mbP(\hat R_{\theta'}^\pi(T)\geq cT^\alpha)\right\}}{T^\beta} \geq \liminf_T \frac{-T^{\gamma + 2\alpha - 2}/(2\sigma^2)}{T^\beta} = 0.
\end{align*}
Since $\ln\left\{\sup_{\theta}\mbP(R_\theta^\pi(T)\geq cT^\alpha)\right\} \leq 0$ always holds, we obtain the result.

$\hfill\Box$

{\noindent\bf Proof of Theorem \ref{thm:exp-K-armed}.}
Without loss of generality, we assume $\theta_1  = \theta_*$. We prove the results one by one.
\begin{enumerate}
\item From Lemma \ref{lemma:bound-noise}, 
\begin{align*}
    \mbE[R_\theta^\pi(T)] = \mbE[\hat R_\theta^\pi(T)] = \sum_{k=2}^K\mbE[n_k]\cdot\Delta_k.
\end{align*}
Let $G$ be the event such that
\begin{align*}
    G = \{\theta_k\in \CI_{t, k},\ \forall (t, k)\}.
\end{align*}
Then
\begin{align*}
    \mbP(\bar G) \leq \sum_{(t, k)}\mbP(\theta_k\notin\CI_{t, k}) \leq K\sum_{n=1}^T2\exp(-\frac{\eta T\ln T}{2n}) \leq 2KT^{1-\eta/2}.
\end{align*}
Thus, 
\begin{align*}
    \mbE[R_\theta^\pi] & = \sum_{k=2}^K
    \left(\mbE[n_k|G]\mbP(G) + \mbE[n_k|\bar G]\mbP(\bar G)\right)\Delta_k \\
    & \leq \sum_{k=2}^K\mbE[n_k|G] + T\cdot\mbP(\bar G) \\
    & \leq \sum_{k=2}^K\mbE[n_k|G] + 2KT^{2-\eta/2} \\
    & \leq \sum_{k=2}^K\mbE[n_k|G] + 2K.
\end{align*}
\begin{itemize}
    \item[(a)] Let $\pi = \SE$. Fix any arm $k\neq 1$. We let $t_k'$ be the largest time period such that we have traversed all the arms in $\cA$, and meanwhile arm $k$ is not eliminated from $\cA$. Then $n_k = n_{t_k', k} + 1$. When doing the elimination after $t_k$, arm $1$ and $k$ are both pulled $n_{t_k', k}$ times. Under $G$, we have
    \begin{align*}
        \theta_1 - 2\sigma\frac{\sqrt{\eta T\ln T}}{n_{t_k', k}} & \leq \theta_1 - 2\sigma\frac{\sqrt{\eta T\ln T}}{n_{t_k', 1}} \leq \hat\theta_{t_k', 1} - \sigma\frac{\sqrt{\eta T\ln T}}{n_{t_k', 1}} \leq \hat\theta_{t_k', k} + \sigma\frac{\sqrt{\eta T\ln T}}{n_{t_k', k}} \leq \theta_k + 2\sigma\frac{\sqrt{\eta T\ln T}}{n_{t_k', k}}.
    \end{align*}
    Therefore, 
    \begin{align*}
        n_{t_k', k} \leq 1 + 4\sigma\frac{\sqrt{\eta T\ln T}}{\Delta_k}
    \end{align*}
    and thus, 
    \begin{align*}
        n_k \leq 2 + 4\sigma\frac{\sqrt{\eta T\ln T}}{\Delta_k}.
    \end{align*}
    As a result, 
    \begin{align*}
        \mbE[\hat R_\theta^\pi(T)] \leq 2\sum_{k=2}^K\Delta_k + 4\sum_{k=2}^K\sigma\sqrt{\eta T\ln T} + 2K \leq 4K + 4K\sigma\sqrt{\eta T\ln T}.
    \end{align*}
    \item[(b)] Let $\pi = \UCB$. Fix any arm $k\neq 1$. We let $t_k$ be the largest time period such that arm $k$ is pulled. Then $n_k = n_{t_k, k} = n_{t_k-1, k}+1$. Under $G$, we have
    \begin{align*}
        \theta_1 \leq \hat\theta_{t_k-1, 1} + \sigma\frac{\sqrt{\eta T\ln T}}{n_{t_k-1, 1}} \leq \hat\theta_{t_k-1, k} + \sigma\frac{\sqrt{\eta T\ln T}}{n_{t_k-1, k}} \leq \theta_k + 2\sigma\frac{\sqrt{\eta T\ln T}}{n_{t_k-1, k}}
    \end{align*}
    Therefore, 
    \begin{align*}
        n_{t_k-1, k} \leq 2\sigma\frac{\sqrt{\eta T\ln T}}{\Delta_k}
    \end{align*}
    and thus, 
    \begin{align*}
        n_k \leq 1 + 2\sigma\frac{\sqrt{\eta T\ln T}}{\Delta_k}.
    \end{align*}
    As a result, 
    \begin{align*}
        \mbE[\hat R_\theta^\pi(T)] \leq \sum_{k=2}^K\Delta_k + 2\sum_{k=2}^K\sigma\sqrt{\eta T\ln T} + 2K \leq 3K + 2K\sigma\sqrt{\eta T\ln T}.
    \end{align*}
\end{itemize}

\item We have
\begin{align*}
    \mbP(\hat R_{\theta}^\pi(T) \geq x) & \leq \mbP\left(R_\theta^\pi(T) \geq x(1-1/\sqrt{K})\right) + \mbP\left(N_\theta^\pi(T) \leq -x/\sqrt{K}\right)
\end{align*}
From Lemma \ref{lemma:bound-noise}, the second term can be bounded as
\begin{align} \label{eq:bound-noise-K}
    \mbP\left(N_\theta^\pi(T) \leq -x/K\right) \leq \exp\left(-\frac{x^2}{2K\sigma^2T}\right).
\end{align}
We are left to bound $\mbP\left(R_\theta^\pi(T) \geq x(1-1/\sqrt{K})\right)$.

\begin{itemize}
    \item[(a)] Let $\pi = \SE$. For any $k\neq 1$, let $S_k$ be the event defined as
    \begin{align*}
        S_k = \{\text{Arm }1\text{ is not eliminated before arm }k\}.
    \end{align*}
    Then
    \begin{align*}
        \bar S_k = \{\text{Arm }1\text{ is eliminated before arm }k\}.
    \end{align*}
    So
    \begin{align*}
        & \quad \mbP\left(R_\theta^\pi(T) \geq x(1-1/\sqrt{K})\right) \\
        & \leq \sum_{k=2}^K\mbP\left(n_k\Delta_k\geq x/(K+\sqrt K)\right) \\
        & = \sum_{k=2}^K\mbP(n_k\Delta_k\geq x/2K, S_k) + \mbP(n_k\Delta_k\geq x/2K, \bar S_k).
    \end{align*}
    Let $x > 0$. Fix any $k\neq 1$. With a slight abuse of notation, we let $n_0 = \lceil x/(2K\Delta_k)\rceil$ - 1, then
    \begin{align*}
        n_0 \geq x/(2K\Delta_k) - 1 \geq (x-2K)/(2K\Delta_k).
    \end{align*}
    Also, if $n_k\Delta_k \geq x/2K$, we must have
    \begin{align*}
        T \geq x/(2K\Delta_k),
    \end{align*}
    which means $\Delta_k\geq x/2KT$. By the definition of $n_0$, arm $k$ is not eliminated after being pulled $n_0$ times. So under $S_k$, after arm $1$ being pulled $n_0$ times, it is still in the active set. We have
    \begin{align}
        & \quad \mbP(n_k\Delta_k\geq x/2K, S_k) \nonumber\\
        & = \mbP(n_k \geq x/(2K\Delta_k), S_k) \nonumber\\
        & \leq \mbP(\text{arm }1\text{ and }k\text{ are both not eliminated after each of them being pulled }n_0\text{ times}) \nonumber\\
        & \leq \mbP\left(\hat\theta_{t_1(n_0), 1} - \frac{\sigma\sqrt{\eta T\ln T}}{n_0}\leq \hat\theta_{t_k(n_0), k} + \frac{\sigma\sqrt{\eta T\ln T}}{n_0}\right) \quad\quad\text{(otherwise arm $k$ is eliminated)}\nonumber\\
        & = \mbP\left(\theta_1 - \frac{\sum_{m=1}^{n_0}\epsilon_{t_1(m), 1} + \sigma\sqrt{\eta T\ln T}}{n_0}\leq \theta_k + \frac{\sum_{m=1}^{n_0}\epsilon_{t_k(m), k} + \sigma\sqrt{\eta T\ln T}}{n_0}\right) \nonumber\\
        & = \mbP\left(\frac{\sum_{m=1}^{n_0}(\epsilon_{t_1(m), 1} - \epsilon_{t_k(m), k})}{n_0} \geq \Delta_k - \frac{2\sigma\sqrt{\eta T\ln T}}{n_0}\right) \nonumber\\
        & \leq \mbP\left(\frac{\sum_{m=1}^{n_0}\epsilon_{t_1(m), 1}}{n_0} \geq \frac{\Delta_k}{2} - \frac{\sigma\sqrt{\eta T\ln T}}{n_0}\right) + \mbP\left(\frac{\sum_{m=1}^{n_0}-\epsilon_{t_k(m), k}}{n_0} \geq \frac{\Delta_k}{2} - \frac{\sigma\sqrt{\eta T\ln T}}{n_0}\right) \nonumber\\
        & \leq 2\exp\left(-n_0\left(\frac{\Delta_k}{2} - \frac{\sigma\sqrt{\eta T\ln T}}{n_0}\right)_+^2\big/(2\sigma^2)\right) \nonumber\\
        & = 2\exp\left(-n_0\Delta_k^2\left(1 - \frac{2\sigma\sqrt{\eta T\ln T}}{n_0\Delta_k}\right)_+^2\big/8\sigma^2\right) \nonumber\\
        & \leq 2\exp\left(-\frac{x(x-2K)_+}{4KT}\left(1 - \frac{4K^2\sigma\sqrt{\eta T\ln T}}{x-2K}\right)_+^2\big/8\sigma^2\right) \quad\quad(n_0\geq(x-2K)/(2K\Delta_k), \Delta_k\geq x/(2KT))\nonumber\\
        & \leq 2\exp\left(-\frac{(x-2K-4K\sigma\sqrt{\eta T\ln T})_+^2}{32\sigma^2K^2T}\right). \label{eq:bound-pseudo-identified-K}
    \end{align}
    In the following, we bound $\mbP(n_k\Delta_k \geq x/2K, \bar S_k)$. Suppose that after $n$ phases, arm $1$ is eliminated by arm $k'$ ($k'$ is not necessarily $k$). By the definition of $\bar S_k$, arm $k$ is not eliminated. Therefore, we have
    \begin{align}
        \hat\theta_{t_{k'}(n), k'} - \frac{\sigma\sqrt{\eta T\ln T}}{n} \geq \hat\theta_{t_1(n), 1} + \frac{\sigma\sqrt{\eta T\ln T}}{n}, \quad
        \hat\theta_{t_k(n), k} + \frac{\sigma\sqrt{\eta T\ln T}}{n} \geq \hat\theta_{t_1(n), 1} + \frac{\sigma\sqrt{\eta T\ln T}}{n} \label{eq:event-bad}
    \end{align}
    holds simultaneously. The first inequality holds because arm $1$ is eliminated. The second inequality holds because arm $k$ is not eliminated. Now for fixed $n$, 
    \begin{align*}
        & \quad \mbP\left(\text{\eqref{eq:event-bad} happens}; \Delta_k \geq \frac{x}{2KT}\right) \\
        & \leq \mbP\left(\exists k': \hat\theta_{t_{k'}(n), k'} - \frac{\sigma\sqrt{\eta T\ln T}}{n} \geq \hat\theta_{t_1(n), 1} + \frac{\sigma\sqrt{\eta T\ln T}}{n}\right) \\ & \quad\quad \wedge \mbP\left(\hat\theta_{t_k(n), k} + \frac{\sigma\sqrt{\eta T\ln T}}{n} \geq \hat\theta_{t_1(n), 1} + \frac{\sigma\sqrt{\eta T\ln T}}{n}; \Delta_k \geq x/2KT\right) \\
        & \leq \mbP\left(\exists k':\frac{\sum_{m=1}^{n}(\epsilon_{t_{k'}(m), k'} - \epsilon_{t_1(m), 1})}{n} \geq \frac{2\sigma\sqrt{\eta T\ln T}}{n}\right) \\
        & \quad \quad \wedge \mbP\left(\frac{\sum_{m=1}^{n}(\epsilon_{t_k(m), k} - \epsilon_{t_1(m), 1})}{n} > \Delta_k; \Delta_k \geq x/2KT\right) \\
        & \leq \sum_{k'\neq 1}\left(\mbP\left(\frac{\sum_{m=1}^{n}\epsilon_{t_{k'}(m), k'}}{n} \geq \frac{\sigma\sqrt{\eta T\ln T}}{n}\right) + \mbP\left(\frac{\sum_{m=1}^{n}\epsilon_{t_1(m), 1}}{n} \leq - \frac{\sigma\sqrt{\eta T\ln T}}{n}\right)\right) \\
        & \quad \quad \wedge \left(\mbP\left(\frac{\sum_{m=1}^{n}\epsilon_{t_{k}(m), k}}{n} \geq \frac{x}{4KT}\right) + \mbP\left(\frac{\sum_{m=1}^{n}\epsilon_{t_1(m), 1}}{n} \leq - \frac{x}{4KT}\right)\right) \\
        & \leq 2K\exp\left(-\frac{\eta T\ln T}{2n}\right) \wedge 2K\exp\left(-\frac{nx^2}{32\sigma^2K^2T^2}\right) \\
        & = 2K\exp\left(-\left(\frac{\eta T\ln T}{2n}\vee \frac{nx^2}{32\sigma^2K^2T^2} \right)\right) \\
        & \leq 2K\exp\left(-\frac{x\sqrt{\eta \ln T}}{8\sigma K\sqrt{T}}\right)
    \end{align*}
    Therefore, 
    \begin{align}
        & \quad \mbP(n_k\Delta_k \geq x/2K, \bar S) \nonumber\\
        & =  \mbP\left(\exists n\leq T/2: \text{\eqref{eq:event-bad} happens}; n_k\Delta_k \geq x/2K\right) \nonumber\\
        & \leq \sum_{n=1}^{\lfloor T/2\rfloor}\mbP\left(\text{\eqref{eq:event-bad} happens}; \Delta_k \geq \frac{x}{2KT}\right) \nonumber\\
        & \leq KT \exp\left(-\frac{x\sqrt{\eta \ln T}}{8\sigma K\sqrt{T}}\right).
        \label{eq:bound-pseudo-unidentified-K}
    \end{align}
    
    Note that the equations above hold for any instance $\theta$. Combining (\ref{eq:bound-noise-K}), (\ref{eq:bound-pseudo-identified-K}), (\ref{eq:bound-pseudo-unidentified-K}) yields
    \begin{align*}
        \sup_{\theta}\mbP(\hat R_\theta^\pi(T)\geq x) \leq \exp\left(-\frac{x^2}{2K\sigma^2T}\right) + 2K\exp\left(-\frac{(x-2K-4K\sigma\sqrt{\eta T\ln T})_+^2}{32\sigma^2K^2T}\right) + K^2T\exp\left(-\frac{x\sqrt{\eta\ln T}}{8\sigma K\sqrt{T}}\right)
    \end{align*}
    
    \item[(b)] Let $\pi=\UCB$. From (a), we know that \begin{align*}
        \mbP\left(R_\theta^\pi(T) \geq x(1-1/\sqrt{K})\right) & \leq \sum_{k=2}^K\mbP\left(n_k\Delta_k\geq x/(K+\sqrt K)\right) \leq \sum_{k=2}^K\mbP(n_k\Delta_k\geq x/2K).
    \end{align*}
    Let $x > 0$. Fix $k\neq 1$. With a slight abuse of notation, we let $n_0 = \lceil x/(2K\Delta_k)\rceil - 1$. Remember that $t_k(n_0+1)$ is the time period that arm $k$ is pulled for the $(n_0+1)$th time. We emphasize again that $\Delta_k\geq x/(2KT)$. Then
    \begin{align}
        & \quad \mbP(n_k\Delta_k\geq x/2K) \nonumber\\
        & = \mbP(n_k \geq x/(2K\Delta_k)) \nonumber\\
        & \leq \mbP\left(\hat\theta_{t_k(n_0+1)-1, 1} + \frac{\sigma\sqrt{\eta T\ln T}}{n_{t_k(n_0+1)-1, 1}}\leq \hat\theta_{t_k(n_0+1)-1, k} + \frac{\sigma\sqrt{\eta T\ln T}}{n_0}\right) \nonumber\\
        & = \mbP\left(\theta_1 + \frac{\sum_{m=1}^{n_{t_k(n_0+1)-1, 1}}\epsilon_{t_1(m), 1} + \sigma\sqrt{\eta T\ln T}}{n_{t_k(n_0+1)-1, 1}}\leq \theta_k + \frac{\sum_{m=1}^{n_0}\epsilon_{t_k(m), k} + \sigma\sqrt{\eta T\ln T}}{n_0}\right) \nonumber\\
        & \leq \mbP\left(\exists n\in[T]: \frac{\sum_{m=1}^{n_0}\epsilon_{t_k(m), k} + \sigma\sqrt{\eta T\ln T}}{n_0} - \frac{\sum_{m=1}^{n}\epsilon_{t_1(m), 1} + \sigma\sqrt{\eta T\ln T}}{n} \geq \Delta_k \right) \nonumber\\
        & \leq \mbP\left(\frac{\sum_{m=1}^{n_0}\epsilon_{t_k(m), k} + \sigma\sqrt{\eta T\ln T}}{n_0} \geq \frac{\Delta_k}{2}\right) + \mbP\left(\exists n\in[T]: \frac{\sum_{m=1}^{n}\epsilon_{t_1(m), 1} + \sigma\sqrt{\eta T\ln T}}{n} \leq - \frac{\Delta_k}{2}\right) \nonumber\\
        & \leq \mbP\left(\frac{\sum_{m=1}^{n_0}\epsilon_{t_k(m), k} + \sigma\sqrt{\eta T\ln T}}{n_0} \geq \frac{\Delta_k}{2}\right) + \sum_{n=1}^{T}\mbP\left(\frac{\sum_{m=1}^{n}-\epsilon_{t_1(m), 1}}{n} \geq \frac{\Delta_k}{2} + \frac{\sigma\sqrt{\eta T\ln T}}{n}\right) \nonumber\\
        & \leq \mbP\left(\frac{\sum_{m=1}^{n_0}\epsilon_{t_k(m), k} + \sigma\sqrt{\eta T\ln T}}{n_0} \geq \frac{\Delta_k}{2}\right) + \sum_{n=1}^{T}\mbP\left(\frac{\sum_{m=1}^{n}-\epsilon_{t_1(m), 1}}{n} \geq \sqrt{\frac{x\sigma\sqrt{\eta T\ln T}}{nKT}}\right) \nonumber\\
        & \leq \exp\left(-\frac{(x-2K-4K\sigma\sqrt{\eta T\ln T})_+^2}{32\sigma^2KT}\right) + T\exp\left(-\frac{x\sqrt{\eta T\ln T}}{2\sigma KT}\right). \label{eq:bound-pseudo-K-ucb}
    \end{align}
    The last inequality holds from \eqref{eq:bound-pseudo-identified-K} and concentration of subGaussian variables. Note that the equations above hold for any instance $\theta$. Combining (\ref{eq:bound-noise-K}), (\ref{eq:bound-pseudo-K-ucb}) yields
    \begin{align*}
        \sup_{\theta}\mbP(\hat R_\theta^\pi(T)\geq x) \leq \exp\left(-\frac{x^2}{2K\sigma^2T}\right) + K\exp\left(-\frac{(x-2K-4K\sigma\sqrt{\eta T\ln T})_+^2}{32\sigma^2KT}\right) + K^2T\exp\left(-\frac{x\sqrt{\eta T\ln T}}{8\sigma KT}\right).
    \end{align*}
\end{itemize}
\end{enumerate}

$\hfill\Box$

{\noindent\bf Remark: $\SE$ or $\UCB$ with \eqref{rad:standard} may lead to a sub-optimal regret when $\eta$ is too small.}

Let $\theta = (1, 0)$ and $\sigma = 1$ with independent Gaussian noise. We first consider $\pi=\SE$. The probability that arm $1$ is eliminated after the first phase is
\begin{align*}
    & \quad \mbP\left(\hat\theta_{1, 1} + \sqrt{\eta\ln T} < \hat\theta_{2, 2} - \sqrt{\eta\ln T}\right) \\
    & = \mbP\left(\epsilon_{1, 1} - \epsilon_{2, 2} < - 1 - 2\sqrt{\eta\ln T}\right) \\
    & \geq \frac{1}{\sqrt{2\pi}}\frac{1/\sqrt{2} + \sqrt{2\eta\ln T}}{1 + (1/\sqrt{2} + \sqrt{2\eta\ln T})^2}\exp(-(1/\sqrt{2} + \sqrt{2\eta\ln T})^2/2) \\
    & = \Theta\left(\frac{T^{-\eta}}{\sqrt{\ln T}}\right) 
\end{align*}
The inequality holds because for a standard normal variable $X$, it is established that
\begin{align*}
    \mbP(X > t) \geq \frac{1}{\sqrt{2\pi}}\frac{t}{1+t^2}\exp(-t^2/2)
\end{align*}
Therefore, the expected regret is at least
\begin{align*}
    \Theta\left(\frac{T^{-\eta}}{\sqrt{\ln T}}\right) \cdot (T-2) = \Theta\left(\frac{T^{1-\eta}}{\sqrt{\ln T}}\right).
\end{align*}
If $\eta$ is very small, then apparently the regret is sub-optimal.

Now we consider $\pi=\UCB$. The probability that arm $1$ is pulled only once is
\begin{align*}
    & \quad\ \mbP\left(\forall 2\leq t \leq T: \hat\theta_{1, 1} + \sqrt{\eta\ln T} < \hat\theta_{t, 2} + \sqrt{\frac{\eta\ln T}{t-1}}\right) \\
    & \geq \mbP\left(\forall 2\leq t \leq T: \hat\theta_{1, 1} + \sqrt{\eta\ln T} < \hat\theta_{t, 2}\right) \\
    & \geq \mbP\left(\left\{\hat\theta_{1, 1} < -1 - 2\sqrt{\eta\ln T}\right\}\bigcap\left\{\forall 1\leq t <  T: \sum_{s=2}^{t+1}\epsilon_{s, 2} > -1 - t\sqrt{\eta\ln T}\right\}\right) \\
    & = \mbP\left(\left\{\epsilon_{1, 1} < -2 - 2\sqrt{\eta\ln T}\right\}\right)\cdot\mbP\left(\left\{\forall 1\leq t <  T: \sum_{s=2}^{t+1}\epsilon_{s, 2} > -1 - t\sqrt{\eta\ln T}\right\}\right)
\end{align*}
We have
\begin{align*}
    & \quad\ \mbP\left(\left\{\epsilon_{1, 1} < -2 - 2\sqrt{\eta\ln T}\right\}\right) \\
    & \geq \frac{1}{\sqrt{2\pi}}\frac{2+2\sqrt{\eta\ln T}}{1 + (2+2\sqrt{\eta\ln T})^2}\exp(-(2 + 2\sqrt{\eta\ln T})^2/2) \\
    & = \Theta\left(\frac{T^{-2\eta}}{\sqrt{\ln T}}\right)
\end{align*}
We use a martingale argument and the optional sampling theorem to bound the second probability. Define $Z_{t} = \sum_{s=2}^{t+1}\epsilon_{s, 2}$. Define the stopping time
\begin{align*}
    \tau = \inf_t\{Z_t \leq -1 - t\sqrt{\eta\ln T}\}
\end{align*}
Then
\begin{align*}
    & \quad\ \mbP\left(\left\{\forall 1\leq t <  T: \sum_{s=2}^{t+1}\epsilon_{s, 2} > -1 - t\sqrt{\eta\ln T}\right\}\right) = \mbP(\tau \geq T)
\end{align*}
For fixed $T$, $\tau\wedge(T-1)$ is finite. Notice that 
\begin{align*}
    \exp(-2\sqrt{\eta\ln T}Z_t - 2\eta \ln T\cdot t)
\end{align*}
is a martingale with mean $1$. By the optional sampling theorem, we have
\begin{align*}
    1 & = \mbE\left[\exp(-2\sqrt{\eta\ln T}Z_{\tau\wedge (T-1)} - 2\eta \ln T\cdot (\tau\wedge (T-1)))\right] \\
    & \geq \mbE\left[\exp(-2\sqrt{\eta\ln T}Z_{\tau} - 2\eta \ln T\cdot \tau)\mathds 1\{\tau < T\}\right] \\
    & \geq \exp(2\sqrt{\eta\ln T})\mbP\left(\tau < T\right)
\end{align*}
Therefore, the second probability is bounded by
\begin{align*}
    1 - \exp(-2\sqrt{\eta\ln T}).
\end{align*}
The expected regret is at least
\begin{align*}
    \Theta\left(\frac{T^{-2\eta}}{\sqrt{\ln T}}\right)\cdot \left(1 - \exp(-2\sqrt{\eta\ln T})\right) \cdot (T-2) = \Omega\left(\frac{T^{1-2\eta}}{\sqrt{\ln T}}\right).
\end{align*}

$\hfill\Box$

{\noindent\bf Proof of Theorem \ref{thm:exp-K-armed-optimal}.}
Without loss of generality, we assume $\theta_1  = \theta_*$. We prove the results one by one.
\begin{enumerate}
\item From Lemma \ref{lemma:bound-noise}, 
\begin{align*}
    \mbE[R_\theta^\pi] = \mbE[\hat R_\theta^\pi] = \sum_{k=2}^K\mbE[n_k]\cdot\Delta_k.
\end{align*}
Let $G$ be the event such that
\begin{align*}
    G = \{\theta_k\in \CI_{t, k},\ \forall (t, k)\}.
\end{align*}
Then
\begin{align*}
    \mbP(\bar G) \leq \sum_{(t, k)}\mbP(\theta_k\notin\CI_{t, k}) \leq K\sum_{n=1}^T2\exp(-\frac{\eta T\ln T}{2n}) \leq 2KT^{1-\eta/2}.
\end{align*}
Thus, 
\begin{align*}
    \mbE[R_\theta^\pi] & = \sum_{k=2}^K
    \left(\mbE[n_k|G]\mbP(G) + \mbE[n_k|\bar G]\mbP(\bar G)\right)\Delta_k \\
    & \leq \sum_{k=2}^K\mbE[n_k|G]\Delta_k + T\cdot\mbP(\bar G) \\
    & \leq \sum_{k=2}^K\mbE[n_k|G]\Delta_k + 2KT^{2-\eta/2} \\
    & \leq \sum_{k=2}^K\mbE[n_k|G]\Delta_k + 2K.
\end{align*}
Define the (random) arm set
\begin{align*}
    \cA_0 = \left\{k\neq 1: n_k \leq 1 + \frac{T}{K}\right\}
\end{align*}
as the set of arms that are pulled no more than $1 + \frac{T}{K}$ times. Then
\begin{align*}
    \mbE[R_\theta^\pi] \leq \mbE\left[\sum_{k\in \cA_0}n_k\Delta_k\Big|G\right] + \mbE\left[\sum_{k\notin \cA_0}n_k\Delta_k\Big|G\right] + 2K
\end{align*}

\begin{itemize}
    \item[(a)] Let $\pi = \SE$. Fix any $k\neq 1$. We let $t_k'$ be the largest time period such that we have traversed all the arms in $\cA$, and meanwhile arm $k$ is not eliminated from $\cA$. Then $n_k = n_{t_k', k}$ or $n_k = n_{t_k', k} + 1$. When doing the elimination after $t_k$, arm $1$ and $k$ are both pulled $n_{t_k', k}$ times. Under $G$, we have
    \begin{align*}
        \theta_1 - 2\rad(n_{t_k', k}) \leq \hat\theta_{t_k', 1} - \rad(n_{t_k', 1})
        \leq \hat\theta_{t_k', k} + \rad(n_{t_k', k}) \leq \theta_k + 2\rad(n_{t_k', k}).
    \end{align*}
    
    \begin{itemize}
        \item Fix any $k\in \cA_0$. Then under $G$, we have
        \begin{align*}
            \Delta_k\leq 4\rad(n_{t_k', k}) \leq 4\sigma\frac{\sqrt{(\eta_1\vee\eta_2)T\ln T}}{\sqrt{K} n_{t'_k, k}}.
        \end{align*}
        Therefore,
        \begin{align*}
            n_{t_k', k} \leq 4\sigma\frac{\sqrt{(\eta_1\vee\eta_2) T\ln T}}{\sqrt{K}\Delta_k}
        \end{align*}
        and thus, 
        \begin{align*}
            n_k \leq 2 + 4\sigma\frac{\sqrt{(\eta_1\vee\eta_2) T\ln T}}{\sqrt{K}\Delta_k}.
        \end{align*}
        As a result, 
        \begin{align*}
            \mbE\left[\sum_{k\in \cA_0} n_k\Delta_k\Big|G\right] & \leq 2\mbE\left[\sum_{k\in \cA_0}^K\Delta_k\right] + 4\mbE\left[\sum_{k\in \cA_0}\sigma\frac{\sqrt{(\eta_1\vee\eta_2) T\ln T}}{\sqrt K}\right] \\
            & \leq 2\mbE[|S|] + 4\sigma\frac{\mbE[|S|]}{\sqrt K}\sqrt{(\eta_1\vee\eta_2) T\ln T}.
        \end{align*}
        \item Fix any $k\notin \cA_0$. Then under $G$, we have
        \begin{align*}
            \Delta_k\leq 4\rad(n_{t_k', k}) \leq 4\sigma\frac{\sqrt{(\eta_1\vee\eta_2)\ln T}}{\sqrt{n_{t_k', k}}}.
        \end{align*}
        Therefore,
        \begin{align*}
            \Delta_k \leq 4\sigma\frac{\sqrt{(\eta_1\vee\eta_2)\ln T}}{\sqrt{n_{t_k', k}}}
        \end{align*}
        and thus, 
        \begin{align*}
            \Delta_k \leq 4\sigma\frac{\sqrt{(\eta_1\vee\eta_2)\ln T}}{\sqrt{\max\{n_{k}-2, 0\}}}.
        \end{align*}
        As a result, 
        \begin{align*}
            \mbE\left[\sum_{k\notin \cA_0} n_k\Delta_k\Big|G\right] & \leq 2\mbE\left[\sum_{k\notin \cA_0}^K\Delta_k\right] + 4\mbE\left[\sum_{k\notin \cA_0}\sigma\sqrt{(\eta_1\vee\eta_2)\max\{n_k-2, 0\}\ln T}\right] \\
            & \leq 2(K-\mbE[|S|]) + 4\mbE\left[\sum_{k\notin \cA_0}\sigma\sqrt{(\eta_1\vee\eta_2)n_k\ln T}\right] \\
            & \leq 2(K-\mbE[|S|]) + 4\sigma\sqrt{(\eta_1\vee\eta_2)(K - \mbE[|S|]) T\ln T}.
        \end{align*}
    \end{itemize}
    Now we have
    \begin{align*}
        \mbE[R_\theta^\pi] & \leq \mbE\left[\sum_{k\in S}n_k\Delta_k\Big|G\right] + \mbE\left[\sum_{k\notin \cA_0}n_k\Delta_k\Big|G\right] + 2K \\
        & \leq 4K + 4\sigma\sqrt{T\ln T}\left(\sqrt{(\eta_1\vee\eta_2)}\frac{\mbE[|S|]}{\sqrt K} + \sqrt{(\eta_1\vee\eta_2)(K - \mbE[|S|])}\right) \\
        & \leq 4K + 8\sigma\sqrt{(\eta_1\vee\eta_2)KT\ln T}.
    \end{align*}
    
    \item[(b)] Let $\pi = \UCB$. With a slight abuse of notation, we let $t_k$ be the largest time period such that arm $k$ is pulled. Then $n_k = n_{t_k, k} = 1 + n_{t_k-1, k}$. Under $G$, we have
    \begin{align*}
        \theta_1 \leq \hat\theta_1 + \rad(n_{t_k-1, 1}) \leq \hat\theta_{k} + \rad(n_{t_k-1, k}) \leq \theta_{k} + 2\rad(n_{t_k-1, k}).
    \end{align*}
    
    \begin{itemize}
        \item Fix any $k\in \cA_0$. Then under $G$, we have
        \begin{align*}
            \theta_1 \leq \theta_k + 2\sigma\frac{\sqrt{(\eta_1\vee\eta_2) T\ln T}}{\sqrt{K}(n_{t_k, k}-1)}.
        \end{align*}
        Therefore,
        \begin{align*}
            n_k = n_{t_k, k} \leq 1 + 2\sigma\frac{\sqrt{(\eta_1\vee\eta_2) T\ln T}}{\sqrt{K}\Delta_k}.
        \end{align*}
        As a result, 
        \begin{align*}
            \mbE\left[\sum_{k\in \cA_0} n_k\Delta_k\Big|G\right] & \leq \mbE\left[\sum_{k\in \cA_0}^K\Delta_k\right] + 2\mbE\left[\sum_{k\in \cA_0}\sigma\frac{\sqrt{(\eta_1\vee\eta_2) T\ln T}}{\sqrt K}\right] \\
            & \leq \mbE[|S|] + 2\sigma\frac{\mbE[|S|]}{\sqrt K}\sqrt{(\eta_1\vee\eta_2) T\ln T}.
        \end{align*}
        \item Fix any $k\notin \cA_0$. Then under $G$, we have
        \begin{align*}
            \theta_1 \leq \theta_k + 2\sigma\frac{\sqrt{(\eta_1\vee\eta_2)\ln T}}{\sqrt{n_{t_k, k}-1}}.
        \end{align*}
        Therefore,
        \begin{align*}
            \Delta_k \leq 2\sigma\frac{\sqrt{(\eta_1\vee\eta_2)\ln T}}{\sqrt{n_{t_k, k}-1}}
        \end{align*}
        and thus, 
        \begin{align*}
            \Delta_k \leq 2\sigma\frac{\sqrt{(\eta_1\vee\eta_2)\ln T}}{\sqrt{\max\{n_{k}-1, 0\}}}.
        \end{align*}
        As a result, 
        \begin{align*}
            \mbE\left[\sum_{k\notin \cA_0} n_k\Delta_k\Big|G\right] & \leq \mbE\left[\sum_{k\notin \cA_0}^K\Delta_k\right] + 2\mbE\left[\sum_{k\notin \cA_0}\sigma\sqrt{(\eta_1\vee\eta_2)\max\{n_k-2, 0\}\ln T}\right] \\
            & \leq (K-\mbE[|S|]) + 2\mbE\left[\sum_{k\notin \cA_0}\sigma\sqrt{(\eta_1\vee\eta_2)n_k\ln T}\right] \\
            & \leq (K-\mbE[|S|]) + 2\sigma\sqrt{(\eta_1\vee\eta_2)(K - \mbE[|S|]) T\ln T}.
        \end{align*}
    \end{itemize}
    Now we have
    \begin{align*}
        \mbE[R_\theta^\pi] & \leq \mbE\left[\sum_{k\in S}n_k\Delta_k\Big|G\right] + \mbE\left[\sum_{k\notin \cA_0}n_k\Delta_k\Big|G\right] + 2K \\
        & \leq 3K + 2\sigma\sqrt{T\ln T}\left(\sqrt{(\eta_1\vee\eta_2)}\frac{\mbE[|S|]}{\sqrt K} + \sqrt{(\eta_1\vee\eta_2)(K - \mbE[|S|])}\right) \\
        & \leq 4K + 4\sigma\sqrt{(\eta_1\vee\eta_2) KT\ln T}.
    \end{align*}
\end{itemize}

\item Let $x\geq 2K$. We have
\begin{align*}
    \mbP(\hat R_{\theta}^\pi(T) \geq x) & \leq \mbP\left(R_\theta^\pi(T) \geq x(1-1/(2\sqrt{K}))\right) + \mbP\left(N_\theta^\pi(T) \leq -x/(2\sqrt{K})\right)
\end{align*}
From Lemma \ref{lemma:bound-noise}, the second term can be bounded as
\begin{align} \label{eq:bound-noise-K-opt}
    \mbP\left(N_\theta^\pi(T) \leq -x/(2\sqrt{K})\right) \leq \exp\left(-\frac{x^2}{8K\sigma^2T}\right).
\end{align}
We are left to bound $\mbP\left(R_\theta^\pi(T) \geq x(1-1/(2\sqrt{K}))\right)$.

\begin{itemize}
    \item[(a)] Let $\pi = \SE$. For any $k\neq 1$, let $S_k$ be the event defined as
\begin{align*}
    S_k = \{\text{Arm }1\text{ is not eliminated before arm }k\}.
\end{align*}
Then
\begin{align*}
    \bar S_k = \{\text{Arm }1\text{ is eliminated before arm }k\}.
\end{align*}
So
\begin{align*}
    & \quad\ \mbP\left(R_\theta^\pi(T)\geq x(1-1/(2\sqrt{K}))\right) \\
    & = \mbP\left(\sum_{k\in \cA_0}n_k\Delta_k + \sum_{k\notin \cA_0}n_k\Delta_k\geq x(1-1/(2\sqrt{K}))\right) \\
    & \leq \mbP\left(\sum_{k\in \cA_0}(n_k-1)\Delta_k + \sum_{k\notin \cA_0}(n_k-1)\Delta_k\geq x(1-1/(2\sqrt{K})) - K\right) \\
    & \leq \mbP\left(\left(\bigcup_{k\in\cA_0}\left\{(n_k-1)\Delta_k\geq\frac{x-2K}{4K}\right\}\right)\bigcup\left(\bigcup_{k\notin\cA_0}\left\{(n_k-1)\Delta_k\geq \frac{(n_k-1)(x-2K)}{4T}\right\}\right)\right) \\
    & \leq \sum_{k\neq 1}\left(\mbP\left((n_k-1)\Delta_k\geq \frac{x-2K}{4K}, \ k\in\cA_0\right) + \mbP\left((n_k-1)\Delta_k\geq \frac{(n_k-1)(x-2K)}{4T}, \ k\notin\cA_0\right)\right) \\
    & \leq \sum_{k\neq 1}\left(\mbP\left((n_k-1)\Delta_k\geq \frac{x-2K}{4K}, \ k\in\cA_0\right) + \mbP\left(\Delta_k\geq \frac{x-2K}{4T}, \ k\notin\cA_0\right)\right)
\end{align*}
The reason that the first inequality holds is as follows. To prove it, we only need to show that the following cannot holds:
\begin{align*}
    (n_k-1)\Delta_k < \frac{x-2K}{4K}, \quad \forall k\in\cA_0; \quad \quad (n_k-1)\Delta_k < \frac{(n_k-1)(x-2K)}{4T}, \quad \forall k\notin\cA_0.
\end{align*}
If not, then we have
\begin{align*}
    \sum_{k\neq 1} (n_k-1)\Delta_k & = \sum_{k\in\cA_0}(n_k-1)\Delta_k + \sum_{k\notin\cA_0}(n_k-1)\Delta_k \\
    & < \frac{(x-2K)|\cA_0|}{4K} + \frac{x-2K}{4} \\
    & \leq \frac{x-2K}{4} + \frac{x-2K}{4} \\
    & = \frac{x-2K}{2}\leq x(1-1/(2\sqrt{K})) - K.
\end{align*}
Therefore, 
\begin{align*}
    & \quad \mbP\left(R_\theta^\pi(T)\geq x(1-1/\sqrt{K})\right) \\
    & \leq \sum_{k\neq 1}\left(\mbP\left((n_k-1)\Delta_k\geq \frac{x-2K}{4K}, \ k\in\cA_0\right) + \mbP\left((n_k-1)\Delta_k\geq \frac{(n_k-1)(x-2K)}{4T}, \ k\notin\cA_0\right)\right) \\
    & = \sum_{k\neq 1}\mbP\left((n_k-1)\Delta_k\geq \frac{x-2K}{4K}, \ k\in\cA_0, S_k\right) + \sum_{k\neq 1}\mbP\left((n_k-1)\Delta_k\geq \frac{x-2K}{4K}, \ k\in\cA_0, \bar S_k\right) \\
    & \quad\quad + \sum_{k\neq 1}\mbP\left((n_k-1)\Delta_k\geq \frac{(n_k-1)(x-2K)}{4T}, \ k\notin\cA_0, S_k\right) + \sum_{k\neq 1}\mbP\left(\Delta_k\geq \frac{x-2K}{4T}, \ k\notin\cA_0, \bar S_k\right)
\end{align*}

Fix $k\neq 1$. Now for each $k$, we consider bounding the four terms separately.
\begin{itemize}
\item $k\in \cA_0$. With a slight abuse of notation, we let $n_0 = \lceil\frac{x-2K}{4K\Delta_k}\rceil\leq n_k-1$. Also, 
\begin{align*}
    \Delta_k\geq \frac{x-2K}{4K(n_k-1)} \geq \frac{(x-2K)K}{4KT} = \frac{x-2K}{4T}.
\end{align*}
We have
\begin{align}
    & \quad \mbP\left((n_k-1)\Delta_k\geq \frac{x-2K}{4K}, \ k\in\cA_0, S_k\right) \nonumber\\
    & \leq \mbP\left(\hat\theta_{t_1(n_0), 1} - \rad(n_0)\leq \hat\theta_{t_k(n_0), k} + \rad(n_0)\right)\mathds 1\left\{\Delta_k\geq \frac{x-2K}{4T}\right\} \nonumber\\
    & = \mbP\left(\theta_1 - \frac{\sum_{m=1}^{n_0}\epsilon_{t_1(m), 1}}{n_0} - \rad(n_0)\leq \theta_k + \frac{\sum_{m=1}^{n_0}\epsilon_{t_k(m), k}}{n_0} + \rad(n_0)\right)\mathds 1\left\{\Delta_k\geq \frac{x-2K}{4T}\right\} \nonumber\\
    & = \mbP\left(\frac{\sum_{m=1}^{n_0}(\epsilon_{t_1(m), 1} - \epsilon_{t_k(m), k})}{n_0} \geq \Delta_k - 2\rad(n_0)\right)\mathds 1\left\{\Delta_k\geq \frac{x-2K}{4T}\right\} \nonumber\\
    & \leq \mbP\left(\frac{\sum_{m=1}^{n_0}\epsilon_{t_1(m), 1}}{n_0} \geq \frac{\Delta_k}{2} - \rad(n_0)\right)\mathds 1\left\{\Delta_k\geq \frac{x-2K}{4T}\right\} \nonumber\\
    & \quad \quad + \mbP\left(\frac{\sum_{m=1}^{n_0}\epsilon_{t_k(m), k}}{n_0} \geq \frac{\Delta_k}{2} - \rad(n_0)\right)\mathds 1\left\{\Delta_k\geq \frac{x-2K}{4T}\right\} \nonumber\\
    & \leq 2\exp\left(-n_0\left(\frac{\Delta_k}{2} - \rad(n_0)\right)_+^2\big/(2\sigma^2)\right)\mathds 1\left\{\Delta_k\geq \frac{x-2K}{4T}\right\} \nonumber \\
    & = 2\exp\left(-n_0\left(\frac{\Delta_k}{2} - \frac{\sqrt{(\eta_1\vee\eta_2) T\ln T}}{n_0\sqrt{K}}\right)_+^2\big/(2\sigma^2)\right)\mathds 1\left\{\Delta_k\geq \frac{x-2K}{4T}\right\} \nonumber\\
    & = 2\exp\left(-n_0\Delta_k^2\left(1 - \frac{2\sqrt{(\eta_1\vee\eta_2) T\ln T}}{n_0\Delta_k\sqrt{K}}\right)_+^2\big/8\sigma^2\right)\mathds 1\left\{\Delta_k\geq \frac{x-2K}{4T}\right\} \nonumber\\
    & \leq 2\exp\left(-\frac{(x-2K)_+^2}{16KT}\left(1 - \frac{8\sigma\sqrt{(\eta_1\vee\eta_2) KT\ln T}}{x-2K}\right)_+^2\big/8\sigma^2\right) \nonumber\\
    & \leq 2\exp\left(-\frac{(x-2K-8\sigma\sqrt{(\eta_1\vee\eta_2)KT\ln T})_+^2}{128\sigma^2KT}\right). \label{eq:bound-pseudo-identified-K-opt-1}
\end{align}

Then we bound $\mbP(n_k\Delta_k \geq (x-2K)/4K, k\in\cA_0, \bar S_k)$. Suppose that after $n$ phases, arm $1$ is eliminated by arm $k'$ ($k'$ is not necessarily $k$). By the definition of $\bar S_k$, arm $k$ is not eliminated. Therefore, we have
    \begin{align}
        \hat\theta_{t_{k'}(n), k'} - \frac{\sigma\sqrt{\eta_1 T\ln T}}{n\sqrt{K}} \geq \hat\theta_{t_1(n), 1} + \frac{\sigma\sqrt{\eta_1 T\ln T}}{n\sqrt{K}}, \quad
        \hat\theta_{t_k(n), k} + \frac{\sigma\sqrt{\eta_1 T\ln T}}{n\sqrt{K}} \geq \hat\theta_{t_1(n), 1} + \frac{\sigma\sqrt{\eta_1 T\ln T}}{n\sqrt{K}} \label{eq:event-bad-opt}
    \end{align}
    holds simultaneously. The first inequality holds because arm $1$ is eliminated. The second inequality holds because arm $k$ is not eliminated. Now for fixed $n$, 
    \begin{align*}
        & \quad \mbP\left(\text{\eqref{eq:event-bad-opt} happens}; \Delta_k \geq \frac{x-2K}{4T}\right) \\
        & \leq \mbP\left(\exists k': \hat\theta_{t_{k'}(n), k'} - \frac{\sigma\sqrt{\eta_1 T\ln T}}{n\sqrt{K}} \geq \hat\theta_{t_1(n), 1} + \frac{\sigma\sqrt{\eta_1 T\ln T}}{n\sqrt{K}}\right) \\ & \quad\quad \wedge \mbP\left(\hat\theta_{t_k(n), k} + \frac{\sigma\sqrt{\eta_1 T\ln T}}{n\sqrt{K}} \geq \hat\theta_{t_1(n), 1} + \frac{\sigma\sqrt{\eta_1 T\ln T}}{n\sqrt{K}}; \Delta_k \geq \frac{x-2K}{4T}\right) \\
        & \leq \mbP\left(\exists k':\frac{\sum_{m=1}^{n}(\epsilon_{t_{k'}(m), k'} - \epsilon_{t_1(m), 1})}{n} \geq \frac{2\sigma\sqrt{\eta_1 T\ln T}}{n\sqrt{K}}\right) \\
        & \quad \quad \wedge \mbP\left(\frac{\sum_{m=1}^{n}(\epsilon_{t_k(m), k} - \epsilon_{t_1(m), 1})}{n} > \Delta_k; \Delta_k \geq \frac{x-2K}{4T}\right) \\
        & \leq \sum_{k'\neq 1}\left(\mbP\left(\frac{\sum_{m=1}^{n}\epsilon_{t_{k'}(m), k'}}{n} \geq \frac{\sigma\sqrt{\eta_1 T\ln T}}{n\sqrt{K}}\right) + \mbP\left(\frac{\sum_{m=1}^{n}\epsilon_{t_1(m), 1}}{n} \leq - \frac{\sigma\sqrt{\eta_1 T\ln T}}{n\sqrt{K}}\right)\right) \\
        & \quad \quad \wedge \left(\mbP\left(\frac{\sum_{m=1}^{n}\epsilon_{t_{k}(m), k}}{n} \geq \frac{(x-2K)_+}{8T}\right) + \mbP\left(\frac{\sum_{m=1}^{n}\epsilon_{t_1(m), 1}}{n} \leq - \frac{(x-2K)_+}{8T}\right)\right) \\
        & \leq 2K\exp\left(-\frac{\eta_1 T\ln T}{2nK}\right) \wedge 2K\exp\left(-\frac{n(x-2K)_+^2}{128\sigma^2T^2}\right) \\
        & = 2K\exp\left(-\left(\frac{\eta_1 T\ln T}{2nK}\vee \frac{n(x-2K)_+^2}{128\sigma^2T^2} \right)\right) \\
        & \leq 2K\exp\left(-\frac{(x-2K)_+\sqrt{\eta_1 \ln T}}{16\sigma\sqrt{KT}}\right)
    \end{align*}
    Therefore, 
    \begin{align}
        & \quad \mbP(n_k\Delta_k \geq (x-2K)/4K, \bar S_k, k\in\cA_0) \nonumber\\
        & =  \mbP\left(\exists n\leq T/2: \text{\eqref{eq:event-bad-opt} happens}; n_k\Delta_k \geq (x-2K)/4K, k\in\cA_0\right) \nonumber\\
        & \leq \sum_{n=1}^{\lfloor T/2\rfloor}\mbP\left(\text{\eqref{eq:event-bad-opt} happens}; \Delta_k \geq \frac{x-2K}{4T}\right) \nonumber\\
        & \leq KT \exp\left(-\frac{(x-2K)_+\sqrt{\eta_1 \ln T}}{16\sigma\sqrt{KT}}\right).
        \label{eq:bound-pseudo-unidentified-K-opt-1}
    \end{align}

\item $k\notin\cA_0$. With a slight abuse of notation, we let $n_0 = \lceil\frac{T}{K}\rceil\leq n_k - 1$. Also, 
\begin{align*}
    \Delta_k \geq \frac{x-2K}{4T}.
\end{align*}
We have
\begin{align}
    & \quad \mbP\left((n_k-1)\Delta_k \geq \frac{(n_k-1)(x-2K)}{4T}, \ k\notin\cA_0, S_k\right) \nonumber\\
    & \leq \mbP\left(\hat\theta_{t_1(n_0), 1} - \rad(n_0)\leq \hat\theta_{t_k(n_0), k} + \rad(n_0)\right) \nonumber\\
    & = \mbP\left(\theta_1 - \frac{\sum_{m=1}^{n_0}\epsilon_{t_1(m), 1}}{n_0} - \rad(n_0)\leq \theta_k + \frac{\sum_{m=1}^{n_0}\epsilon_{t_k(m), k}}{n_0} + \rad(n_0)\right) \nonumber\\
    & = \mbP\left(\frac{\sum_{m=1}^{n_0}(\epsilon_{t_1(m), 1} - \epsilon_{t_k(m), k})}{n_0} \geq \Delta_k - 2\rad(n_0)\right) \nonumber\\
    & \leq \mbP\left(\frac{\sum_{m=1}^{n_0}\epsilon_{t_1(m), 1}}{n_0} \geq \frac{\Delta_k}{2} - \rad(n_0)\right) + \mbP\left(\frac{\sum_{m=1}^{n_0}\epsilon_{t_k(m), 2}}{n_0} \geq \frac{\Delta_k}{2} - \rad(n_0)\right) \nonumber\\
    & \leq 2\exp\left(-n_0\left(\frac{\Delta_k}{2} - \rad(n_0)\right)_+^2\big/(2\sigma^2)\right) \nonumber \\
    & = 2\exp\left(-\left(\frac{\sqrt{n_0}\Delta_k}{2} - \sigma\sqrt{(\eta_1\vee\eta_2)\ln T}\right)_+^2\big/(2\sigma^2)\right) \nonumber\\
    & \leq 2\exp\left(-\left(\frac{x-2K}{8\sqrt{KT}} - \sigma\sqrt{(\eta_1\vee\eta_2)\ln T}\right)_+^2\big/(2\sigma^2)\right) \nonumber\\
    & \leq 2\exp\left(-\frac{\left(x-2K-8\sigma\sqrt{(\eta_1\vee\eta_2)KT\ln T}\right)_+^2}{128\sigma^2KT}\right). \label{eq:bound-pseudo-identified-K-opt-2}
\end{align}

Then we bound $\mbP((n_k-1)\Delta_k \geq (n_k-1)(x-2K)/4T, k\notin\cA_0, \bar S_k)$. The procedure is nearly the same as in the case where $k\in\cA_0$. Suppose that after $n$ phases, arm $1$ is eliminated by arm $k'$ ($k'$ is not necessarily $k$). By the definition of $\bar S_k$, arm $k$ is not eliminated. Therefore, we have
    \begin{align}
        \hat\theta_{t_{k'}(n), k'} - \frac{\sigma\sqrt{\eta_1 T\ln T}}{n\sqrt{K}} \geq \hat\theta_{t_1(n), 1} + \frac{\sigma\sqrt{\eta_1 T\ln T}}{n\sqrt{K}}, \quad
        \hat\theta_{t_k(n), k} + \frac{\sigma\sqrt{\eta_1 T\ln T}}{n\sqrt{K}} \geq \hat\theta_{t_1(n), 1} + \frac{\sigma\sqrt{\eta_1 T\ln T}}{n\sqrt{K}} \label{eq:event-bad-opt-2}
    \end{align}
    holds simultaneously. The first inequality holds because arm $1$ is eliminated. The second inequality holds because arm $k$ is not eliminated. Now for fixed $n$, 
    \begin{align*}
        & \quad \mbP\left(\text{\eqref{eq:event-bad-opt-2} happens}; \Delta_k \geq \frac{x-2K}{4T}, k\notin\cA_0\right) \\
        & \leq \mbP\left(\exists k': \hat\theta_{t_{k'}(n), k'} - \frac{\sigma\sqrt{\eta_1 T\ln T}}{n\sqrt{K}} \geq \hat\theta_{t_1(n), 1} + \frac{\sigma\sqrt{\eta_1 T\ln T}}{n\sqrt{K}}\right) \\ & \quad\quad \wedge \mbP\left(\hat\theta_{t_k(n), k} + \frac{\sigma\sqrt{\eta_1 T\ln T}}{n\sqrt{K}} \geq \hat\theta_{t_1(n), 1} + \frac{\sigma\sqrt{\eta_1 T\ln T}}{n\sqrt{K}}; \Delta_k \geq \frac{x-2K}{4T}\right) \\
        & \leq \mbP\left(\exists k':\frac{\sum_{m=1}^{n}(\epsilon_{t_{k'}(m), k'} - \epsilon_{t_1(m), 1})}{n} \geq \frac{2\sigma\sqrt{\eta_1 T\ln T}}{n\sqrt{K}}\right) \\
        & \quad \quad \wedge \mbP\left(\frac{\sum_{m=1}^{n}(\epsilon_{t_k(m), k} - \epsilon_{t_1(m), 1})}{n} > \Delta_k; \Delta_k \geq \frac{x-2K}{4T}\right) \\
        & \leq \sum_{k'\neq 1}\left(\mbP\left(\frac{\sum_{m=1}^{n}\epsilon_{t_{k'}(m), k'}}{n} \geq \frac{\sigma\sqrt{\eta_1 T\ln T}}{n\sqrt{K}}\right) + \mbP\left(\frac{\sum_{m=1}^{n}\epsilon_{t_1(m), 1}}{n} \leq - \frac{\sigma\sqrt{\eta_1 T\ln T}}{n\sqrt{K}}\right)\right) \\
        & \quad \quad \wedge \left(\mbP\left(\frac{\sum_{m=1}^{n}\epsilon_{t_{k}(m), k}}{n} \geq \frac{(x-2K)_+}{8T}\right) + \mbP\left(\frac{\sum_{m=1}^{n}\epsilon_{t_1(m), 1}}{n} \leq - \frac{(x-2K)_+}{8T}\right)\right) \\
        & \leq 2K\exp\left(-\frac{\eta_1 T\ln T}{2nK}\right) \wedge 2K\exp\left(-\frac{n(x-2K)_+^2}{128\sigma^2T^2}\right) \\
        & = 2K\exp\left(-\left(\frac{\eta_1 T\ln T}{2nK}\vee \frac{n(x-2K)_+^2}{128\sigma^2T^2} \right)\right) \\
        & \leq 2K\exp\left(-\frac{(x-2K)_+\sqrt{\eta_1 \ln T}}{16\sigma\sqrt{KT}}\right)
    \end{align*}
    Therefore, 
    \begin{align}
        & \quad \mbP((n_k-1)\Delta_k \geq (n_k-1)(x-2K)/4T, \bar S_k, k\notin\cA_0) \nonumber\\
        & =  \mbP\left(\exists n\leq T/2: \text{\eqref{eq:event-bad-opt-2} happens}; (n_k-1)\Delta_k \geq (n_k-1)(x-2K)/4T, k\notin\cA_0\right) \nonumber\\
        & \leq \sum_{n=1}^{\lfloor T/2\rfloor}\mbP\left(\text{\eqref{eq:event-bad-opt-2} happens}; \Delta_k \geq \frac{x-2K}{4T}\right) \nonumber\\
        & \leq KT \exp\left(-\frac{(x-2K)_+\sqrt{\eta_1 \ln T}}{16\sigma\sqrt{KT}}\right).
        \label{eq:bound-pseudo-unidentified-K-opt-2}
    \end{align}
\end{itemize}

Note that the equations above hold for any instance $\theta$. Combining \eqref{eq:bound-noise-K-opt}, \eqref{eq:bound-pseudo-identified-K-opt-1}, \eqref{eq:bound-pseudo-unidentified-K-opt-1}, \eqref{eq:bound-pseudo-identified-K-opt-2}, \eqref{eq:bound-pseudo-unidentified-K-opt-2} yields
\begin{align*}
    & \quad\sup_{\theta}\mbP(\hat R_\theta^\pi(T)\geq x) \\
    & \leq \exp\left(-\frac{x^2}{8K\sigma^2T}\right) + 4K\exp\left(-\frac{(x-2K-8\sigma\sqrt{(\eta_1\vee\eta_2) KT\ln T})_+^2}{128\sigma^2KT}\right)+ 2K^2T\exp\left(-\frac{(x-2K)_+\sqrt{\eta_1\ln T}}{16\sigma\sqrt{KT}}\right)
\end{align*}

\item[(b)] From (a), we have
\begin{align*}
    & \quad \mbP\left(R_\theta^\pi(T)\geq x(1-1/\sqrt{K})\right) \\
    & \leq \sum_{k\neq 1}\left(\mbP\left((n_k-1)\Delta_k\geq \frac{x-2K}{4K}, \ k\in\cA_0\right) + \mbP\left((n_k-1)\Delta_k\geq \frac{(n_k-1)(x-2K)}{4T}, \ k\notin\cA_0\right)\right).
\end{align*}
Fix $k\neq 1$. Now for each $k$, we consider bounding the two terms separately.
\begin{itemize}
    \item $k\in \cA_0$. With a slight abuse of notation, we let $n_0 = \lceil\frac{x-2K}{4K\Delta_k}\rceil\leq n_k-1$. Remember that $t_k(n_0+1)$ is the time period that arm $k$ is pulled for the $(n_0+1)$th time. We have
    \begin{align}
        & \quad \mbP\left((n_k-1)\Delta_k\geq \frac{x-2K}{4K},\ k\in\cA_0\right) \nonumber\\
        & = \mbP\left(n_k \geq 1 + \frac{x-2K}{4K\Delta_k},\ k\in\cA_0\right) \nonumber\\
        & \leq \mbP\left(\hat\theta_{t_k(n_0+1)-1, 1} + \rad(n_{t_k(n_0+1)-1, 1})\leq \hat\theta_{t_k(n_0+1)-1, k} + \rad(n_0)\right) \nonumber\\
        & = \mbP\left(\theta_1 + \frac{\sum_{m=1}^{n_{t_k(n_0+1)-1, 1}}\epsilon_{t_1(m), 1}}{n_{t_k(n_0+1)-1, 1}} + \rad(n_{t_k(n_0+1)-1, 1}) \leq \theta_k + \frac{\sum_{m=1}^{n_0}\epsilon_{t_k(m), k}}{n_0} + \rad(n_0)\right) \nonumber\\
        & \leq \mbP\left(\exists n\in[T]: \theta_1 + \frac{\sum_{m=1}^{n}\epsilon_{t_1(m), 1}}{n} + \rad(n) \leq \theta_k + \frac{\sum_{m=1}^{n_0}\epsilon_{t_k(m), k} }{n_0} + \rad(n_0)\right) \nonumber\\
        & \leq \mbP\left(\exists n\in[T]: \left(\frac{\sum_{m=1}^{n_0}\epsilon_{t_k(m), k}}{n_0} +\rad(n_0)\right) - \left(\frac{\sum_{m=1}^{n}\epsilon_{t_1(m), 1}}{n} + \rad(n)\right) \geq \Delta_k \right) \nonumber\\
        & \leq \mbP\left(\frac{\sum_{m=1}^{n_0}\epsilon_{t_k(m), k}}{n_0} + \rad(n_0) \geq \frac{\Delta_k}{2}\right) + \mbP\left(\exists n\in[T]: \frac{\sum_{m=1}^{n}\epsilon_{t_1(m), 1}}{n} +\rad(n) \leq - \frac{\Delta_k}{2}\right) \nonumber\\
        & \leq \mbP\left(\frac{\sum_{m=1}^{n_0}\epsilon_{t_k(m), k}}{n_0} + \frac{\sigma\sqrt{(\eta_1\vee\eta_2)T\ln T}}{n_0\sqrt{K}} \geq \frac{\Delta_k}{2}\right) + \sum_{n=1}^{T}\mbP\left(\frac{\sum_{m=1}^{n}-\epsilon_{t_1(m), 1}}{n} \geq \frac{\Delta_k}{2} + \frac{\sigma\sqrt{\eta_1 T\ln T}}{n\sqrt{K}}\right) \nonumber\\
        & \leq \mbP\left(\frac{\sum_{m=1}^{n_0}\epsilon_{t_k(m), k}}{n_0} + \frac{\sigma\sqrt{(\eta_1\vee\eta_2)T\ln T}}{n_0\sqrt{K}} \geq \frac{\Delta_k}{2}\right) \nonumber\\
        & \quad\quad + \sum_{n=1}^{T}\mbP\left(\frac{\sum_{m=1}^{n}-\epsilon_{t_1(m), 1}}{n} \geq \sqrt{\frac{(x-2K)_+\sigma\sqrt{\eta_1 T\ln T}}{2n\sqrt{K}T}}\right) \nonumber\\
        & \leq \exp\left(-\frac{(x-2K-8\sigma\sqrt{(\eta_1\vee\eta_2) KT\ln T})_+^2}{128\sigma^2KT}\right) + T\exp\left(-\frac{(x-2K)_+\sqrt{\eta_1\ln T}}{16\sigma\sqrt{KT}}\right). \label{eq:bound-pseudo-K-opt-1-ucb}
    \end{align}
    The last inequality holds from \eqref{eq:bound-pseudo-identified-K-opt-1} and \eqref{eq:bound-pseudo-unidentified-K-opt-1}.
    \item $k\notin \cA_0$. With a slight abuse of notation, we let $n_0 = \lceil\frac{T}{K}\rceil\leq n_k - 1$. Remember that $t_k(n_0+1)$ is the time period that arm $k$ is pulled for the $(n_0+1)$th time. We have
    \begin{align}
        & \quad \mbP\left((n_k-1)\Delta_k\geq \frac{(n_k-1)(x-2K)}{4T}, \ k\notin\cA_0\right) \nonumber\\
        & = \mbP\left(\Delta_k\geq\frac{x-2K}{4T}, \ k\notin\cA_0\right) \nonumber\\
        & \leq \mbP\left(\hat\theta_{t_k(n_0+1)-1, 1} + \rad(n_{t_k(n_0+1)-1, 1})\leq \hat\theta_{t_k(n_0+1)-1, k} + \rad(n_0)\right) \nonumber\\
        & = \mbP\left(\theta_1 + \frac{\sum_{m=1}^{n_{t_k(n_0+1)-1, 1}}\epsilon_{t_1(m), 1}}{n_{t_k(n_0+1)-1, 1}} + \rad(n_{t_k(n_0+1)-1, 1}) \leq \theta_k + \frac{\sum_{m=1}^{n_0}\epsilon_{t_k(m), k}}{n_0} + \rad(n_0)\right) \nonumber\\
        & \leq \mbP\left(\exists n\in[T]: \theta_1 + \frac{\sum_{m=1}^{n}\epsilon_{t_1(m), 1}}{n} + \rad(n) \leq \theta_k + \frac{\sum_{m=1}^{n_0}\epsilon_{t_k(m), k} }{n_0} + \rad(n_0)\right) \nonumber\\
        & \leq \mbP\left(\exists n\in[T]: \left(\frac{\sum_{m=1}^{n_0}\epsilon_{t_k(m), k}}{n_0} +\rad(n_0)\right) - \left(\frac{\sum_{m=1}^{n}\epsilon_{t_1(m), 1}}{n} + \rad(n)\right) \geq \Delta_k \right) \nonumber\\
        & \leq \mbP\left(\frac{\sum_{m=1}^{n_0}\epsilon_{t_k(m), k}}{n_0} +\rad(n_0) \geq \frac{\Delta_k}{2}\right) + \mbP\left(\exists n\in[T]: \frac{\sum_{m=1}^{n}\epsilon_{t_1(m), 1}}{n} + \rad(n) \leq - \frac{\Delta_k}{2}\right) \nonumber\\
        & \leq \exp\left(-\frac{\left(x-2K-8\sigma\sqrt{(\eta_1\vee\eta_2)KT\ln T}\right)_+^2}{128\sigma^2KT}\right) + T\exp\left(-\frac{(x-2K)_+\sqrt{\eta_1\ln T}}{16\sigma\sqrt{KT}}\right). \label{eq:bound-pseudo-K-opt-2-ucb}
    \end{align}
    The last inequality holds from \eqref{eq:bound-pseudo-identified-K-opt-2} and \eqref{eq:bound-pseudo-unidentified-K-opt-2}. Note that the equations above hold for any instance $\theta$. Combining \eqref{eq:bound-noise-K-opt}, \eqref{eq:bound-pseudo-K-opt-1-ucb}, \eqref{eq:bound-pseudo-K-opt-2-ucb} yields
    \begin{align*}
        & \quad\sup_{\theta}\mbP(\hat R_\theta^\pi(T)\geq x) \\
        & \leq \exp\left(-\frac{x^2}{8K\sigma^2T}\right) + 4K\exp\left(-\frac{(x-2K-8\sigma\sqrt{(\eta_1\vee\eta_2) KT\ln T})_+^2}{128\sigma^2KT}\right) + 2K^2T\exp\left(-\frac{(x-2K)_+\sqrt{\eta_1\ln T}}{16\sigma\sqrt{KT}}\right).
    \end{align*}
\end{itemize}
\end{itemize}
\end{enumerate}

$\hfill\Box$

{\noindent\bf Proof of Theorem \ref{thm:any-time}.} Fix a time horizon of $T$. We write $t_k = t_k(n_{T, k})$ as the last time that arm $k$ is pulled throughout the $T$ time periods. By the definition of $n_k$ and $t_k$, the following event happens w.p. $1$:
\begin{align*}
    \hat\theta_{t_k-1, 1} + \rad_{t_k}(n_{t_k-1, 1})\leq \hat\theta_{t_k-1, k} + \rad_{t_k}(n_{t_k-1, k})
\end{align*}
Define
\begin{align*}
    \cA_1 = \left\{k\neq 1: n_k\leq 1 + \frac{t_k^{3/4}T^{1/4}}{K}\right\}.
\end{align*}
Fix $x\geq 2K$. We have
\begin{align*}
    & \quad\ \mbP\left(R_\theta^\pi(T)\geq x(1-1/(2\sqrt{K}))\right) \\
    & = \mbP\left(\sum_{k\in \cA_1}n_k\Delta_k + \sum_{k\notin \cA_1}n_k\Delta_k\geq x(1-1/(2\sqrt{K}))\right) \\
    & \leq \mbP\left(\sum_{k\in \cA_1}(n_k-1)\Delta_k + \sum_{k\notin \cA_1}(n_k-1)\Delta_k\geq x(1-1/(2\sqrt{K})) - K\right) \\
    & \leq \mbP\left(\left(\bigcup_{k\in\cA_1}\left\{(n_k-1)\Delta_k\geq\frac{x-2K}{4K}\right\}\right)\bigcup\left(\bigcup_{k\notin\cA_1}\left\{(n_k-1)\Delta_k\geq \frac{(n_k-1)(x-2K)}{4\sqrt{t_k T}}\right\}\right)\right) \\
    & \leq \sum_{k\neq 1}\left(\mbP\left((n_k-1)\Delta_k\geq \frac{x-2K}{4K}, \ k\in\cA_1\right) + \mbP\left((n_k-1)\Delta_k\geq \frac{(n_k-1)(x-2K)}{4\sqrt{t_k T}}, \ k\notin\cA_1\right)\right) \\
    & \leq \sum_{k\neq 1}\left(\mbP\left((n_k-1)\Delta_k\geq \frac{x-2K}{4K}, \ k\in\cA_1\right) + \mbP\left(\Delta_k\geq \frac{x-2K}{4\sqrt{t_k T}}, \ k\notin\cA_1\right)\right)
\end{align*}
The reason that the second inequality holds is as follows. To prove it, we only need to show that the following cannot holds:
\begin{align*}
    (n_k-1)\Delta_k < \frac{x-2K}{4K}, \quad \forall k\in\cA_1; \quad \quad (n_k-1)\Delta_k < \frac{(n_k-1)(x-2K)}{8\sqrt{t_k T}}, \quad \forall k\notin\cA_1.
\end{align*}
If not, then we have
\begin{align*}
    \sum_{k\neq 1} (n_k-1)\Delta_k & = \sum_{k\in\cA_1}(n_k-1)\Delta_k + \sum_{k\notin\cA_1}(n_k-1)\Delta_k \\
    & < \frac{(x-2K)|\cA_1|}{4K} + \frac{x-2K}{8\sqrt{T}}\sum_{k\notin\cA_1}\frac{n_k}{\sqrt{t_k}} \\
    & \leq \frac{x-2K}{4} + \frac{x-2K}{8\sqrt{T}}\sum_{k\notin\cA_1}\frac{n_k}{\sqrt{t_k}} \\
    & \leq \frac{x-2K}{2}\leq x(1-1/(2\sqrt{K})) - K.
\end{align*}
In fact, to bound $\sum_{k\notin\cA_1}\frac{n_k}{\sqrt{t_k}}$, we can assume $0=t_{k_0} < t_{k_1} < t_{k_2} < \cdots$. Then we have
\begin{align*}
    t_{k_i} \geq n_{k_1} + \cdots + n_{k_i}
\end{align*}
because before up to time $t_{k_i}$, arms $k_1, \cdots, k_i$ have been pulled completely, and after time $t_{k_i}$ none of them will be pulled. Thus, 
\begin{align*}
    \sum_{k\notin\cA_1}\frac{n_k}{\sqrt{t_k}} & = \sum_{i=1}^{|\cA_1|}\frac{t_{k_i} - t_{k_{i-1}}}{\sqrt{t_{k_i}}} \leq 2\sum_{i=1}^{|\cA_1|}\frac{t_{k_i} - t_{k_{i-1}}}{\sqrt{t_{k_i}} + \sqrt{t_{k_{i-1}}}} = 2\sqrt{t_{k_{|\cA_1|}}} \leq 2\sqrt{T}.
\end{align*}

Now fix $k\neq 1$. For each $k$, we consider bounding the two terms separately.
\begin{itemize}
    \item $k\in \cA_1$. Remember that $n_k$ is the last time period that arm $k$ is pulled. Then from $k\in\cA_1$, we know
    \begin{align*}
        \frac{t_k^{3/4}T^{1/4}}{K} \geq n_k - 1 \geq \frac{x-2K}{4K\Delta_k}
    \end{align*}
    Thus, 
    \begin{align*}
        \Delta_k \geq \frac{x-2K}{4t_k^{3/4}T^{1/4}}.
    \end{align*}
    We have
    \begin{align*}
        & \quad \mbP\left((n_k-1)\Delta_k\geq \frac{x-2K}{4K},\ k\in\cA_1\right) \nonumber\\
        & = \mbP\left(n_k \geq 1 + \frac{x-2K}{4K\Delta_k},\ k\in\cA_1\right) \nonumber\\
        & = \mbP\left(\hat\theta_{t_k-1, 1} + \rad_{t_k}(n_{t_k-1, 1})\leq \hat\theta_{t_k-1, k} + \rad_{t_k}(n_{t_k-1, k}); \frac{t_k^{3/4}T^{1/4}}{K} \geq n_k - 1 \geq \frac{x-2K}{4K\Delta_k}\right) \nonumber\\
        & = \mbP\left(\theta_1 + \frac{\sum_{m=1}^{n_{t_k-1, 1}}\epsilon_{t_1(m), 1}}{n_{t_k-1, 1}} + \rad_{t_k}(n_{t_k-1, 1}) \leq \theta_k + \frac{\sum_{m=1}^{n_k-1}\epsilon_{t_k(m), k}}{n_k-1} + \rad_{t_k}(n_k-1); \right.\nonumber\\ 
        & \quad \quad \left.\frac{t_k^{3/4}T^{1/4}}{K} \geq n_k - 1 \geq \frac{x-2K}{4K\Delta_k}\right) \nonumber\\
        & \leq \mbP\left(\exists n\in[T]: \theta_1 + \frac{\sum_{m=1}^{n}\epsilon_{t_1(m), 1}}{n} + \rad_{t_k}(n) \leq \theta_k + \frac{\sum_{m=1}^{n_k-1}\epsilon_{t_k(m), k}}{n_k-1} + \rad_{t_k}(n_k-1); \right.\nonumber\\ 
        & \quad \quad \left.\frac{t_k^{3/4}T^{1/4}}{K} \geq n_k - 1 \geq \frac{x-2K}{4K\Delta_k}\right) \nonumber\\
        & \leq \sum_{\substack{n_0, t:\\ Kn_0 \leq t^{3/4}T^{1/4}}}\mbP\left(\exists n\in[T]: \left(\frac{\sum_{m=1}^{n_0}\epsilon_{t_k(m), k}}{n_0} +\rad_t(n_0)\right) - \left(\frac{\sum_{m=1}^{n}\epsilon_{t_1(m), 1}}{n} + \rad_{t}(n)\right) \geq \frac{x-2K}{4Kn_0}\right)
    \end{align*}
    Note that here $n_k$ and $t_k$ are both random variables, so we need to decompose the probability by $n_k-1=n_0$ and $t_k=t$ through all possible $(n_0, t)$. Now for any $n_0$ and $t$ such that $Kn_0 \leq t^{3/4}T^{1/4}$, we have
    \begin{align*}
        & \quad \mbP\left(\exists n\in[T]: \left(\frac{\sum_{m=1}^{n_0}\epsilon_{t_k(m), k}}{n_0} +\rad_t(n_0)\right) - \left(\frac{\sum_{m=1}^{n}\epsilon_{t_1(m), 1}}{n} + \rad_{t}(n)\right) \geq \frac{x-2K}{4Kn_0}\right) \nonumber\\
        & \leq \mbP\left(\frac{\sum_{m=1}^{n_0}\epsilon_{t_k(m), k}}{n_0} + \rad_t(n_0) \geq \frac{x-2K}{8Kn_0}\right) + \mbP\left(\exists n\in[T]: \frac{\sum_{m=1}^{n}\epsilon_{t_1(m), 1}}{n} +\rad_{t}(n)\leq - \frac{x-2K}{8Kn_0}\right) \nonumber\\
        & \leq \mbP\left(\frac{\sum_{m=1}^{n_0}\epsilon_{t_k(m), k}}{n_0} + \frac{\sigma\sqrt{\eta t(1\vee\ln Kt)}}{n_0\sqrt{K}} \geq \frac{x-2K}{8Kn_0}\right) + \sum_{n=1}^{T}\mbP\left(\frac{\sum_{m=1}^{n}-\epsilon_{t_1(m), 1}}{n} \geq \frac{x-2K}{8Kn_0} + \frac{\sigma\sqrt{\eta t\ln (Kt)}}{n\sqrt{K}}\right) \nonumber\\
        & \leq \mbP\left(\frac{\sum_{m=1}^{n_0}\epsilon_{t_k(m), k}}{n_0} + \frac{\sigma\sqrt{2\eta t\ln T}}{n_0\sqrt{K}} \geq \frac{x-2K}{8Kn_0}\right) + \sum_{n=1}^{T}\mbP\left(\frac{\sum_{m=1}^{n}-\epsilon_{t_1(m), 1}}{n} \geq \sqrt{\frac{(x-2K)_+\sigma\sqrt{\eta\ln (2t)}}{2n\sqrt{K}t^{1/4}T^{1/4}}}\right) \nonumber\\
        & \leq \exp\left(-n_0\left(\frac{x-2K}{8Kn_0} - \frac{\sigma\sqrt{2\eta t\ln T}}{n_0\sqrt{K}}\right)_+^2\big/(2\sigma^2)\right) + \sum_{n=1}^T\exp\left(-\frac{(x-2K)_+\sqrt{\eta\ln T}}{8\sigma\sqrt{KT}}\right) \nonumber\\
        & \leq \exp\left(-\frac{1}{n_0}\left(\frac{x-2K}{8K} - \frac{\sigma\sqrt{2\eta t\ln T}}{\sqrt{K}}\right)_+^2\big/(2\sigma^2)\right) + T\exp\left(-\frac{(x-2K)_+\sqrt{\eta\ln T}}{8\sigma\sqrt{KT}}\right) \nonumber\\
        & \leq \exp\left(-\frac{K}{T}\left(\frac{x-2K}{8K} - \frac{\sigma\sqrt{2\eta T\ln T}}{\sqrt{K}}\right)_+^2\big/(2\sigma^2)\right) + T\exp\left(-\frac{(x-2K)_+\sqrt{\eta\ln T}}{8\sigma\sqrt{KT}}\right) \nonumber\\
        & \leq \exp\left(-\frac{(x-2K-8\sigma\sqrt{2\eta KT\ln T})_+^2}{128\sigma^2KT}\right) + T\exp\left(-\frac{(x-2K)_+\sqrt{\eta\ln T}}{8\sigma\sqrt{KT}}\right).
    \end{align*}
    Note that here we use the fact that for any $1\leq t\leq T$, 
    \begin{align*}
        \frac{\ln(2t)}{\sqrt{t}} \geq \frac{1}{4}\cdot \frac{\ln T}{\sqrt{T}}.
    \end{align*}
    So we have
    \begin{align}
        & \quad \mbP\left((n_k-1)\Delta_k\geq \frac{x-2K}{4K},\ k\in\cA_1\right) \nonumber\\
        & \leq \sum_{\substack{n_0, t:\\ Kn_0\leq t^{3/4}T^{1/4}}}\mbP\left(\exists n\in[T]: \left(\frac{\sum_{m=1}^{n_0}\epsilon_{t_k(m), k}}{n_0} +\rad_T(n_0)\right) - \left(\frac{\sum_{m=1}^{n}\epsilon_{t_1(m), 1}}{n} + \rad_{Kn_0}(n)\right) \geq \frac{x-2K}{4Kn_0}\right) \nonumber\\
        & \leq T^2\exp\left(-\frac{(x-2K-8\sigma\sqrt{2\eta KT\ln T})_+^2}{128\sigma^2KT}\right) + T^3\exp\left(-\frac{(x-2K)_+\sqrt{\eta\ln T}}{8\sigma\sqrt{KT}}\right) \label{eq:bound-pseudo-K-opt-1-ucb-any}
    \end{align}
    \item $k\notin \cA_1$. Remember that $t_k$ is the last time period that arm $k$ is pulled. Then from $k\notin\cA_1$, we know
    \begin{align*}
        t_k\geq n_k \geq 1 + \frac{t_k^{3/4}T^{1/4}}{K}.
    \end{align*}
    We have
    \begin{align*}
        & \quad \mbP\left((n_k-1)\Delta_k\geq \frac{(n_k-1)(x-2K)}{8\sqrt{t_k T}}, \ k\notin\cA_1\right) \nonumber\\
        & = \mbP\left(\Delta_k\geq\frac{x-2K}{8\sqrt{t_k T}}, \ n_k \geq 1 + \frac{t_k^{3/4}T^{1/4}}{K}\right) \nonumber\\
        & = \mbP\left(\hat\theta_{t_k-1, 1} + \rad_{t_k}(n_{t_k-1, 1})\leq \hat\theta_{t_k-1, k} + \rad_{t_k}(n_k-1); \Delta_k\geq\frac{x-2K}{8\sqrt{t_k T}}, \ n_k \geq 1 + \frac{t_k^{3/4}T^{1/4}}{K} \right) \nonumber\\
        & = \mbP\left(\theta_1 + \frac{\sum_{m=1}^{n_{t_k-1, 1}}\epsilon_{t_1(m), 1}}{n_{t_k-1, 1}} + \rad_{t_k}(n_{t_k-1, 1}) \leq \theta_k + \frac{\sum_{m=1}^{n_k-1}\epsilon_{t_k(m), k}}{n_k-1} + \rad_{t_k}(n_k-1); \right.\nonumber\\
        & \quad \quad \quad \left.\Delta_k\geq\frac{x-2K}{8\sqrt{t_k T}}, \ n_k \geq 1 + \frac{t_k^{3/4}T^{1/4}}{K}\right) \nonumber\\
        & \leq \mbP\left(\exists n\in[T]: \theta_1 + \frac{\sum_{m=1}^{n}\epsilon_{t_1(m), 1}}{n} + \rad_{t_k}(n) \leq \theta_k + \frac{\sum_{m=1}^{n_k-1}\epsilon_{t_k(m), k} }{n_k-1} + \rad_{t_k}(n_k-1); \right.\nonumber\\
        & \quad \quad \quad \left.\Delta_k\geq\frac{x-2K}{8\sqrt{t_k T}}, \ n_k - 1 \geq \frac{t_k^{3/4}T^{1/4}}{K}\right) \nonumber\\
        & \leq \sum_{\substack{n_0\leq t:\\ Kn_0\geq t^{3/4}T^{1/4}}}\mbP\left(\exists n\in[T]: \left(\frac{\sum_{m=1}^{n_0}\epsilon_{t_k(m), k}}{n_0} +\rad_t(n_0)\right) - \left(\frac{\sum_{m=1}^{n}\epsilon_{t_1(m), 1}}{n} + \rad_t(n)\right) \geq \frac{x-2K}{8\sqrt{tT}} \right)
    \end{align*}
    Note that here $n_k$ and $t_k$ are both random variables, so we need to decompose the probability by $n_k-1=n_0$ and $t_k=t$ through all possible $(n_0, t)$. Now for any $(n_0, t)$ such that $t\geq n_0$ and $Kn_0\geq t^{3/4}T^{1/4}$, we know that $Kt \geq Kn_0\geq T^{1/4}$, and so
    \begin{align*}
        \ln (Kt) \geq \frac{1}{4}\ln T.
    \end{align*}
    We have
    \begin{align*}
        & \quad \mbP\left(\exists n\in[T]: \left(\frac{\sum_{m=1}^{n_0}\epsilon_{t_k(m), k}}{n_0} +\rad_t(n_0)\right) - \left(\frac{\sum_{m=1}^{n}\epsilon_{t_1(m), 1}}{n} + \rad_t(n)\right) \geq \frac{x-2K}{8\sqrt{tT}} \right) \nonumber\\
        & \leq \mbP\left(\frac{\sum_{m=1}^{n_0}\epsilon_{t_k(m), k}}{n_0} +\rad_t(n_0) \geq \frac{x-2K}{16\sqrt{tT}}\right) + \mbP\left(\exists n\in[T]: \frac{\sum_{m=1}^{n}\epsilon_{t_1(m), 1}}{n} + \rad_t(n) \leq - \frac{x-2K}{16\sqrt{tT}}\right) \nonumber\\
        & \leq \mbP\left(\frac{\sum_{m=1}^{n_0}\epsilon_{t_k(m), k}}{n_0} + \frac{\sigma\sqrt{\eta t\ln(KT)}}{n_0\sqrt{K}} \geq \frac{x-2K}{16\sqrt{tT}}\right) + \sum_{n=1}^{T}\mbP\left(\frac{\sum_{m=1}^{n}-\epsilon_{t_1(m), 1}}{n} \geq \frac{x-2K}{16\sqrt{tT}} + \frac{\sigma\sqrt{\eta t\ln (Kt)}}{n\sqrt{K}}\right) \nonumber\\
        & \leq \mbP\left(\frac{\sum_{m=1}^{n_0}\epsilon_{t_k(m), k}}{n_0} + \frac{\sigma\sqrt{2\eta t\ln T}}{n_0\sqrt{K}} \geq \frac{x-2K}{16\sqrt{tT}}\right) + \sum_{n=1}^{T}\mbP\left(\frac{\sum_{m=1}^{n}-\epsilon_{t_1(m), 1}}{n} \geq \sqrt{\frac{(x-2K)_+\sigma\sqrt{\eta\ln T}}{8n\sqrt{KT}}}\right) \nonumber\\
        & \leq \exp\left(-n_0\left(\frac{x-2K}{16\sqrt{tT}} - \frac{\sqrt{2\eta t\ln T}}{n_0\sqrt{K}}\right)_+^2\big/(2\sigma^2)\right) + \sum_{n=1}^T\exp\left(-\frac{(x-2K)_+\sqrt{\eta\ln T}}{16\sigma\sqrt{KT}}\right) \nonumber\\
        & \leq \exp\left(-\frac{t}{K}\left(\frac{x-2K}{16\sqrt{tT}} - \frac{\sqrt{2\eta t\ln T}}{t/\sqrt{K}}\right)_+^2\big/(2\sigma^2)\right) + \sum_{n=1}^T\exp\left(-\frac{(x-2K)_+\sqrt{\eta\ln T}}{16\sigma\sqrt{KT}}\right) \nonumber\\
        & \leq \exp\left(-\frac{(x-2K-16\sigma\sqrt{2\eta KT})_+^2}{512\sigma^2KT}\right) + T\exp\left(-\frac{(x-2K)_+\sqrt{\eta\ln T}}{16\sigma\sqrt{KT}}\right).
    \end{align*}
    So we have
    \begin{align}
        & \quad \mbP\left((n_k-1)\Delta_k\geq \frac{(n_k-1)(x-2K)}{8\sqrt{t_k T}}, \ k\notin\cA_1\right) \nonumber\\
        & \leq \sum_{\substack{n_0, t:\\ n_0\geq t/K}}\mbP\left(\exists n\in[T]: \left(\frac{\sum_{m=1}^{n_0}\epsilon_{t_k(m), k}}{n_0} +\rad_t(n_0)\right) - \left(\frac{\sum_{m=1}^{n}\epsilon_{t_1(m), 1}}{n} + \rad_t(n)\right) \geq \frac{x-2K}{8\sqrt{tT}} \right) \nonumber\\
        & \leq T^2\exp\left(-\frac{(x-2K-16\sigma\sqrt{2\eta KT})_+^2}{512\sigma^2KT}\right) + T^3\exp\left(-\frac{(x-2K)_+\sqrt{\eta\ln T}}{16\sigma\sqrt{KT}}\right) \label{eq:bound-pseudo-K-opt-2-ucb-any}
    \end{align}
    Note that all the equations above hold for any instance $\theta$. Combining \eqref{eq:bound-noise-K-opt}, \eqref{eq:bound-pseudo-K-opt-1-ucb-any}, \eqref{eq:bound-pseudo-K-opt-2-ucb-any} yields
    \begin{align*}
        & \quad\sup_{\theta}\mbP(R_\theta^\pi(T)\geq x) \\
        & \leq \exp\left(-\frac{x^2}{8K\sigma^2T}\right) + 2KT^2\exp\left(-\frac{(x-2K-16\sigma\sqrt{2\eta KT\ln T})_+^2}{512\sigma^2KT}\right) + 2KT^3\exp\left(-\frac{(x-2K)_+\sqrt{\eta\ln T}}{16\sigma\sqrt{KT}}\right).
    \end{align*}
\end{itemize}

$\hfill\Box$

{\noindent\bf Proof of Theorem \ref{thm:linear}.} To simplify notations, we write $\Delta_t\triangleq \theta^\top(a_t^*-a_t)$. Also, we write
\begin{align*}
    A_t = [a_1, \cdots, a_t], \quad R_t = [r_1, \cdots, r_t]^\top, \quad \mathcal E_t = [\epsilon_{1, a_1}, \cdots, \epsilon_{t, a_t}]^\top.
\end{align*}
Then
\begin{align*}
    \hat\theta_t & = V_t^{-1}A_tR_t = V_t^{-1}A_t(A_t^\top\theta+\mathcal E_t) = \theta - V_t^{-1}\theta + V_t^{-1}A_t\mathcal E_t.
\end{align*}
Note that
\begin{align*}
    R_\theta^\pi(T) = \sum_{t}\Delta_t = \sum_{t}\frac{\Delta_t}{a_t^\top V_{t-1}^{-1}a_t}\cdot a_t^\top V_{t-1}^{-1}a_t
\end{align*}
and from Lemma 11 in \cite{abbasi2011improved}, 
\begin{align*}
    \sum_{t}a_t^\top V_{t-1}^{-1}a_t & \leq 2\ln\det V_{T-1} - 2\ln\det V_1 \leq 2d\ln\left(\frac{tr(V_{T-1})}{d}\right) \leq 2d\ln\frac{T}{d} \leq 2d\ln T.
\end{align*}
Another fact we will be using in the proof is from Theorem 1 in \cite{abbasi2011improved}, where it is shown that for any $\delta>0$, w.p. at least $1-\delta$, the following holds:
\begin{align*}
    (A_{t-1}\mathcal E_{t-1})^\top  V_{t-1}^{-1}A_{t-1}\mathcal E_{t-1} \leq 2\sigma^2\ln\left(\frac{\det(V_{t-1})/\det(V_0)}{\delta}\right) \leq 2\sigma^2\ln\left(\frac{(T/d)^{2d}}{\delta}\right)
\end{align*}
Thus, for any $y\geq 0$, we have
\begin{align*}
    \mbP\left(\sqrt{(A_{t-1}\mathcal E_{t-1})^\top  V_{t-1}^{-1}A_{t-1}\mathcal E_{t-1}} \geq x\right) \leq (T/d)^{2d}\exp\left(-\frac{x^2}{2\sigma^2}\right)
\end{align*}
We have
\begin{align*}
    \sup_{\theta}\mbP(\hat R_\theta^\pi(T)\geq x) & \leq \mbP\left(N^\pi(T) \leq -x/d\right) + \sup_{\theta}\mbP\left(R_\theta^\pi(T) \geq x(1-1/d)\right) \\
    & \leq \exp\left(-\frac{x^2}{2\sigma^2d^2T}\right) + \sup_{\theta}\mbP\left(R_\theta^\pi(T) \geq x/2\right)
\end{align*}
Also, for any $\theta$, 
\begin{align} \label{eq:linear-decompose}
    \mbP(R_\theta^\pi(T) \geq x/2) & \leq \mbP\left(\bigcup_{t\geq 2}\left\{\Delta_t \geq \frac{x-4\sqrt{d}}{4t\ln T},\ a_t^\top V_{t-1}^{-1}a_t\leq d/t \right\}\right) \nonumber\\
    & \quad\quad + \mbP\left(\bigcup_{t\geq 2}\left\{\frac{\Delta_t}{a_t^\top V_{t-1}^{-1}a_t}\geq \frac{x-4\sqrt{d}}{8d\ln T}, \ a_t^\top V_{t-1}^{-1}a_t > d/t\right\}\right)
\end{align}
The reason that \eqref{eq:linear-decompose} holds is as follows. To prove it, we only need to show that the following events cannot hold simultaneously:
\begin{align*}
    \Delta_t < \frac{x-4\sqrt{d}}{4t\ln T},\ \text{ if }a_t^\top V_{t-1}^{-1}a_t\leq d/t; \quad \frac{\Delta_t}{a_t^\top V_{t-1}^{-1}a_t} < \frac{x-4\sqrt{d}}{8d\ln T}, \ \text{ if }a_t^\top V_{t-1}^{-1}a_t > d/t.
\end{align*}
If not, then
\begin{align*}
    R_{\theta}^\pi(T) & = \theta^\top(a_1^*-a_1) + \sum_{t\geq 2} \Delta_t\mathds 1\{a_t^\top V_{t-1}^{-1}a_t\leq d/t\} \\
    & \quad \quad + \frac{\Delta_t}{a_t^\top V_{t-1}^{-1}a_t}\cdot (a_t^\top V_{t-1}^{-1}a_t)\mathds 1\{a_t^\top V_{t-1}^{-1}a_t > d/t\} \\
    & < 2\sqrt{d} + \sum_{t\geq 2}\frac{x-4\sqrt{d}}{4t\ln T} + \sum_{t\geq 2}\frac{x-4\sqrt{d}}{8d\ln T}a_t^\top V_{t-1}^{-1}a_t \\
    & \leq 2\sqrt{d} + \frac{x}{4} - \sqrt{d} + \frac{x}{4} - \sqrt{d} = x/2.
\end{align*}
This is a contradiction.
At time $t$, the policy takes action $a_t$, which means
\begin{align}
    & \quad\ \quad\ \hat\theta_{t-1}^\top a_t + (a_t^\top V_{t-1}^{-1}a_t)\sigma\sqrt{\frac{\eta t}{d}} + \sqrt{d(a_t^\top V_{t-1}^{-1}a_t)} \geq \nonumber\\
    & \quad\quad\quad \hat\theta_{t-1}^\top a_t^* + (a_t^{*\top} V_{t-1}^{-1}a_t^{*})\sigma\sqrt{\frac{\eta t}{d}} + \sqrt{d(a_t^{*\top} V_{t-1}^{-1}a_t^{*})} \nonumber\\
    & \Leftrightarrow\quad \theta^\top a_t - \theta^\top V_{t-1}^{-1}a_t + (V_{t-1}^{-1}A_{t-1}\mathcal E_{t-1})^\top a_t + (a_t^\top V_{t-1}^{-1}a_t)\sigma\sqrt{\frac{\eta t}{d}} + \sqrt{d(a_t^\top V_{t-1}^{-1}a_t)} \geq \nonumber\\
    & \quad\quad\quad \theta^\top a_t^* - \theta^\top V_{t-1}^{-1}a_t^* + (V_{t-1}^{-1}A_{t-1}\mathcal E_{t-1})^\top a_t^* + (a_t^{*\top} V_{t-1}^{-1}a_t^{*})\sigma\sqrt{\frac{\eta t}{d}} + \sqrt{d(a_t^{*\top} V_{t-1}^{-1}a_t^{*})} \nonumber\\
    & \Leftrightarrow\quad (V_{t-1}^{-1}A_{t-1}\mathcal E_{t-1})^\top a_t + (a_t^\top V_{t-1}^{-1}a_t)\sigma\sqrt{\frac{\eta t}{d}} + \sqrt{d(a_t^\top V_{t-1}^{-1}a_t)} - \theta^\top V_{t-1}^{-1}a_t \geq \nonumber\\
    & \quad\quad\quad \Delta_t + (V_{t-1}^{-1}A_{t-1}\mathcal E_{t-1})^\top a_t^* + (a_t^{*\top} V_{t-1}^{-1}a_t^{*})\sigma\sqrt{\frac{\eta t}{d}} + \sqrt{d(a_t^{*\top} V_{t-1}^{-1}a_t^{*})} - \theta^\top V_{t-1}^{-1}a_t^* \nonumber\\
    & \Rightarrow\quad a_t^\top V_{t-1}^{-1}A_{t-1}\mathcal E_{t-1} + (a_t^\top V_{t-1}^{-1}a_t)\sigma\sqrt{\frac{\eta t}{d}} + 2\sqrt{d(a_t^\top V_{t-1}^{-1}a_t)} \geq \nonumber\\
    & \quad\quad\quad \Delta_t + a_t^{*\top}V_{t-1}^{-1}A_{t-1}\mathcal E_{t-1} + (a_t^{*\top} V_{t-1}^{-1}a_t^{*})\sigma\sqrt{\frac{\eta t}{d}} \nonumber\\
    & \Rightarrow\quad a_t^\top V_{t-1}^{-1}A_{t-1}\mathcal E_{t-1} \geq \frac{\Delta_t}{2} - (a_t^\top V_{t-1}^{-1}a_t)\sigma\sqrt{\frac{\eta t}{d}} - 2\sqrt{d(a_t^\top V_{t-1}^{-1}a_t)} \quad \text{or} \nonumber\\
    & \quad\quad\quad -a_t^{*\top}V_{t-1}^{-1}A_{t-1}\mathcal E_{t-1} \geq \frac{\Delta_t}{2} + (a_t^{*\top}V_{t-1}^{-1}a_t^{*})\sigma\sqrt{\frac{\eta t}{d}} \label{eq:linear-action}
\end{align}
Note that in \eqref{eq:linear-action} we use the following inequality: for any $a\in\cA_t$, 
\begin{align*}
    |\theta^\top V_{t-1}^{-1}a| \leq \sqrt{\theta^\top V_{t-1}^{-1}\theta}\sqrt{a^\top V_{t-1}^{-1}a} \leq \sqrt{d(a^\top V_{t-1}^{-1}a)}
\end{align*}
Combining \eqref{eq:linear-decompose} and \eqref{eq:linear-action} yields
\begin{align*}
    & \quad \mbP\left(R_\theta^\pi(T) \geq x/2\right) \\
    & \leq \sum_{t}\mbP\left(\Delta_t \geq \frac{x-4\sqrt{d}}{4t\ln T},\ a_t^\top V_{t-1}^{-1}a_t\leq d/t,\ a_t^\top V_{t-1}^{-1}A_{t-1}\mathcal E_{t-1} \geq \frac{\Delta_t}{2} - (a_t^\top V_{t-1}^{-1}a_t)\sigma\sqrt{\frac{\eta t}{d}} - 2\sqrt{d(a_t^\top V_{t-1}^{-1}a_t)}\right) \\
    & + \sum_t\mbP\left(\Delta_t \geq \frac{x-4\sqrt{d}}{4t\ln T},\ a_t^\top V_{t-1}^{-1}a_t\leq d/t,\ -a_t^{*\top}V_{t-1}^{-1}A_{t-1}\mathcal E_{t-1} \geq \frac{\Delta_t}{2} + (a_t^{*\top}V_{t-1}^{-1}a_t^*)\sigma\sqrt{\frac{\eta t}{d}}\right) \\
    & + \sum_{t}\mbP\left(\frac{\Delta_t}{a_t^\top V_{t-1}^{-1}a_t}\geq \frac{x-4\sqrt{d}}{8d\ln T},\ a_t^\top V_{t-1}^{-1}a_t > \frac{d}{t},  \right.\nonumber\\
    & \quad\quad\quad\quad\quad\left.a_t^\top V_{t-1}^{-1}A_{t-1}\mathcal E_{t-1} \geq \frac{\Delta_t}{2} - (a_t^\top V_{t-1}^{-1}a_t)\sigma\sqrt{\frac{\eta t}{d}} - 2\sqrt{d(a_t^\top V_{t-1}^{-1}a_t)}\right) \\
    & + \sum_t\mbP\left(\frac{\Delta_t}{a_t^\top V_{t-1}^{-1}a_t}\geq \frac{x-4\sqrt{d}}{8d\ln T},\ a_t^\top V_{t-1}^{-1}a_t > \frac{d}{t},\ -a_t^{*\top}V_{t-1}^{-1}A_{t-1}\mathcal E_{t-1} \geq \frac{\Delta_t}{2} + (a_t^{*\top}V_{t-1}^{-1}a_t^*)\sigma\sqrt{\frac{\eta t}{d}}\right)
\end{align*}
We bound each term separately. We have
\begin{align*}
    & \quad \mbP\left(\Delta_t \geq \frac{x-4\sqrt{d}}{4t\ln T},\ a_t^\top V_{t-1}^{-1}a_t\leq d/t,\ a_t^\top V_{t-1}^{-1}A_{t-1}\mathcal E_{t-1} \geq \frac{\Delta_t}{2} - (a_t^\top V_{t-1}^{-1}a_t)\sigma\sqrt{\frac{\eta t}{d}} - 2\sqrt{d(a_t^\top V_{t-1}^{-1}a_t)}\right) \\
    & \leq \mbP\left(a_t^\top V_{t-1}^{-1}a_t\leq d/t,\ a_t^\top V_{t-1}^{-1}A_{t-1}\mathcal E_{t-1} \geq \frac{x-4\sqrt{d}}{8t\ln T} - \sigma\sqrt{\eta d/t} - 2d/\sqrt{t}\right) \\
    & \leq \mbP\left(\frac{|a_t^\top V_{t-1}^{-1}A_{t-1}\mathcal E_{t-1}|}{\sqrt{a_t^\top V_{t-1}^{-1}a_t}}\geq \frac{\left(\frac{x-4\sqrt{d}}{8t\ln T} - \sigma\sqrt{\eta d/t}-2d/\sqrt{t}\right)_+}{\sqrt{d/t}}\right) \\
    & \leq \mbP\left(\sqrt{(A_{t-1}\mathcal E_{t-1})^\top  V_{t-1}^{-1}A_{t-1}\mathcal E_{t-1}} \geq \frac{\left(x-4\sqrt{d} - 16d\sqrt{T}\ln T - 8\sigma\sqrt{\eta dT}\ln T\right)_+}{8\sqrt{dT}\ln T}\right) \\
    & \leq (T/d)^{2d}\exp\left(-\frac{\left(x-4\sqrt{d} - 16d\sqrt{T}\ln T - 8\sigma\sqrt{\eta dT}\ln T\right)_+^2}{128\sigma^2dT\ln^2T}\right)
\end{align*}
and
\begin{align*}
    & \quad \mbP\left(\Delta_t \geq \frac{x-4\sqrt{d}}{4t\ln T},\ a_t^\top V_{t-1}^{-1}a_t\leq d/t,\ -a_t^{*\top}V_{t-1}^{-1}A_{t-1}\mathcal E_{t-1} \geq \frac{\Delta_t}{2} + (a_t^{*\top}V_{t-1}^{-1}a_t^*)\sigma\sqrt{\frac{\eta t}{d}}\right) \\
    & \leq \mbP\left(-a_t^{*\top}V_{t-1}^{-1}A_{t-1}\mathcal E_{t-1} \geq \sqrt{2\frac{x-4\sqrt{d}}{4t\ln T}(a_t^{*\top}V_{t-1}^{-1}a_t^*)\sigma\sqrt{\frac{\eta t}{d}}}\right) \\
    & \leq \mbP\left(\frac{|a_t^\top V_{t-1}^{-1}A_{t-1}\mathcal E_{t-1}|}{\sqrt{a_t^\top V_{t-1}^{-1}a_t}} \geq \sqrt{\frac{(x-4\sqrt{d})_+\sigma\sqrt{\eta}}{2\sqrt{dT}\ln T}}\right) \\
    & \leq \mbP\left(\sqrt{(A_{t-1}\mathcal E_{t-1})^\top  V_{t-1}^{-1}A_{t-1}\mathcal E_{t-1}} \geq \frac{\sqrt{(x-4\sqrt{d})_+\sigma\sqrt{\eta}}}{\sqrt{2\sqrt{dT}\ln T}}\right) \\
    & \leq (T/d)^{2d}\exp\left(-\frac{(x-4\sqrt{d})_+\sqrt{\eta}}{4\sigma\sqrt{dT}\ln T}\right)
\end{align*}
and
\begin{align*}
    & \quad \mbP\left(\frac{\Delta_t}{a_t^\top V_{t-1}^{-1}a_t}\geq \frac{x-4\sqrt{d}}{8d\ln T}, \ a_t^\top V_{t-1}^{-1}a_t > \frac{d}{t}, \ a_t^\top V_{t-1}^{-1}A_{t-1}\mathcal E_{t-1} \geq \frac{\Delta_t}{2} - (a_t^\top V_{t-1}^{-1}a_t)\sigma\sqrt{\frac{\eta t}{d}} - 2\sqrt{d(a_t^\top V_{t-1}^{-1}a_t)}\right) \\
    & = \mbP\left(\frac{\Delta_t}{a_t^\top V_{t-1}^{-1}a_t}\geq \frac{x-4\sqrt{d}}{8d\ln T}, \ a_t^\top V_{t-1}^{-1}a_t > \frac{d}{t}, \ \frac{a_t^\top V_{t-1}^{-1}A_{t-1}\mathcal E_{t-1}}{\sqrt{a_t^\top V_{t-1}^{-1}a_t}\sqrt{a_t^\top V_{t-1}^{-1}a_t}} \geq \frac{\Delta_t}{2a_t^\top V_{t-1}^{-1}a_t} - \sigma\sqrt{\frac{\eta t}{d}} - 2\sqrt{t}\right) \\
    & \leq \mbP\left(a_t^\top V_{t-1}^{-1}a_t > \frac{d}{t}, \ \frac{a_t^\top V_{t-1}^{-1}A_{t-1}\mathcal E_{t-1}}{\sqrt{a_t^\top V_{t-1}^{-1}a_t}\sqrt{a_t^\top V_{t-1}^{-1}a_t}} \geq \frac{x-4\sqrt{d}}{16d\ln T} - \sigma\sqrt{\frac{\eta t}{d}} - 2\sqrt{t}\right) \\
    & \leq \mbP\left(\frac{|a_t^\top V_{t-1}^{-1}A_{t-1}\mathcal E_{t-1}|}{\sqrt{a_t^\top V_{t-1}^{-1}a_t}} \geq \left(\frac{x-4\sqrt{d}}{16d\ln T} - \sigma\sqrt{\frac{\eta t}{d}} - 2\sqrt{t}\right)_+\sqrt{\frac{d}{t}}\right) \\
    & \leq \mbP\left(\sqrt{(A_{t-1}\mathcal E_{t-1})^\top  V_{t-1}^{-1}A_{t-1}\mathcal E_{t-1}} \geq \frac{\left(x-4\sqrt{d}-32d\sqrt{T}\ln T - 16\sigma\sqrt{\eta dT}\ln T\right)_+}{16\sqrt{dT}\ln T}\right) \\
    & \leq (T/d)^{2d}\exp\left(-\frac{\left(x-4\sqrt{d} - 32d\sqrt{T}\ln T - 16\sigma\sqrt{\eta dT}\ln T\right)_+^2}{512\sigma^2dT\ln^2T}\right)
\end{align*}
and
\begin{align*}
    & \quad \mbP\left(\frac{\Delta_t}{a_t^\top V_{t-1}^{-1}a_t}\geq \frac{x-4\sqrt{d}}{8d\ln T},\ a_t^\top V_{t-1}^{-1}a_t > \frac{d}{t}, -a_t^{*\top}V_{t-1}^{-1}A_{t-1}\mathcal E_{t-1}\geq \frac{\Delta_t}{2} + (a_t^{*\top}V_{t-1}^{-1}a_t^*)\sigma\sqrt{\frac{\eta t}{d}}\right) \\
    & \leq \mbP\left(-a_t^{*\top}V_{t-1}^{-1}A_{t-1}\mathcal E_{t-1}\geq \sqrt{2\frac{x-4\sqrt{d}}{8t\ln T}(a_t^{*\top}V_{t-1}^{-1}a_t^*)\sigma\sqrt{\frac{\eta t}{d}}}\right) \\
    & \leq \mbP\left(\frac{|a_t^{*\top} V_{t-1}^{-1}A_{t-1}\mathcal E_{t-1}|}{\sqrt{a_t^{*\top} V_{t-1}^{-1}a_t^*}} \geq \sqrt{\frac{(x-4\sqrt{d})\sqrt{\eta}}{4\sqrt{dT}\ln T}}\right) \\
    & \leq \mbP\left(\sqrt{(A_{t-1}\mathcal E_{t-1})^\top  V_{t-1}^{-1}A_{t-1}\mathcal E_{t-1}} \geq \frac{\sqrt{(x-4\sqrt{d})_+\sigma\sqrt{\eta}}}{\sqrt{4\sqrt{dT}\ln T}}\right) \\
    & \leq (T/d)^{2d}\exp\left(-\frac{(x-4\sqrt{d})_+\sqrt{\eta}}{8\sigma\sqrt{dT}\ln T}\right).
\end{align*}

Plugging the four bounds above into \eqref{eq:linear-decompose} yields the final result
\begin{align*}
    \sup_{\theta}\mbP(\hat R_{\theta}^\pi(T)\geq x) & \leq \exp\left(-\frac{x^2}{2\sigma^2d^2T}\right) + 2d(T/d)^{2d+1}\exp\left(-\frac{\left(x-4\sqrt{d} - 32d\sqrt{T}\ln T - 16\sigma\sqrt{\eta dT}\ln T\right)_+^2}{512\sigma^2dT\ln^2T}\right) \\
    & \quad\quad + 2d(T/d)^{2d+1}\exp\left(-\frac{(x-4\sqrt{d})_+\sqrt{\eta}}{8\sigma\sqrt{dT}\ln T}\right).
\end{align*}

\end{APPENDICES}
%
%


\end{document}